\documentclass[lettersize,journal]{IEEEtran}

\usepackage{amsmath,amsfonts}
\usepackage{algorithmic}
\usepackage{array}
\usepackage[caption=false,font=normalsize,labelfont=sf,textfont=sf]{subfig}
\usepackage{textcomp}
\usepackage{stfloats}
\usepackage{url}
\usepackage{graphicx}

\usepackage{balance}
\usepackage{xspace} 
\makeatletter
\DeclareRobustCommand\onedot{\futurelet\@let@token\@onedot}
\def\@onedot{\ifx\@let@token.\else.\null\fi\xspace}

\def\eg{\emph{e.g}\onedot} 
\def\ie{\emph{i.e}\onedot}

\makeatother

\usepackage{amssymb}
\usepackage{booktabs}
\usepackage{siunitx}
\usepackage{placeins}

\usepackage{tikz}
\usepackage{pgfplots}
\usetikzlibrary{spy,calc}
\usetikzlibrary{positioning}

\usepackage{adjustbox}

\usepackage{multirow}
\usepackage{multicol}

\usepackage[final]{changes} 

\usepackage{ifthen}
\usepackage{hyperref}

\usepackage[
backend=biber,
style=numeric, 
isbn=false,
url=false,
doi=false,
sorting=none
]{biblatex}
  
\AtEveryBibitem{\clearfield{language}}
\AtEveryBibitem{\clearfield{doi}}
\AtEveryBibitem{\clearfield{pages}}
\AtEveryBibitem{\clearfield{note}}
\AtEveryBibitem{\clearfield{pagetotal}}

\AtEveryBibitem{
  \ifentrytype{inproceedings}{
    \clearfield{series}
    \clearfield{location}
    \clearlist{publisher}
    \clearlist{location}
    \clearfield{publisher}
    \clearfield{editor}
    \clearlist{editor}
    \clearname{editor}
    \clearfield{url}
    \clearfield{urldate}
    \clearfield{volume}
    \iffieldundef{booktitle}{}{\clearfield{eventtitle}}
    \clearfield{language}
  }{}
}

\AtEveryBibitem{\ifentrytype{article}{
    \clearfield{url}
    \clearfield{eprint}
    }{}
}
\AtEveryBibitem{\ifentrytype{misc}{\clearfield{url}\iffieldundef{journaltitle}{}{\clearfield{eprint}\clearfield{eprinttype}}}{}}

\newenvironment{matplotlibFont}{\fontfamily{bch}\selectfont}{\par} 
\DeclareTextFontCommand{\textmyfont}{\matplotlibFont}

\newcommand{\sep}{-0.2cm}
\newcommand{\sepPlot}{-0.025cm} 
\newcommand{\fsize}[1]{\tiny{\textmyfont{#1}}}

\newenvironment{mysubfigure}[3]{
  \begin{tabular}{@{}c@{}}
    #2 \\[\abovecaptionskip]
    \small #3
  \end{tabular}
}{}

\begin{document}
\title{Examining the Impact of Optical Aberrations to Image Classification and Object Detection Models}
\author{Patrick M\"uller~\IEEEmembership{Student Member, IEEE}, Alexander Braun, 
Margret Keuper
\thanks{Patrick M\"uller is with the University of Siegen, 57068 Siegen, Germany, E-mail: patrick.mueller@uni-siegen.de, 
Alexander Braun is with the Hochschule D\"usseldorf, University of Applied Sciences, M\"unsterstraße 156, 40476 D\"usseldorf, Germany, E-Mail: alexander.braun@hs-duesseldorf.de, %
Margret Keuper is with the University of Mannheim, 68165 Mannheim, Germany, and
also with the Max Planck Institute for Informatics, Saarland Informatics
Campus, 66123 Saarbr\"{u}cken, Saarland, Germany. 
E-mail: \url{margret.keuper@uni-mannheim.de} \\
Patrick Müller and Margret Keuper acknowledge funding from the DFG research unit DFG-FOR 5336 ``Learning to Sense". 
Our code is available at \url{https://github.com/PatMue/classification_robustness}. \\
 }} 


\maketitle

\begin{abstract}
Deep neural networks (DNNs) have proven to be successful in various computer vision applications such that models even infer in safety-critical situations. 
Therefore, vision models have to behave in a robust way to disturbances such as noise or blur. While seminal benchmarks exist to evaluate model robustness to diverse corruptions, blur is often approximated in an overly simplistic way to model defocus, while ignoring the different blur kernel shapes that result from optical systems. 
To study model robustness against realistic optical blur effects, this paper proposes two datasets of blur corruptions, which we denote OpticsBench and LensCorruptions. 
OpticsBench examines primary aberrations such as coma, defocus, and astigmatism, i.e. aberrations that can be represented by varying a single parameter of Zernike polynomials. 
To go beyond the principled but synthetic setting of primary aberrations, LensCorruptions samples linear combinations in the vector space spanned by Zernike polynomials, corresponding to 100 real lenses.
Evaluations for image classification and object detection on ImageNet and MSCOCO show that for a variety of different pre-trained models, the performance on OpticsBench and LensCorruptions varies significantly, indicating the need to consider realistic image corruptions to evaluate a model's robustness against blur. 
\end{abstract}

\begin{IEEEkeywords}
DNN, PSF, Blur, Robustness, Object detection, Image Classification, Data Augmentation, Camera Lens
\end{IEEEkeywords}

\section{Introduction}
\IEEEPARstart{O}{ptical aberration} effects are inevitable in practice~\cite{born_principles_1999,noauthor_handbook_2006}. Under high cost pressure, \eg in automotive mass production of cameras, compromises in optical quality can not be avoided due to production tolerances~\cite{noauthor_handbook_2006,smith_modern_2000,yoder_opto_mechanical_2018}. Further, over the lifetime of a camera, changes to image quality occur, \eg due to thermal expansion and aging, leading to increased optical aberrations~\cite{yoder_opto_mechanical_2018,siew_perspectives_2023}. End-of-line (EOL) the produced camera is tested to ensure that these aberrations stay within the specified limits. Given a large number of cameras,  these optical test limits thus define the business case, as 'bad' cameras are scrapped, destroying maximum value-add~\cite{braun_automotive_2022}.
Realistic optical effects are likely to challenge the robust behavior of learned vision models. Yet, while the literature has studied diverse test cases for natural distribution shifts~\cite{hendrycks_benchmarking_2019,michaelis_benchmarking_2020,kar_3d_2022,vasiljevic2016examining}, model robustness to \emph{realistic} optical aberration effects has not yet been thoroughly investigated. 
In this paper, we propose two test frameworks,  \emph{OpticsBench}~\cite{mueller_opticsbench2023} and \emph{LensCorruptions} to close this gap. 

OpticsBench is a benchmark that includes primary optical aberrations such as coma, astigmatism and spherical aberration as image corruptions. To additionally cover realistic combinations of such effects, LensCorruptions for the first time comprises a wide range of real lens types and qualities that we show to impact image classification and object detection. The 100 selected image corruptions are based on realistic lens kernels obtained by tracing rays according to the Huygens principle~\cite{born_principles_1999,noauthor_opticstudio_nodate}, spatially resolved over the field of view and parameterised by radius and azimuth.  
\begin{figure}
    \centering
    \begin{tikzpicture}
        \node (image) at (0,0){\includegraphics[page=5,trim=3.2cm 23cm 3.2cm 2cm,clip,width=.975\linewidth]{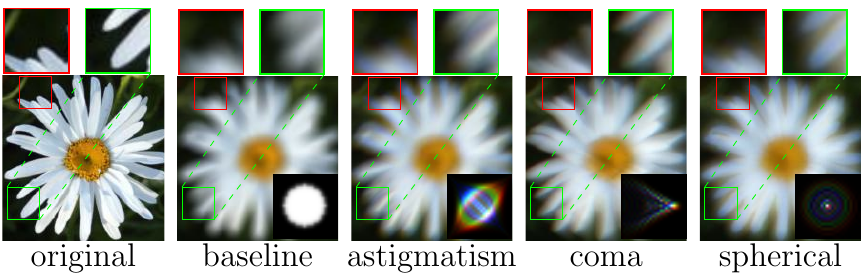}};
    \end{tikzpicture}
    \caption{Blur image corruptions applied to an ImageNet image. 
    The effects of baseline blur~\cite{hendrycks_benchmarking_2019} and OpticsBench blur kernels~\cite{mueller_opticsbench2023} are visualized for severity 4. Although the image looks similarly blurred for the different kernels, the details are different: for example, the petals of the flower remain white for the baseline blur, while they have reddish and bluish color fringes for the multi-channel (R, G, B) primary aberrations considered in OpticsBench. The baseline is uniformly blurred, while astigmatism and coma introduce directional blur. In this article, we investigate how realistic blur kernel properties impact current image classification and object detection models.}
    \label{fig:teaser}
\end{figure}
In both cases the optical kernels are derived theoretically from an expansion of the wavefront of light into Zernike polynomials, which can be mixed by the user to create any realistic optical kernel. The basis for the dataset are 3D kernels (x, y, color) matched in size to the defocus blur corruption from~\cite{hendrycks_benchmarking_2019}, which is generated using symmetric, disk-shaped monochromatic kernels. 
We evaluate more than 70 different vision models on a total of 1M images from the benchmark applied to ImageNet. Our evaluation shows that the performance of ImageNet models varies for optical degradations and that the disk-shaped blur kernels provide only weak proxies to estimate the models' robustness when confronted with optical degradations.

Further, since our analysis indicates a lack of robustness to optical corruptions, we propose an efficient tool to use optical kernels for data augmentation during model training, which we denote \emph{OpticsAugment}. 
In experiments on ImageNet-100, OpticsAugment achieves on average 18\% performance gain compared to conventionally trained DNNs on OpticsBench.
We also show that OpticsAugment allows to improve on 2D common corruptions~\cite{hendrycks_benchmarking_2019} on average by 5.3\% points on ImageNet-100, \ie the learned robustness transfers to other domain shifts.

Our main contributions can be summarized as follows:
\begin{itemize}
    \item We provide two different sets of blur corruptions: 
    OpticsBench for analyzing single primary aberrations, complemented by
    LensCorruptions for testing against mixed aberrations obtained from real lenses. Both types of blur corruption are extensively evaluated on image classification (ImageNet~\cite{russakovsky_imagenet_2015}) and object detection datasets (MSCOCO~\cite{fleet_microsoft_2014}, NuImages~\cite{caesar_nuscenes_2020}).
    \item By systematically collecting, sorting and curating the large lens database, we ensure that our sampling of 100 lenses covers the whole range of optical qualities encountered in more than 718 lenses, yielding a comprehensive variety of realistic blur kernels. 
    \item Additionally, we provide a data augmentation method (OpticsAugment) showing that compensation of the specific blur kernels is possible to a large extent.
\end{itemize}

\section{Related work}
\label{sec:related_work}
Vasiljevic et al.~\cite{vasiljevic2016examining} investigate the robustness of CNNs to defocus and camera shake. Hendrycks et al.~\cite{hendrycks_benchmarking_2019} provide a benchmark to 2D common corruptions. The benchmark includes several general modifications to images such as change in brightness or contrast as well as weather influences such as fog, frost and snow. They also include different types of blur, but only consider luminance kernels and more general types such as Gaussian or disk-shaped kernels.
Kar et al.~\cite{kar_3d_2022} build on this work and extend common corruptions to 3D. These include \emph{extrinsic} camera parameter changes such as field of view changes or translation and rotation. Michaelis et al.~\cite{michaelis_benchmarking_2020} and Dong et al.~\cite{dong_benchmarking_2023} provide robustness benchmarks for object detection on common vision datasets.
As the field of research grows, more subtle changes in image quality, potentially posing a distribution shift, are uncovered. This work aims to build on a more general treatment of blur types with OpticsBench~\cite{mueller_opticsbench2023}, which are known in optics research, but less common in computer vision. OpticsBench has been previously published as an ICCV 2023 Workshop Paper. Here, we propose a largely consolidated and extended version, including an improved evaluation of OpticsBench on Optics Detection. In addition, we extend the systematic evaluation of blur types by introducing LensCorruptions offering to test against realistic lens models.

A related field of research investigates robustness to adversarial examples created by targeted~\cite{croce_reliable_2020,moosavi-dezfooli_deepfool_2016} and untargeted~\cite{andriushchenko_square_2020} attacks. The goal is to introduce small perturbations of the input data in a way such that the model makes wrong classifications. Successful attacks pose a security risk to a particular DNN, while human observers would not even notice a difference and safely classify~\cite{carlini_adversarial_2017}.
Croce et al.~\cite{croce2021robustbench} provide a robustness benchmark, originally intended for adversarial robustness testing using AutoAttack~\cite{croce_reliable_2020}, while more practical $l_p$-bounds are discussed in~\cite{lorenz2022is}.
However, these methods are model specific in that the particular attack is \emph{optimized} for the model using \eg projected gradient descent in the backward pass. Therefore white-box methods require full knowledge about the underlying model. 
In contrast, model evaluation 
with OpticsBench corruptions can be done by applying simple filters to the validation data and thus requires only a clean image dataset. Therefore, it also works for black-box models. Convolving tensors with kernels is a base task in computer vision and so GPU-optimized implementations exist. These include parallel evaluation of OpticsBench for many models and on-the-fly training with OpticsAugment~\cite{mueller_opticsbench2023}. 

Since these benchmarks reveal potential distribution shifts or lack of robustness for a given model, concurrent methods are researched to improve robustness~\cite{gowal2021improving,geirhos2018imagenet,cubuk_autoaugment_2019} on various benchmarks. Hendrycks et al.~\cite{hendrycks_augmix_2020,hendrycks_many_2021} propose different data augmentation methods to improve classification robustness towards 2D common corruptions. Saikia et al.~\cite{saikia_improving_2021} further build on this and achieve high accuracy on both clean and corrupted samples. Similarly, methods exist to improve adversarial robustness~\cite{salman2020adversarially} on benchmarks like~\cite{PINTOR2023109064,croce2021robustbench}.

Modeling or retrieving Zernike polynomials~\cite{born_principles_1999,zernike_beugungstheorie_1934} is a common process in ophthalmological optics~\cite{thibos_retinal_2009,iskander_optimal_2001} to investigate aberrations of the human eye. The expansion is also widely used in other areas of optics, such as lens design~\cite{noauthor_opticstudio_nodate,laikin_lens_2007}, microscopy~\cite{cumming_direct_2020} and astronomy~\cite{ndiaye_calibration_2013,lane_wave-front_1992}. Therefore several tools~\cite{noauthor_opticstudio_nodate,dube_prysm_2019,kirshner2011} exist to generate PSFs from Zernike coefficients or wavefronts. The design is application specific and intended for optical engineers or physicists such as optical design software like Zemax Optic Studio~\cite{noauthor_opticstudio_nodate}. 
Our OpticsBench and OpticsAugment leverage such optical models. Further LensCorruptions uses realistic lens blur obtained from  different lens designs to create realistic optical effects.

\section{Optical aberrations}
\label{sec:optical_aberrations}
Optical aberrations are diverse imperfections that occur to varying degrees in any real lens and can degrade image quality in a camera lens system~\cite{phillips_camera_2018}. 
Aberrations affect the (perceived) sharpness of an image and 
give it a characteristic lens fingerprint. This fingerprint results mainly from the physical dimensions, shape, arrangement and material of the different lens elements, and can be divided into chromatic and monochromatic aberrations. Most common  aberrations, \eg spherical aberrations, coma, and astigmatism, result in a blurred\footnote{Distortion, which geometrically deforms the image, is not considered in this article, as it is an orthogonal question.} image, while each aberration has its own characteristic symmetry properties~\cite{noauthor_handbook_2005,hecht_optics_2017}. 
Modern lens optimization and manufacturing methods can produce lenses with negligible aberrations. 
However, depending on the particular application, budget constraints and other lens specification requirements result in aberration-limited lenses from a wide range of lens qualities, which may have significant aberrations~\cite{noauthor_handbook_2006,braun_automotive_2022,noauthor_handbook_2008}.

The characteristic blur fingerprint of a lens is represented by its \textit{point spread function} (PSF) as visualized in Fig.~\ref{fig:psf_sources_and_image_processing} (left) and can be modeled. Superimposing the input image and the space-variant PSF yields the blurred output image. For small image regions, the space-variant lens blur can be approximated  with a convolution~\cite{nagy_fast_1997} as displayed in Fig.~\ref{fig:psf_sources_and_image_processing} (right). Generally, the PSF depends on wavelength, angle of incidence, and distance in such a way that the observed PSF generally varies from the center of the lens to the edge, and with object distance~\cite{born_principles_1999,hecht_optics_2017,goodman_introduction_2017}. Beyond the lens-dependent hyperfocal distance for fixed-focus lenses~\cite{siew_perspectives_2023}, the PSF varies only with angle and wavelength and the depth dependence can then be dropped. We assume this case here for simplicity. 

\subsection{Optical quality metrics}  
\begin{figure}
    \centering
    \includegraphics[width=\linewidth]{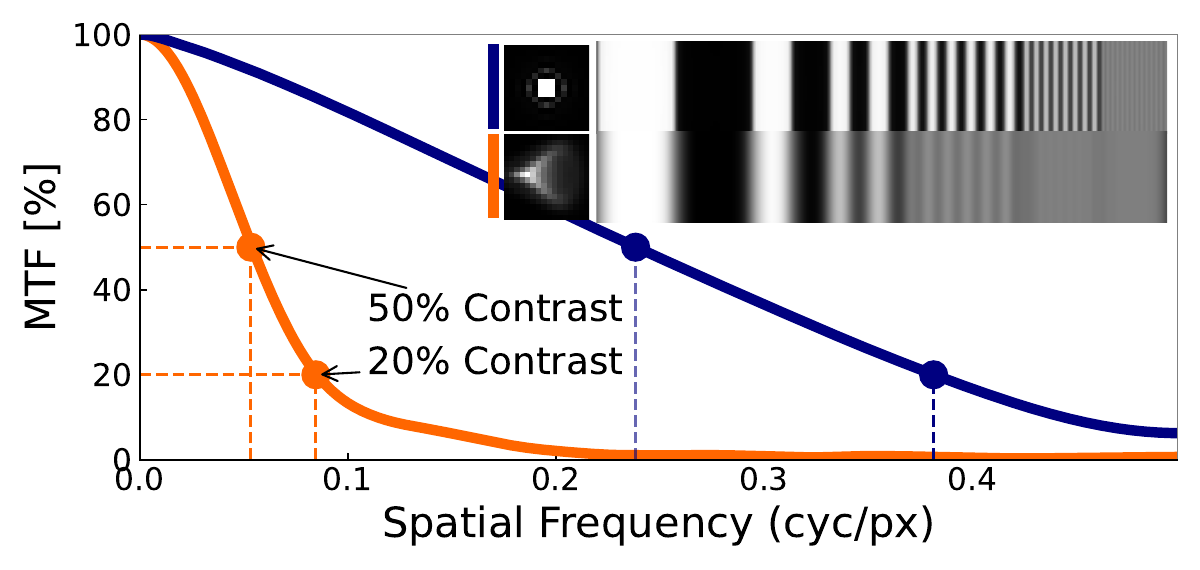}
    \caption{The Modulation Transfer Function (MTF) and its derived metrics, MTF50 and MTF20 frequencies (markers), objectively measure lens quality in terms of sharpness. The higher the contrast, the better, so the higher the frequency assigned to either MTF50 or MTF20, the better. 
    The effect of each MTF is illustrated by convolving the bar targets at the top right with the corresponding PSFs. The six spatial frequencies (cyc/px) double from left to right, while the far right represents the maximum frequency - the Nyquist frequency (0.5 cyc/px). 
    The black MTF represents a high quality lens with high contrast transfer and an MTF50 frequency around \SI{0.25}{cyc/px} (half Nyquist), visible in the bar target's penultimate frequency bin. The orange MTF drops off quickly with an MTF50 frequency around \SI{0.0625}{cyc/px} visible in bar target's third frequency bin. Below MTF20 or even MTF10, the bar patterns are hardly discernible and turn all grayish.}
    \label{fig:mtf_demo}
\end{figure}

There exist many different metrics to assess lens quality. Usually they are based on the system's PSF (Strehl ratio, encircled energy), wavefront (\eg~RMS wavefront error, Zernike Polynomials) or the Modulation Transfer Function (MTF). 
The MTF measures the contrast transfer vs.~the spatial frequency and can be  obtained by taking the normalized magnitude of the Fourier transformed PSF (here). In practical applications, the MTF can also be obtained from test targets with specific features, see \eg~\cite{boreman_modulation_2001,koren_correcting_2020} or from commercial MTF measurement hardware. 
Fig.~\ref{fig:mtf_demo} shows an MTF of a high-performing (blue) and low-performing (orange) lens together with their effect on a test image using the corresponding small and large PSFs. The blur effect reduces contrast according to the decreasing MTF curve with higher frequencies. So, the smaller the detail, the harder it is to distinguish. 
Although the whole curve is informative, summary metrics are often more practical.   
The most common are either the contrast at a given frequency, such as the half Nyquist frequency, or conversely, the frequency at a fixed contrast value. For the latter, the MTF50 frequency at 50\% modulation is widely used in industry and lens design as a sharpness criterion ~\cite{laikin_lens_2007,koren_correcting_2020}. Lower contrast fidelity criteria measure the resolution of the lens, such as the MTF10 or MTF20 frequencies. 
All of these criteria are quality metrics in the sense that the higher the value, the better the performance. Typically, multiple metrics are used to provide a more comprehensive view of quality. 

Lens quality can also be analyzed by decomposing the wavefront into Seidel or Zernike polynomials. This provides insight into which aberrations contribute to the observed PSF, as the Zernike coefficients are directly related to the classical aberrations~\cite{noauthor_handbook_2005}. Table~\ref{tab:coefficients} shows how the first 13 Zernike coefficients are converted to aberrations. 
While the first nine represent the primary aberrations and defocus, higher coefficients refer to higher order aberrations.

\subsection{Lens and system design}
Opto-mechanical system design is a complex and iterative process that has to meet various, often contradictory requirements for an overall camera system performance. The knowledge and exact procedure is often confidential and proprietary information. 
Usually, the field of view, sensor size and resolution -- which defines the Nyquist frequency -- are determined by the intended application~\cite{phillips_camera_2018,siew_perspectives_2023}. Design specifications for camera lenses are derived from further application requirements, usually defining various field points at which the lens MTF is measured and evaluated. 
Typically, test limits are based on specific contrast values such as MTF50 or MTF20. 
In our case, since the lens is already available, we set the Nyquist frequency~\cite{gonzalez_digital_2009,noauthor_handbook_2005,loffler-mang_handbuch_2020} of the system to the mean MTF20 value of different measurement points. Critically, this fixes an otherwise free parameter in our study, the pixel size. As a trade-off, this would still allow objects with low contrast to be detected by the sensor, but avoids too large pixels.

\subsection{PSF sources}
\label{subsec:psf_sources}
\begin{figure}
    \centering
\includegraphics[trim=0cm 0.32cm 0cm 0.4cm,clip,width=\linewidth]{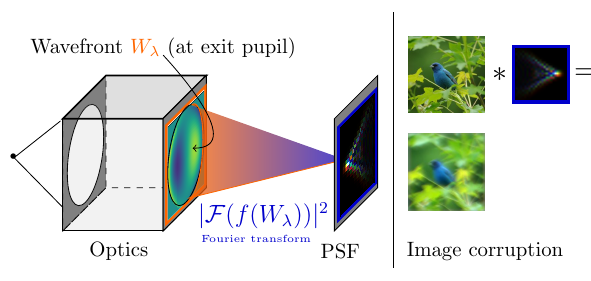}
    \caption{
    (Simplified view) Lens elements are lumped into a black box (left) and its output is represented by the exit pupil~\cite{goodman_introduction_2017}. The PSF (enlarged for display purposes) is obtained by applying a nonlinear operation to the wavefront aberration $W_{\lambda}$ at the exit pupil. 
    We simulate $W_{\lambda}$ using either ray-tracing or Zernike polynomials for different lenses and field positions. The PSFs can be used to generate our image blur corruptions.}
    \label{fig:psf_sources_and_image_processing}
\end{figure}
We consider two sources to obtain the intensity point spread function: tracing rays through an optical system based on Huygens principle~\cite{born_principles_1999,noauthor_opticstudio_nodate}, and from Zernike polynomials~\cite{born_principles_1999}. Both model the wavefront aberration $W_{\lambda}$ in the exit pupil function and obtain a PSF from this function~\cite{goodman_introduction_2017} as visible in Fig.~\ref{fig:psf_sources_and_image_processing}. The approaches differ mainly in the data source, which is either synthetic or from real lens prescriptions. Another approach is to measure the PSF of a lens using specialized hardware. We argue here that such measured lenses differ only in manufacturing tolerances, and therefore the nominal 
models from optical design give a good insight into real lenses. A simple example showing tolerances is given in the supplementary material~\ref{app:production_tolerances}.

With our first approach, forming OpticsBench, we obtain kernel shapes that represent real optical aberrations using a linear system model as in \cite{goodman_introduction_2017,born_principles_1999}, and define the wavefront aberration $W_\lambda$ by Zernike polynomials~\cite{born_principles_1999,noauthor_handbook_2005}: 
\begin{equation}
W_{\lambda}(\rho,\varphi,\lambda) = \lambda \cdot \sum_{i} A_i(\lambda) \cdot Z_i(\rho,\varphi)
\label{eq:zernike_expansion}
\end{equation}
The wavefront $W_\lambda$ is expanded into a complete and orthogonal set of polynomials $Z_i$. The polynomials $Z_i$ are defined on the unit circle and are decomposed into a radial and azimuthal factor~\cite{born_principles_1999}.
The definition is given in the supplementary material~\ref{app:optical_aberrations}.
Each coefficient $A_i$ in multiples of the  wavelengths $\lambda_k$ represents the contribution of a particular type of aberration and therefore different aspects such as the amount of coma, astigmatism or defocus can be turned off or on. 
From this, a PSF is derived using the Fourier transform, which can then be applied to images~\cite{goodman_introduction_2017}. Details are given in~\cite{mueller_opticsbench2023} and in the supplementary material~\ref{app:optical_aberrations}.
\newcommand{\kWidth}{.9}
\begin{table}
\caption{Zernike Fringe coefficients and corresponding monochromatic aberration. The coefficients 1-3 represent an offset (Piston) and shifts (Tilt X, Y), which we exclude here.}
 \label{tab:coefficients}
\centering
\begin{tabular}{@{}p{1.5cm}p{1.5cm}@{}p{1.5cm}  p{1.5cm}@{}p{1.5cm}@{}}
\hline
{Defocus} & \multicolumn{2}{c}{Astigmatism (5, 6)} 
&
\multicolumn{2}{c}{Coma (7, 8)} \\ 
(4) & \multicolumn{2}{c}{(straight, oblique)} & \multicolumn{2}{c}{(horiz., vertical)}\\ 
\includegraphics[width=\kWidth\linewidth]{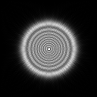} & \includegraphics[width=\kWidth\linewidth]{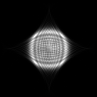} & 
\includegraphics[width=\kWidth\linewidth]{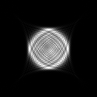} & 

\includegraphics[width=\kWidth\linewidth]{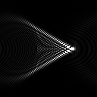} & 
\includegraphics[width=\kWidth\linewidth]{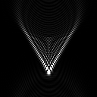} \\
Spherical & 
\multicolumn{2}{c}{Trefoil (10, 11)} & \multicolumn{2}{c}{2nd Astig. (12, 13)}\\
(9) & \multicolumn{2}{c}{(horiz., vertical)} &  \multicolumn{2}{c}{(straight, oblique)} \\
\includegraphics[width=\kWidth\linewidth]{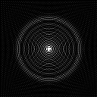} & \includegraphics[width=\kWidth\linewidth]{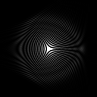} & 
\includegraphics[width=\kWidth\linewidth]{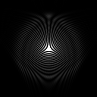} &
\includegraphics[width=\kWidth\linewidth]{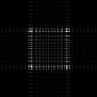} &
\includegraphics[width=\kWidth\linewidth]{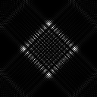} \\
\\
\hline
\end{tabular} 
\end{table}

With our second approach, defining the LensCorruptions dataset, we obtain the wavefront aberration $W_{\lambda}$ in Fig.~\ref{fig:image_processing}, by tracing a grid of rays through all optical elements from a lens prescription~\cite{noauthor_opticstudio_nodate}.  
Finally, the wavefront is traced to the image sensor and integrated, which yields the intensity Huygens PSF~\cite{noauthor_opticstudio_nodate}. 

The next two  sections~\ref{subsec:dataset_curation} and~\ref{subsec:lc_dataset_curation} describe how OpticsBench and LensCorruptions are constructed.

\section{Constructing OpticsBench}
\label{subsec:dataset_curation}
Similar to Hendrycks et al.~\cite{hendrycks_benchmarking_2019} we define five different severities for each optical aberration. To obtain comparable corruptions, the defocus blur corruption from \cite{hendrycks_benchmarking_2019} serves as baseline for OpticsBench~\cite{mueller_opticsbench2023}. 
For each severity, our kernels are matched to the defocus blur kernel types. For brevity, we refer to our kernels as \emph{optical kernels}, although the disk-shaped kernel type  from~\cite{hendrycks_benchmarking_2019} is a model for defocus obtained from geometric optics.
OpticsBench uses eight different Zernike modes that combine pairs of similar shapes into a single corruption. This results in four different optical corruptions. Each corrupted dataset is then obtained by randomly assigning one of the two kernel types per image and convolving the clean image with the kernel, where we use zero-padding. Results for reflective boundaries are given in the suplementary material~\ref{app:subsec:in1k_ob_reflective}. A predefined seed of the pseudo-random number generator ensures reproducibility. The different kernels are further discussed in the supplementary material~\ref{app:optics_bench_kernels}. 
Following~\cite{hendrycks_benchmarking_2019,noauthor_visionreferencesclassification_nodate}, before applying the blur kernels, the ImageNet~\cite{deng_imagenet_2009} images are resized to $256$ keeping the aspect ratio and then center cropped to $224\times224$, to avoid any reduction of the blur effect. For MSCOCO, we keep the original image sizes. More details are reported in the supplementary material~\ref{app:optics_bench}.

The Python code to re-create both the benchmark inference scripts and the benchmark datasets for all optical corruptions is provided\footnote{\url{https://github.com/PatMue/classification_robustness}}, including the set consisting of 40 kernels for OpticsBench. The script is intended for ImageNet-1k and ImageNet-100, but can be extended to other datasets. 
Generating the OpticsBench datasets (five severities, four corruptions) from the 50k ImageNet-1k validation images takes about 120 minutes with six PyTorch workers and batch size 128 on an 8-core i7-CPU and 32GB RAM equipped with NVIDIA GeForce 3080-Ti 12GB GPU. Using smaller kernels than ours ($25\times25\times3$) may speed up the process.
 
\section{Constructing LensCorruptions}
\label{subsec:lc_dataset_curation}
First, this section briefly describes the process of generating the lens blur kernel dataset \added{from a large number of real lens descriptions}. 
Then, we describe the process of constructing lens blur image corruptions (LensCorruptions) from the lens blur dataset.

\subsection{Lens dataset curation \& imaging model}
The collected lens blur dataset is based on publicly available photographic nominal lens prescriptions from the public domain database of the optical engineer Daniel J.  Reiley.~\cite{noauthor_navigation_nodate}
Several stages are carried out to obtain our final lens blur kernel dataset: we obtain realistic blur kernels by tracing rays using the Huygens principle~\cite{born_principles_1999,noauthor_opticstudio_nodate}, downsample these blur kernels and select a subset of the large database. The blur kernel database can then be used to simulate the lens blur effect on images. A detailed description of the process is given in the supplementary material~\ref{app:dataset_curation_details}. 
\paragraph*{Exporting \& preprocessing PSFs}
From 718 lens samples, Huygens PSFs for three different wavelengths are evaluated at high resolution. To resolve the spatial variance of a lens, we obtain the PSFs at five different distances from the lens center, and off-axis positions are additionally obtained at three different orientations. 
Then, we downsample the high-resolution PSFs to represent sampling with a virtual pixel size: during lens design, the lens PSF and camera pixel pitch are typically matched following a MTF-based criterion. 
Here, we relate the sensor's Nyquist frequency $f_N$ to 
the MTF20 value, which relates to vanishing resolution~\cite{garcia-villena_3d-printed_2021,loffler-mang_handbuch_2020}. With this process, we obtain pixel sizes common to CMOS and CCD sensors between \SIlist{1;20}{\micro\metre}~\cite{loffler-mang_handbuch_2020}. 
Each PSF is scaled by the corresponding pixel size with bicubic downsampling and anti-aliasing from the high resolution,
 cropped and aligned to the array center.

\begin{figure}
    \centering
    \begin{tabular}{@{}c@{}r@{}}
\subfloat{\label{fig:coefficient_space}
\begin{minipage}[c][\height][c]{0.7\linewidth}
\includegraphics[width=\linewidth]{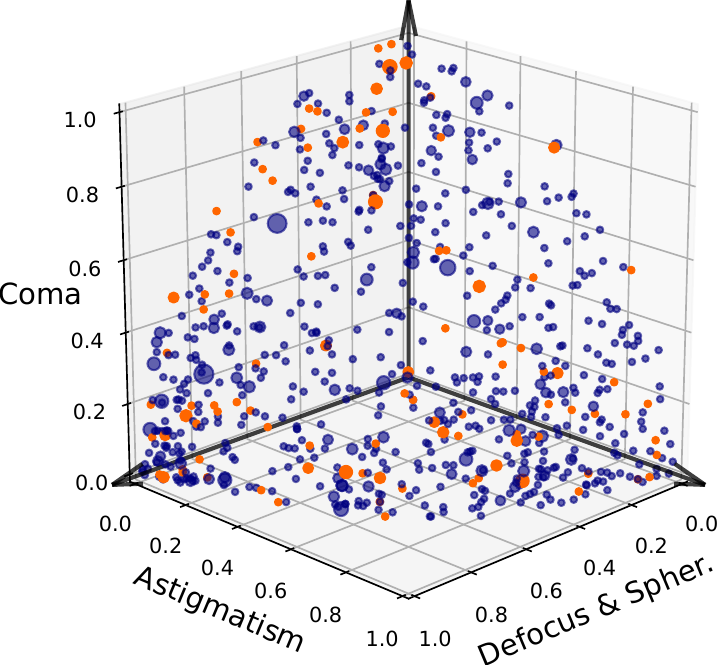}\end{minipage}}&
\hfill (a) \\
    \subfloat{\label{fig:lens_quality}
    \begin{minipage}[c][\height][c]{0.9\linewidth}
    \centering
\includegraphics[width=\linewidth]{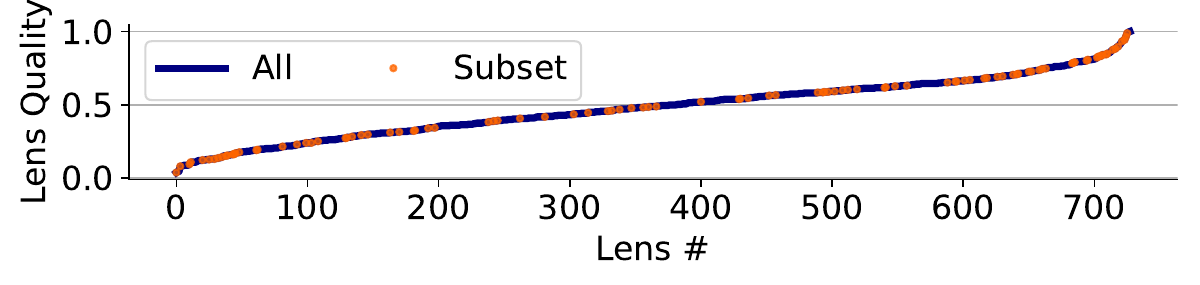}
\end{minipage}}&
\hfill
(b)
\end{tabular}
\caption{Lens selection: (a) 3D Zernike coefficient-space for the medium field position illustrating the diversity of the lens aberrations. The orange dots represent the 100 selected lenses, the blue ones the excluded ones. The lenses are chosen all over the wide-spread coefficient-space. The dot size is controlled by the logarithmic vector norm. Details are given in the supplementary material~\ref{app:dataset_curation_details}. (b) Sampling of a subset of 100 lenses (orange dots) from a wide range of qualities. The lens quality (y-axis) is derived from the downsampled MTFs (higher is better) and represents the normalized MTF50 obtained from MTFs averaged over field positions after downsampling. The x-axis represents the different lenses sorted by \added{lens} quality.}
\end{figure}

\paragraph*{Lens selection}
To keep experiments practical, a subset $(n=100)$ from the remaining lenses is selected that covers a wide range of optical qualities.  We define the lens quality as the average of the field dependent MTF50 value obtained from the MTFs retrieved from the downsampled PSFs.
We use the downsampled data because it correlates with the predicted image quality and therefore the observable effect strength.
First, all lenses are sorted by average quality over the imaging field. This sorting is then sampled equidistantly by quality and the nearest neighbor lens sample is selected. Fig.~\ref{fig:lens_quality} shows the selected subset from all available lenses. This fair sampling of all encountered optical qualities is a central contribution of this work, as it ensures that the diversity of reality is covered, while it allows for a straight-forward management of computational complexity and cost by selecting an appropriate number of samples.

\paragraph*{Lens properties \& coefficient space} 
To illustrate the selection, we 
provide a visualization of the lens dataset using the Zernike coefficients. Fig.~\ref{fig:coefficient_space} shows the distribution of all lenses (blue dots) and the chosen subset (orange dots) in a 3D Zernike coefficient space for a medium field position. For the visualization, we select three primary aberrations  (defocus \& spherical aberration, astigmatism, coma) as base vectors and map for each lens a normalized Zernike coefficient vector into this space showing the dominant category. The dot size refers to the logarithmic vector's magnitude. Details and the coefficient space distributions for all distances from the lens center are given in the supplementary material~\ref{app:dataset_curation_details}. 
The lens vectors in Fig.~\ref{fig:coefficient_space} show a variety of combinations of aberrations, with our selection spread across this coefficient category space. The base vectors (black) defined OpticsBench. 
In the supplementary material~\ref{app:dataset_curation_details}, we also report the general lens properties and the diversity of the selected aberrations for the most characterizing system properties of a lens, namely the field of view (FoV) and aperture, as well as the effective focal length (EFL). The lens selection returns samples from all types and various EFL.

\paragraph*{Imaging model}
The kernels from the dataset are applied to images using the convolution blur model from Sec.~\ref{sec:optical_aberrations}. 
For the simulation, we assume that a small image with negligible spatial variance is a section on a larger image sensor shifted according to the field position. We argue that this is a reasonable assumption since ImageNet or MSCOCO images of less than one megapixel are very small compared to standard multi-megapixel images.

\paragraph*{Limitations \& Discussion}
To summarize, the lens dataset consists of wavelength-dependent kernels, which are directly obtained from optical lens designs and are in this sense realistic. 
As each lens is sampled at five distances from the lens center and three different azimuth values, the space-variance property of lens blur is taken into account. 
For practical reasons, there are some restrictions. Besides depth, the simulation does not cover lens flare or scattering effects, nor polarization related effects. Although these effects are relevant, this study focuses on lens blur. Also, for simplicity, we do not add a color filter array.
In the absence of Exif data with sensor information for the web-crawled ImageNet or MSCOCO images, we had to assume a virtual pixel size to account for lens and sensor matching. We do not use a sensor noise or image signal processor (ISP) model for simplicity. However, our lens corruptions could be used as a starting point to investigate other aspects of imaging, such as sensor noise and color demosaicing. 
Despite the various simplifications, the simulation is capable of investigating realistic blur effects because it contains kernels sampled from real lenses. However, this study only highlights one aspect of the complex imaging process: optical blur aberrations. %

\subsection{Constructing LensCorruptions from the lens dataset}
\label{sec:lens_corruptions_construction}
We generate 100 image corruptions from the  selection of the lens dataset in Fig.~\ref{fig:lens_quality}. Each image corruption represents a complete lens and consists of five field positions with increasing distance to the image sensor's center. 
At each field position, a color PSF is randomly selected from the three azimuth positions $(r,x,y)$ and applied to an image using the convolution blur model described above. This results in a total of 500 new corrupted datasets per computer vision task, which we evaluate in Subsection~\ref{subsec:classification_lens_corruptions} for classification and in Subsection~\ref{subsec:detection_lens_corruptions} for object detection. Following~\cite{hendrycks_benchmarking_2019}, all processed images are saved with a very slight JPEG (quality$=0.9$) compression to keep a total number of \num{2.5e6} and \num{0.5e6} images from the ImageNet-100 and MSCOCO subset experiments manageable.

\section{Testing robustness to primary aberrations - \emph{OpticsBench}}
\subsection{Experiments on ImageNet}
\label{subsec:imagenet1k_opticsbench}
OpticsBench is built on the ImageNet validation dataset,
consisting of 50k images distributed in 1000 classes. All four corruptions (astigmatism, coma, defocus blur \& spherical and trefoil) are divided into 5 different severities each. For this article, we evaluate 65 DNNs from the torchvision model zoo with more recent updated weights from torchvision 0.15.2~\cite{noauthor_models_v015_2023} including a wide range of architectures such as the models from Tab.~\ref{tab:classification_opticsbench_overview}.
All models are pre-trained on ImageNet-1k and most use the standard methods for data augmentation of AutoAugment~\cite{cubuk_autoaugment_2019}, CutMix~\cite{yun_cutmix_2019} and MixUp~\cite{zhang_mixup_2018} combined with long training. A more comprehensive overview gives the supplementary material~\ref{app:models}.
In addition to these networks, we evaluate ResNet50 models from the RobustBench~\cite{croce2021robustbench} leaderboard, which are reportedly robust against common corruptions~\cite{hendrycks_augmix_2020,hendrycks_many_2021,erichson_noisymix_2022}. 

\begin{table}[htbp]
  \centering
  \caption{Accuracies on ImageNet-1k for pre-trained classification models on validation (clean) and OpticsBench (OB).}
  \scriptsize
\begin{tabular}{@{}l c c l c c@{}}
    DNN & Clean & OB & DNN & Clean & OB \\
    \hline     
    ConvNeXt\_large ~\cite{liu2022convnet} & \textit{80.1} & \textbf{49.9}     & 
        MobileNet\_v3\_l~\cite{howard_mobilenets_2017} & 71.7 & 32.3 \\

    ConvNeXt\_small~\cite{liu2022convnet} & 79.4 & 47.2 &     
    
        ResNet50~\cite{he_deep_2016} & 76.9 & 41.0 \\
    DenseNet\_169~\cite{huang_densely_2017} & 72.5 & 36.0 &                     
        ResNeXt50~\cite{xie_aggregated_2017} & 77.8 & 42.1 \\
        
    DenseNet\_201~\cite{huang_densely_2017} & 73.1 & 37.4 &

        Swin\_v2\_base~\cite{liu_swin_2022}& \textbf{80.3} & \textit{49.0} \\
         
    EfficientNet\_b0~\cite{tan_efficientnet_2019} & 73.8 & 35.3 &

        ViT\_base~\cite{ranftl_vision_2021} & 74.7 & 45.4 \\
        
    EfficientNet\_b3~\cite{tan_efficientnet_2019} & 79.3 & 47.6 &
    
        ViT\_large~\cite{ranftl_vision_2021} & 76.0 & \textit{48.8} \\
    EfficientNet\_b5~\cite{tan_efficientnet_2019} & 79.9 & 43.2 &

    $\Sigma$ & 76.6 & 42.7 \\

\end{tabular}
\label{tab:classification_opticsbench_overview}
\end{table}
Table~\ref{tab:classification_opticsbench_overview} lists a selection of diverse models from the full set of evaluated models on ImageNet-1k. We include large and small CNN and transformer models. The first column lists the accuracies on the validation set and the second column lists the average accuracy across the five severities and different OpticsBench corruptions. OpticsBench is intended to test the limits of recognition and is closely following the defocus blur corruption from~\cite{hendrycks_benchmarking_2019}: 
The overall accuracy halves almost from \SI{76.6}{\percent} to on average \SI{42.7}{\percent}. 
Swin\_v2 and ConvNeXt have the highest clean accuracies and also the highest accuracies on OpticsBench, while ViT\_l is more robust on OpticsBench. 
A complete list with accuracies for all 72 models is given in the supplementary material~\ref{app:opticsbench_classification}.

To broaden the perspective on the behavior of the different models and corruptions, the ranking with the baseline \emph{defocus blur} for severity 4 is discussed below for several aspects. The remaining severities are shown in the supplementary material~\ref{app:opticsbench_classification}.

\begin{figure*}
    \centering
    \includegraphics[trim=0cm 1cm 0cm  0cm,clip,width=\linewidth]{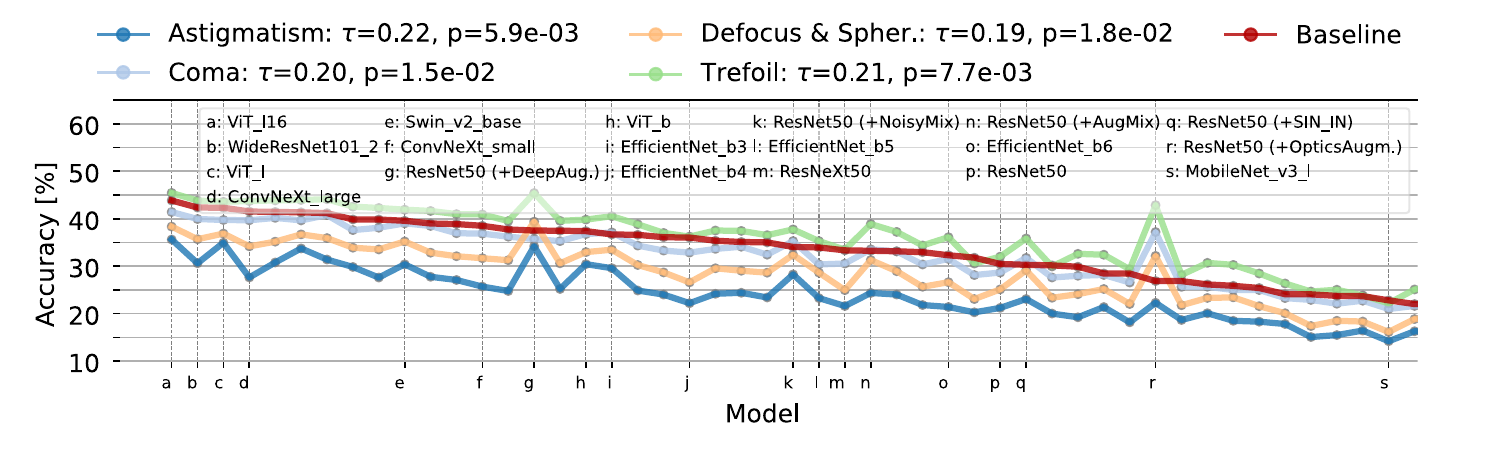}
    \caption{Ranking for the best 50 models on ImageNet-1k with respect to the baseline \emph{defocus blur} corruption from~\cite{hendrycks_benchmarking_2019}. The different peaks are mainly caused by the robust ResNet50 versions. The Kendall tau~\cite{kendall_treatment_1945} $\tau$ is computed between the rankings of the baseline and a specific corruption. All $\tau$ values report a weak correlation to the baseline sorting.}
    \label{fig:classification_opticsbench_imagnet1k_ranking4}
\end{figure*}

Fig.~\ref{fig:classification_opticsbench_imagnet1k_ranking4} shows the accuracy for the best 50 models on ImageNet-1k ranked by the baseline \emph{defocus blur} corruption from~\cite{hendrycks_benchmarking_2019}. Each marker represents a model and each curve represents an image corruption. The accuracy for the baseline (red curve) corruption is compared to the OpticsBench corruptions. If the defocus blur corruption were the same for all models as the OpticsBench corruption, all curves would decrease strictly monotonically. However, there are several minor and major deviations from this including robust models (g, k, n, q and r) that perform comparatively better on OpticsBench compared to their baseline rank. A detailed discussion is given in the supplementary material~\ref{app:evaluation_and_tables}.

To further investigate the behavior of the different corruptions, we report the Kendall rank coefficient~\cite{kendall_treatment_1945} $\tau_b$  between the rankings of the baseline and a specific corruption in Fig.~\ref{fig:classification_opticsbench_imagnet1k_ranking4}. 
All $\tau_b$ values show an average weak correlation of \num{0.21} with the baseline sorting, with the individual $\tau_b$ being similar on all image corruptions. Note, that selecting the OpticsBench corruptions as baselines yields similar weak to moderate rank correlations to the other corruptions, with different outliers. The results are discussed in the supplementary material~\ref{app:rank_correlation_matrix}. 
They confirm that each corruption is different and that a single blur corruption is only a weak proxy for others.
Table~\ref{tab:rank_correlation_classification} lists the rank coefficients after excluding RobustBench models and the OpticsAugment trained model. As expected, this increases the rank correlation to the \emph{defocus blur} baseline for most of the corruptions. The correlation to the baseline for astigmatism and 
 defocus \& spherical rise to 0.34 and 0.36, for trefoil to 0.29. While the p-values are small for these corruptions, indicating that there is a ranking correlation (the null hypothesis has to be rejected), the high p-value for coma implies that there is no significant ranking.
\begin{table}[]
    \centering
    \caption{Rank correlation $\tau$ between corruptions and defocus blur baseline, when excluding the robust ResNet50 RobustBench models and the OpticsAugment trained model.}\label{tab:rank_correlation_classification}
    \begin{tabular}{@{}l c c l c c@{}}
     Corruption & $\tau$ & $p$ & Corruption & $\tau$ & $p$\\
     \hline
     Astigm.    & 0.34 & \num{5e-5} 
     & Def.\& Spher. & 0.36 & \num{2e-5} \\
     Coma & 0.06 & \num{0.5} &
     Trefoil & 0.29 & \num{5e-5}
    \end{tabular}
\end{table}

\begin{figure}
    \centering
    \includegraphics[trim=0cm 1cm 0cm  0cm,clip,width=\linewidth]{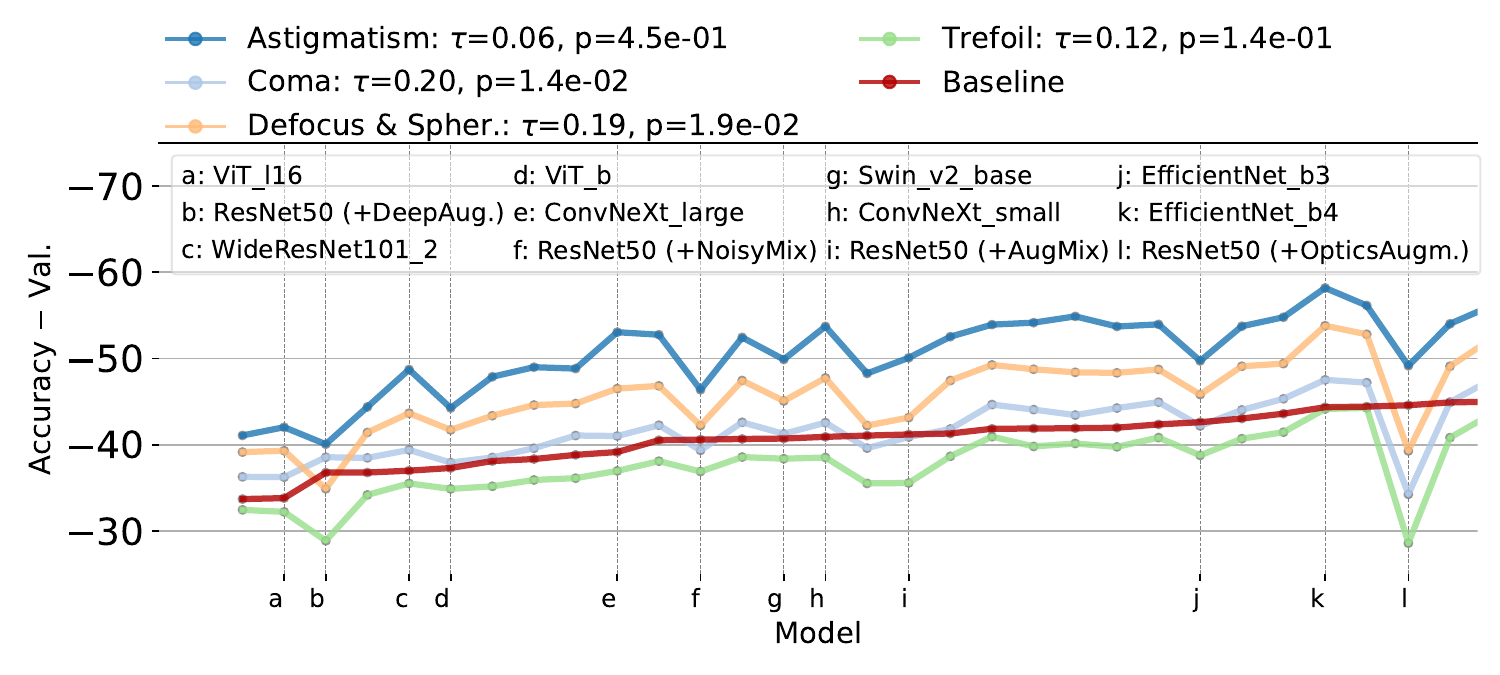}
    \caption{Ranking for the 30 most robust models on ImageNet-1k using the baseline \emph{defocus blur} (red) corruption from~\cite{hendrycks_benchmarking_2019} at severity 4 compared to the validation accuracy. Lower means, that the corruption accuracy is closer to the validation accuracy and thus more robust.}    \label{fig:classification_robustness_opticsbench_imagnet1k_ranking4}
\end{figure}

Fig.~\ref{fig:classification_robustness_opticsbench_imagnet1k_ranking4} shows the 30 most robust models to the baseline \emph{defocus blur}~\cite{hendrycks_benchmarking_2019} (red) on ImageNet-1k and severity 4. The colors refer to the different corruptions and the robustness is measured as difference to the validation accuracy, thus: a smaller absolute value shows a higher robustness. 
Fig.~\ref{fig:classification_robustness_opticsbench_imagnet1k_ranking4} shows that the large ViT models (a, a-1) are the most robust to the baseline and are also among the best at compensating for the OpticsBench corruptions. However, the ResNet50 models trained with DeepAugment (b), NoisyMix (f) 
perform significantly better on OpticsBench than the others at severity 4.
In summary, the DeepAugment, OpticsAugment and SIN\_IN ResNet50 models helped to gain robustness on OpticsBench, whereas the \eg ConvNeXt, EfficientNet\_b4 and Wide ResNet models lack robustness to all or some of the OpticsBench corruptions if trained without such data augmentation. The ViT and Swin transformer models were more robust compared to similarly ranked models. Although the models are sometimes robust to all other corruptions, there are exceptions. The baseline rank is generally not predictive of the behavior of the other corruptions, as shown by the many changes in the gradient direction of the specific corruptions.

\begin{figure}
    \centering
    \includegraphics[width=\linewidth]{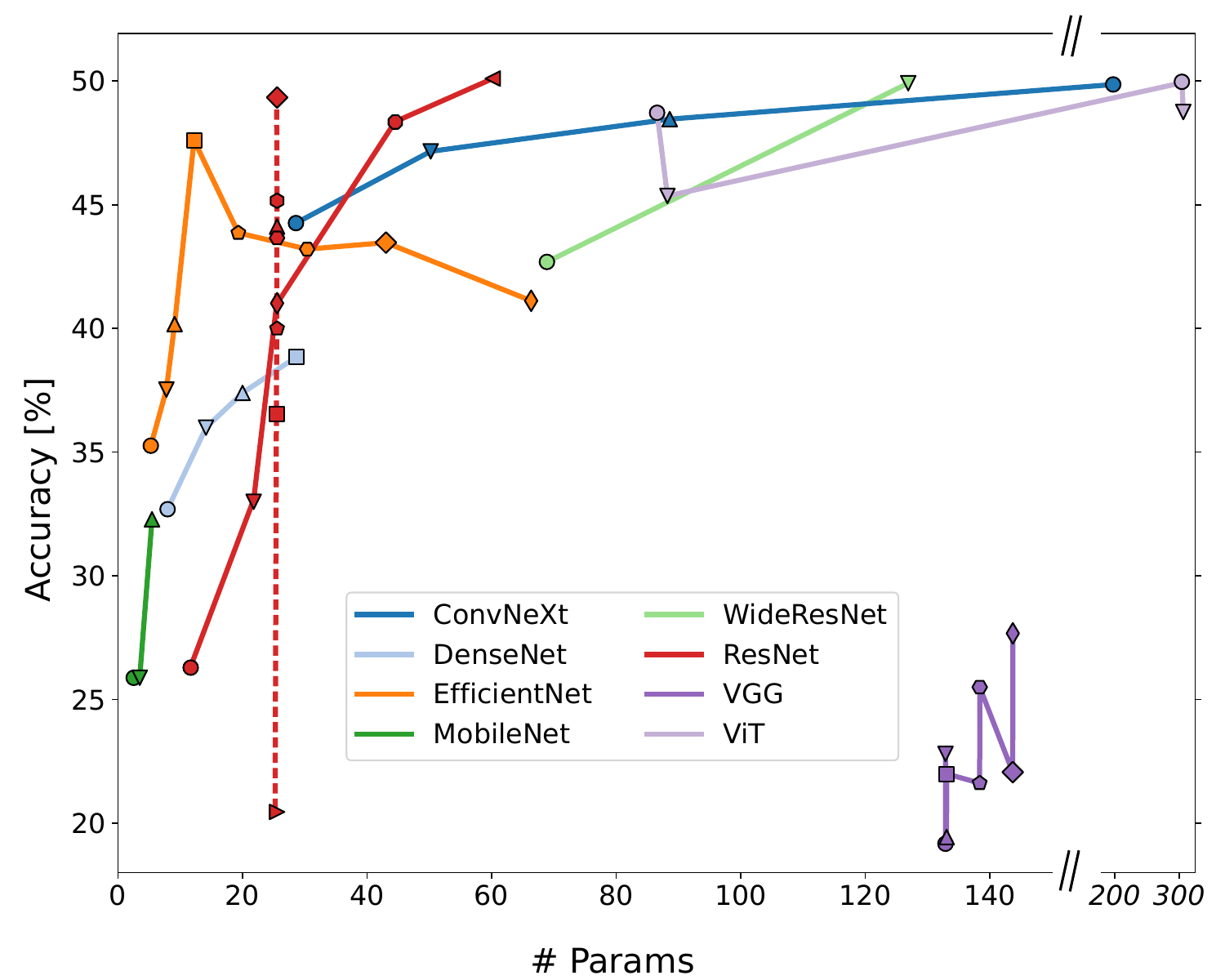}
    \caption{Average robustness on ImageNet-1k-OpticsBench in dependence of several model architectures and the number of model parameters. Models that belong to a single architectural family are connected with lines. The vertical dashed line connects differently trained ResNet50 models.}
    \label{fig:in1k_ob_num_param_vs_acc}
\end{figure}

Finally, we discuss the role of the model architecture for optical aberrations. Fig.~\ref{fig:in1k_ob_num_param_vs_acc} shows the average accuracy on ImageNet-1k-OpticsBench as a function of the number of model parameters, where models belonging to a single architectural family are connected with lines. Generally, larger models tend to be more robust, but the training impact is high. The red line shows several ResNet variants, where the dashed vertical line denotes robust and non-robust ResNet50 models. The low-performing model (red triangle) is adversarially trained and is not robust against optical aberrations. Also, different stylized ImageNet training~\cite{geirhos2018imagenet} variants result in lower accuracy compared to the standard ResNet50 model (red diamond). The high-performing ResNet50 models are trained with DeepAugment (red diamond) or OpticsAugment (red hexagon, below), which results in accuracies that are competitive to highly  parametrized models such as ConvNeXt large (blue circle).

\subsection{Experiments on Flowers-102 and Stanford Cars}
To further investigate the behavior of image classification models, we also generate OpticsBench on the fine-grained image classification datasets Flowers-102~\cite{nilsbackAutomatedFlowerClassification2008} and Stanford Cars~\cite{krause3DObjectRepresentations2013}. 
On both fine-grained datasets we first fine-tune strong ImageNet-1k pre-trained models from diverse architectures and then evaluate these models on the clean test dataset and on OpticsBench image corruptions. We report the results in Table~\ref{tab:fine_grained_datasets} and provide details on the datasets and model training in the supplementary material~\ref{app:classification_models_fine_grained_datasets}. 

Table~\ref{tab:fine_grained_datasets} shows the accuracy for the different models evaluated on the clean datasets and on OpticsBench. While the decrease in accuracy on Flowers-102 (top) for OpticsBench by 34 points is comparable to ImageNet-1k, on Stanford Cars (bottom), the performance degradation is more pronounced with a decrease by 53 points. Differing training settings between the datasets, such as layer-wise learning rate decay, contribute to the pronounced difference in robustness. However, we expect that Stanford Cars is also more challenging compared to Flowers-102 with more classes and higher intra-class variability, which challenges the models when confronted to optical aberrations. 
\begin{table}[h]
    \centering
    \caption{Fine-grained image classification on Flowers-102~\cite{nilsbackAutomatedFlowerClassification2008} (top) and Stanford Cars~\cite{krause3DObjectRepresentations2013} (bottom). We report the accuracy of five different fine-tuned ImageNet pre-trained models on the corresponding test set and OpticsBench.}
    \label{tab:fine_grained_datasets}
    \scriptsize 
    \begin{tabular}{c}
    \footnotesize{Flowers-102} \\ 
    \begin{tabular}{@{}l c ccccc|c@{}}
    DNN & Clean $\uparrow$ & \multicolumn{5}{c}{Severity} & \\
    && 1 & 2 & 3 & 4 & 5 & $\Sigma$\\
    \hline 
    ConvNeXt (small) & 96.9 & 89.1 & 81.7 & 66.9 & 55.5 & 47.0 & 68.0 \\
    DenseNet201 & 96.8 & 86.8 & 76.6 & 54.8 & 38.8 & 30.1 & 57.4 \\
    ResNet50 & 93.7 & 77.4 & 64.3 & 41.8 & 29.4 & 22.3 & 47.0 \\
    Swin\_v2 (small) & 92.9 & 82.4 & 75.1 & 63.3 & 53.5 & 46.0 & 64.1 \\
    ViT (base) & 97.3 & 88.0 & 83.4 & 72.9 & 62.4 & 53.6 & 72.1 \\
    \hline 
    $\Sigma$ & 95.5 & 84.7 & 76.2 & 59.9 & 47.9 & 39.8 & 61.7
    \vspace{.05\linewidth}
    \end{tabular}    
    \\
    \footnotesize{Stanford Cars}
    \\
    \begin{tabular}{@{}l@{}c ccccc|c@{}} 
    DNN & Clean $\uparrow$ & \multicolumn{5}{c}{Severity} &  \\
    && 1 & 2 & 3 & 4 & 5 & $\Sigma$ \\
    \hline 
    ConvNeXt (small) & 90.1 & 72.2 & 57.5 & 33.2 & 19.5 & 13.5 & 39.2 \\
    DenseNet201 & 88.8 & 71.8 & 58.2 & 33.5 & 18.5 & 12.1 & 38.8 \\
    ResNet50 & 85.2 & 57.5 & 40.4 & 19.0 & 10.8 & 7.6 & 27.1 \\
    Swin\_v2 (small) & 89.6 & 69.6 & 55.1 & 31.1 & 17.8 & 11.5 & 37.0 \\
    ViT\_b & 86.5 & 65.5 & 50.4 & 27.0 & 15.4 & 10.6 & 33.8 \\
    \hline 
    $\Sigma$ & 88.0 & 67.3 & 52.3 & 28.8 & 16.4 & 11.1 & 35.2
    \vspace{.025\linewidth}
    \end{tabular}
    \end{tabular}
\end{table}

\subsection{Experiments on MSCOCO} 
\label{sec:opticsbench_detection}
To measure the robustness of DNNs for object detection to optical aberrations, we extend the implementation from OpticsBench~\cite{mueller_opticsbench2023}. We apply the evaluations to models for the seminal computer vision task of object detection, notably on the datasets MSCOCO \cite{fleet_microsoft_2014} and NuImages~\cite{caesar_nuscenes_2020}. While the former is a commonly used dataset to evaluate general purpose object detection, the latter dataset has a strong focus on street scenes and driving scenarios. 
We measure the mean Average Precision (mAP) \deleted{metric} using COCO evaluation tools and \added{perform a } separate analysis of classification (Cls) and localization (Loc) errors following the TIDE evaluation of Bolya et al.~\cite{bolya_tide_2020}. For simplicity, we will not discuss the other error types listed in~\cite{bolya_tide_2020}. The errors refer to oracles that predict how much improvement in mAP50 would be gained by fixing a particular error type individually. 
An overview on the considered detection models is given in Tab.~\ref{tab:mscoco_ob_dnn_overview} together with the average mAP, Cls and Loc errors on the unmodified (clean) validation dataset and on MSCOCO-OpticsBench (OB). The remaining error metrics are given in the supplementary material in Table~\ref{tab:app_tide_metrics_overview}. We evaluate two-stage and single-stage architectures as well as transformer and CNN based architectures. 
\begin{table}[h]
    \centering
    \scriptsize 
    \caption{Evaluated DNNs on MSCOCO~\cite{fleet_microsoft_2014} validation dataset (Clean) and OpticsBench (OB). Here, we report the mAP values for the evaluated DNNs together with Cls and Loc errors\deleted{ with the evaluation method from Bolya et al.}~\cite{bolya_tide_2020}.}  
    \begin{tabular}{@{}l c c c c@{}}
    \multirow{2}{*}{DNN / Backbone} & \multicolumn{2}{c}{mAP \added{\ensuremath{\uparrow}}} & \multicolumn{2}{c}{Cls / Loc \added{\ensuremath{\downarrow}}} \\
     & Clean & OB & Clean & OB \\
    \hline
    Cascade R-CNN~\cite{cai_cascade_2021} & 40.4 & 18.1 & 4.0 / 6.9 & 7.3 / 5.0 \\ 
    Cascade Mask R-CNN / ConvNeXt & 51.9 & 30.1 & 2.3 / 6.2 & 5.2 / 6.1 \\
    Deformable DETR~\cite{zhu_deformable_2020} & 47.0 & 22.0  & 2.3 / 6.3 & 5.6 / \textbf{4.8} \\
    DINO~\cite{zhang_dino_2022} / Swin\_l & \textbf{58.4} & \textbf{37.4} & \textbf{1.2} / \textbf{4.8} & \textbf{2.8} / 6.3 \\
    Faster R-CNN~\cite{lin_feature_2017} & 37.4 & 16.6 & 3.8 / 7.0 & 7.3 / 5.1 \\
    Mask R-CNN~\cite{he_mask_2017} / Swin\_s & 48.2 & 27.1 & 2.1 / 6.7 & 4.6 / 6.3 \\
    RetinaNet~\cite{lin_focal_2017} / Eff.Net\_b3 & 40.5 & 23.7 & 2.7 / 7.0 & 4.5 / 5.5 \\
    YOLO-X\_x~\cite{ge_yolox_2021} / DarkNet & 50.6 & 30.9 & 3.2 / 6.4 & 5.6 / 5.8 \\
    \hline
    $\Sigma$ & 46.8 & 25.7 & 2.7 / 6.4 & 5.3 / 5.6
\end{tabular}
    \label{tab:mscoco_ob_dnn_overview}
\end{table}

\begin{figure}[htb]
    \centering
    \includegraphics[width=\linewidth]{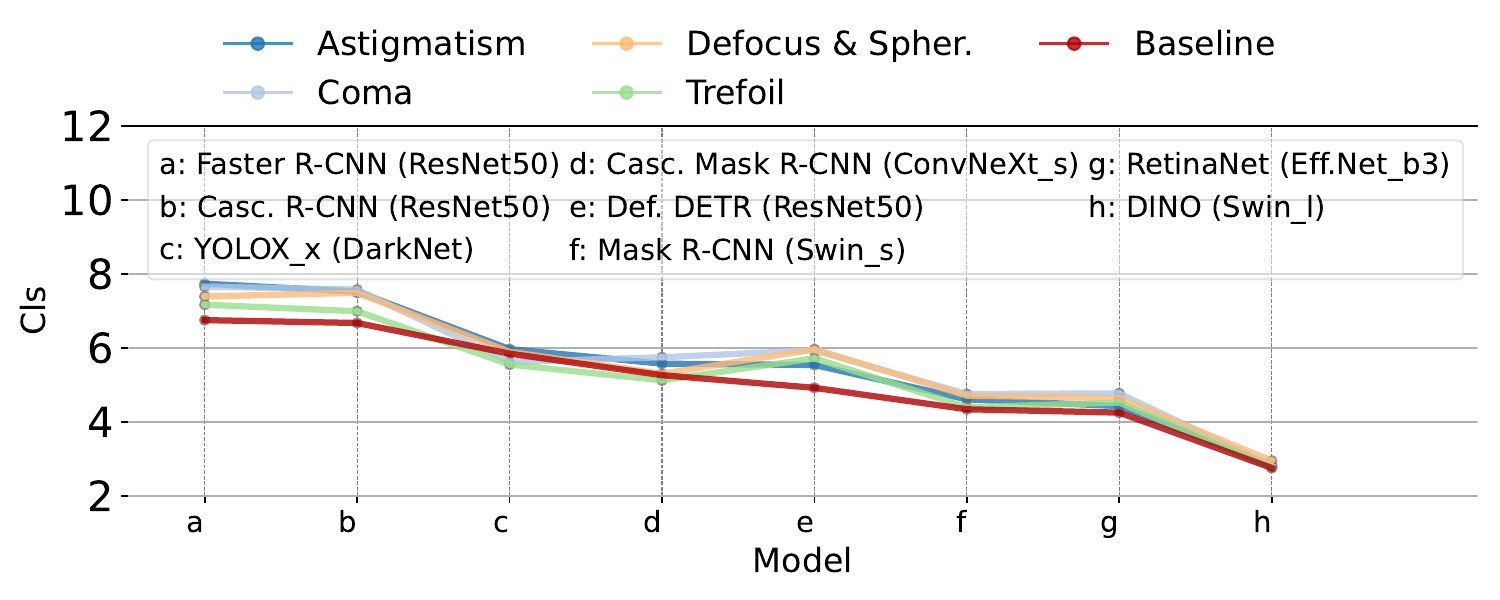}
    \\
    {\includegraphics[width=\linewidth]{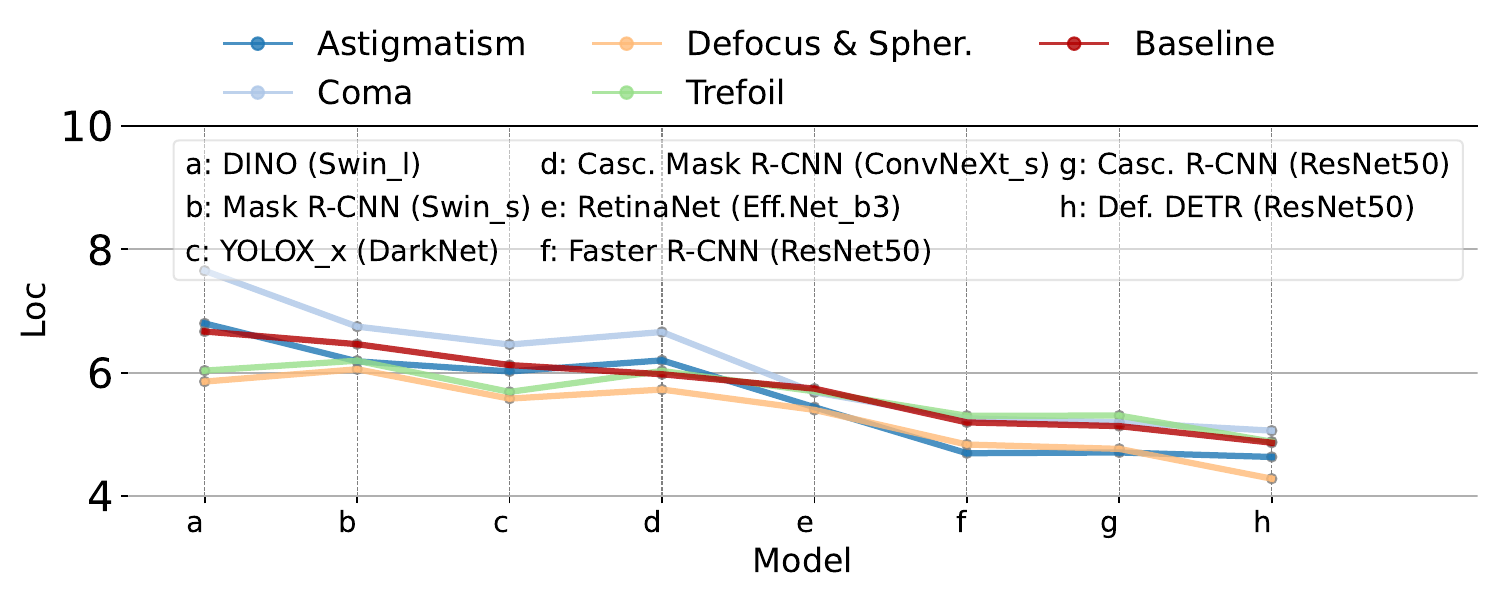}}
    \caption{Cls (top) and Loc (bottom) errors from TIDE~\cite{bolya_tide_2020} for the different OpticsBench~\cite{mueller_opticsbench2023} corruptions and the baseline for severity 3. Lower is better.}  
    \label{fig:cls_and_loc}
\end{figure}

The DNNs have an average mAP of 46.8 on the clean dataset and an average mAP of 25.7 on OpticsBench. This is a reduction to \SI{54.3}{\percent} of the clean mAP, which is very close to the classification results where the accuracy is reduced by an average of \SI{55.7}{\percent}. 
While the Cls errors double from an average of 2.7 to 5.3, the Loc errors reduce for most of the models except the DINO model from an average 6.4 to 5.6. This means that, there would be a larger robustness gain on OpticsBench, when fixing the Cls errors than the Loc errors. All six main errors are further discussed in the supplementary material~\ref{app:mscoco_rankings_a}. 
Here we focus on the evaluation of class (Cls) and localisation (Loc) errors for different DNNs and exemplarily show the results for severity 3 in Fig.~\ref{fig:cls_and_loc}. 
We sort by Cls and Loc errors (lower is better), with the baseline at severity 3. 

In Fig.~\ref{fig:cls_and_loc} (top) for some models the Cls error depends on the image corruption, while for others it is quite similar. YOLO (c), Mask R-CNN (f) and in particular DINO (h) have similar Cls errors for all image corruptions. For the remaining models the Cls errors are more different for the various image corruptions. The Cls errors of Faster R-CNN (a) and Cascade R-CNN (b) are highest for astigmatism and coma (light blue), while the baseline (red) has the lowest Cls error. 
In Fig.~\ref{fig:cls_and_loc} (bottom) the coma corruption (light blue) has the highest Loc errors for half of the models (a-d), while the distance to the baseline becomes negligible for lower ranked models (e-h). A detailed discussion is given in the supplementary material~\ref{app:mscoco_rankings_a}.  
\newcommand{\plotSzCLS}{0.98\linewidth}

\section{Improving model robustness to primary aberrations with \emph{OpticsAugment} training}
\label{sec:opticsaugment}
Since the above results show a clear drop in performance due to the different image corruptions, and motivated by the general benefit of training with data augmentation, we proposed a data augmentation method with kernels from the OpticsBench kernel generation, which supplements existing 
methods.~\cite{mueller_opticsbench2023} 

OpticsAugment~\cite{mueller_opticsbench2023} follows a similar approach as AugMix~\cite{hendrycks_augmix_2020}.
During dataloading in the training process each image is convolved with an individual RGB-kernel from the kernel stack containing \eg 40 kernels for the different primary aberrations and severities. The random selection of the kernel follows a uniform distribution. 
Additionally, each resulting image is a weighted combination of the original image and the blurred image, following a beta-distribution as in~\cite{hendrycks_augmix_2020}, which controls the amount of augmentation. Thus, the total number of images per epoch stays the same. 
The exact combination varies each epoch, which facilitates generalization to the blur types. Since the processing is accelerated by GPU parallelization, the blurred images do not need to be computed in advance, but can be computed on-the-fly with comparable overhead to other methods such as~\cite{kar_3d_2022} and a kernel size of 25x25x3. For this, OpticsAugment is implemented using parallelization on full image \added{mini-}batches, which results in strong acceleration. For an image \added{mini-}batch of 128 images an overhead of about 250ms is generated. Using the PyTorch dataloader takes much longer since the number of workers is typically about 4-8, while the batch size is typically between 32 and 128. The implementation relies heavily on GPU acceleration using Cuda and PyTorch. The processed image \added{mini-}batch is then normalized to the dataset specific mean and standard deviation. This is crucial for convergence. 

Since AugMix~\cite{hendrycks_augmix_2020} provides low-cost data augmentation and promising results but no diverse optical blur kernel augmentation, we also try cascading AugMix and OpticsAugment. Since both methods can feed the training algorithm with highly corrupted images, direct chaining leads to non-convergence and low accuracy. 
Therefore, we model the probability of augmentation with a flat Dirichlet distribution in four dimensions: The first two variables are the probabilities for AugMix and OpticsAugment. The other variables are auxiliary variables and ensure that the overall probability of each augmentation remains uniformly distributed. The output of OpticsAugment is normalized.

\subsection{Experiments on ImageNet-100} 
\paragraph{OpticsBench}
\label{sec:experiments_imagenet100}
We train five different DNNs on ImageNet-100 using OpticsAugment. In addition, we train baseline models on ImageNet-100 with the same hyperparameter settings, but without \deleted{the} data augmentation. These DNNs are then compared to each other on OpticsBench and 2D Common Corruptions~\cite{hendrycks_benchmarking_2019} applied to ImageNet-100. We select five different architectures:  ResNeXt50, ResNet101, EfficientNet\_b0, MobileNet\_v3\_large and DenseNet161. 

The train split is divided with a fixed seed into $5\%$ validation images and $95\%$ training images to ensure that the benchmark data contains only unseen data. On top of the trained DNNs, all models are also trained with the same settings, but include OpticsAugment with a severity of 3 during training and the amount of augmentation is uniformly distributed, so $\alpha=1.0$.
The hyperparameter settings follow the standard simple training recipes as reported in the PyTorch 1.x references~\cite{noauthor_visionreferencesclassification_nodate} using cross-entropy loss, stochastic gradient descent and learning rate scheduling\added{,} but no additional data augmentation such as CutMix~\cite{yun_cutmix_2019} or AutoAugment~\cite{cubuk_autoaugment_2019}. The DNNs are trained with the same batch size and number of epochs for clean, AugMix and OpticsAugment training. Fine-tuning the hyperparameters and more complex training recipes using additional standard data augmentation can further improve the observed benefit. 
\begin{table}
    \caption{Performance \emph{gain} with OpticsAugment on all ImageNet-100 OpticsBench corruptions. Average difference in accuracy across all corruptions in \%-points for each severity. Details are given in \added{the} supplementary~\ref{app:optics_augment}, tables~\ref{tab:tab:imagenet100_corruptions_DenseNet_revisited}-\ref{tab:tab:imagenet100_corruptions_ResNeXt50_revisited}.}
    \label{tab:optics_augment_opticsbench}
    \centering
    \footnotesize
    \begin{tabular}{llllll}
        DNN & 1 & 2 & 3 & 4 & 5\\
        \hline 
        DenseNet161 & 14.77 & 21.96 &  27.26  & 20.98 & 13.84\\
        ResNeXt50 & 17.40 &  24.49 &  29.56 & 22.32 &  14.76\\
        ResNet101 & 9.97 & 16.24 & 21.15 & 17.39 & 12.07\\
        MobileNet & 8.12 & 12.73 & 13.80 & 10.09 & 7.27\\
        EfficientNet  & 8.45 & 12.60 & 12.90 & 9.43 & 7.35\\
    \end{tabular}
\end{table}

\begin{figure}[h]
    \centering
    \begin{tabular}{@{}c@{}}
    \small ResNeXt50 \\
{\centering\includegraphics[trim=0cm 0.5cm 0cm 0cm,clip,width=\linewidth]{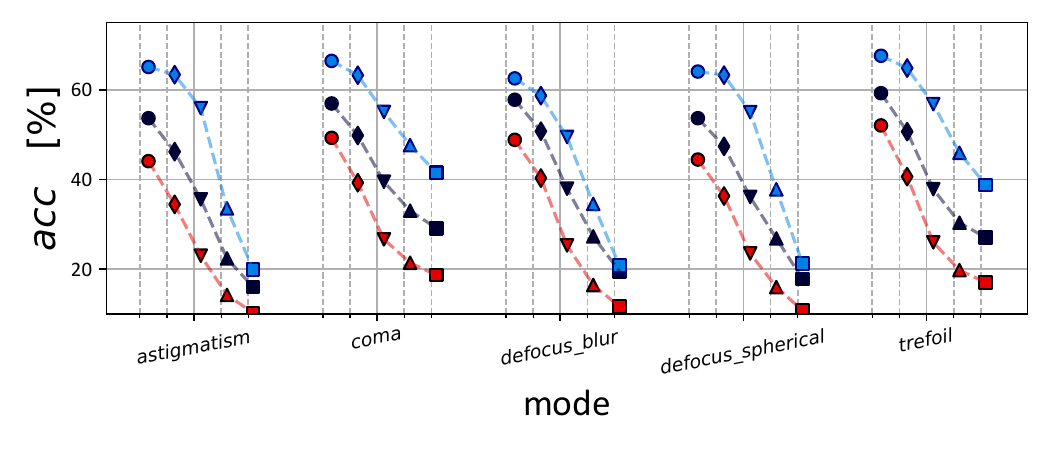}}\\ 
\\ \small DenseNet161 \\ 
   {\centering\includegraphics[trim=0cm 0.5cm 0cm 0cm,clip,width=\linewidth]{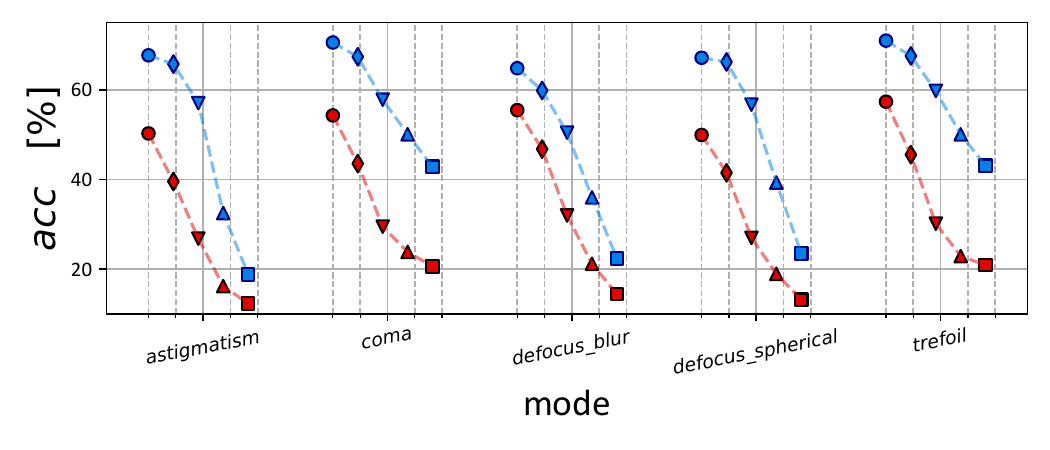}}\\
\end{tabular}
\caption{Accuracy on OpticsBench-ImageNet-100 for  DNNs with (blue) and without (red) OpticsAugment training. The x-axis shows for each of the five corruptions five different severities (increasing from left to right). ResNeXt50 (top) and DenseNet161 (bottom). We also train a ResNeXt50 with Augmix~\cite{hendrycks_augmix_2020} (black). The accuracy increases by on average $29.6$\% points for ResNeXt50 compared to the clean trained DNN (red) \textbf{with OpticsAugment (blue)} and severity $3$ (down-pointing triangles). The performance gain for defocus blur~\cite{hendrycks_benchmarking_2019} is the lowest for all severities.}
    \label{fig:optics_augment}
\end{figure}

Tab.~\ref{tab:optics_augment_opticsbench} gives an overview of the improvement on ImageNet-100 OpticsBench with OpticsAugment. The smaller DNNs (MobileNet, EfficientNet\_b0) show significantly lower accuracy improvements, while ResNeXt50 gains up to $29.6\%$ points with OpticsAugment.
Fig.~\ref{fig:optics_augment} compares the accuracies with/without OpticsAugment during training for each corruption. The ResNeXt50 in Fig.~\ref{fig:optics_augment} (top) improves on average by 21.7\% points with OpticsAugment (blue) compared to the default model (red) and by 11.6\% compared to AugMix (black). Robustness to coma is significantly improved, especially for higher severities.
The results for DenseNet161 in Fig.~\ref{fig:optics_augment} (bottom) are similar with an average improvement of 19.8\% points. The severity 3 accuracies in Fig.~\ref{fig:optics_augment} for the standard DenseNet161 model are comparable, while OpticsAugment is significantly more robust to OpticsBench corruptions. For example, trefoil (last data point) is handled overly well while the performance gain for defocus blur is significantly lower. This seems to hold across different severities and DNNs: The corresponding disk-shaped kernel~\cite{hendrycks_benchmarking_2019} was not present during training, suggesting that diverse blur kernels need to be considered during training. 

\begin{table}[h]\centering
\caption{Accuracies w/wo OpticsAugment evaluated on ImageNet-100 OpticsBench. Average over all corruptions, last column shows the average over fields and corruptions.}
\footnotesize
\begin{tabular}{@{}llllll | l@{}}
Model & 1 & 2 & 3 & 4 & 5 & $\Sigma$ \\ 
\hline 
DenseNet \textbf{(ours)} & \textbf{68.22} & \textbf{65.33} & \textbf{56.33} & \textbf{41.60} & \textbf{30.13} & \textbf{52.32} \\ 
DenseNet & 53.45 & 43.37 & 29.07 & 20.62 & 16.30 & 32.56\\
EfficientNet \textbf{(ours)} & \textbf{61.00} & \textbf{55.34} & \textbf{42.14} & \textbf{30.27} & \textbf{23.35} & \textbf{42.42} \\
EfficientNet & 52.55 & 42.74 & 29.24 & 20.84 & 16.00 & 32.27\\
MobileNet \textbf{(ours)} & \textbf{57.59} & \textbf{52.30} & \textbf{38.58} & \textbf{27.51} & \textbf{20.54} & \textbf{39.04} \\ 
MobileNet & 49.47 & 39.57 & 24.78 & 17.42 & 13.27 & 28.90\\
ResNet101 \textbf{(ours)} & \textbf{69.90} & \textbf{67.68} & \textbf{61.36} & \textbf{49.04} & \textbf{37.80} & \textbf{57.16} \\ 
ResNet101 & 59.92 & 51.44 & 40.21 & 31.65 & 25.73 & 41.79\\
ResNeXt50 \textbf{(ours)} & \textbf{65.14} & \textbf{62.68} & \textbf{54.44} & \textbf{39.90} & \textbf{28.45} & \textbf{50.12} \\ 
ResNeXt50 & 47.74 & 38.19 & 24.88 & 17.58 & 13.69 & 28.42 \\\end{tabular}
\label{tab:tab:imagenet100_avg_absolute}\end{table}

We also provide a ResNet50+OpticsAugment checkpoint for ImageNet-1k with a simple training scheme using 90 epochs. Using a more advanced and longer training recipe may significantly help to gain more robust features and improve the results on ImageNet and ImageNet-OpticsBench. We have added results for stronger models on ImageNet-100 to the supplementary material~\ref{app:optics_augment}. 
The results on ImageNet-1k-OpticsBench are presented in Table~\ref{tab:imagenet1k_opticsbench_trained}. We compare the simply trained ResNet50+OpticsAugment for 90 epochs to the  baseline ResNet50 model from PyTorch, as well as to the RobustBench checkpoint for ResNet50+AugMix. Still, our model is the most robust model across severities. Compared to the validation accuracy, our ResNet50+OpticsAugment model reduces on average \SI{10.4}{\percent} less than the baseline and \SI{4.2}{\percent} compared to the AugMix trained ResNet50 from RobustBench. 
\begin{table}[]
    \centering
        \caption{ImageNet-1k-OpticsBench for all severities for the  ResNet50 model from PyTorch with basic data augmentation, the strong RobustBench  AugMix~\cite{croce2021robustbench} checkpoint~\cite{croce2021robustbench} and a simple training scheme using  OpticsAugment~\cite{mueller_opticsbench2023} training. The last column compares the loss in accuracy compared to the validation accuracy (Lower is better).}
    \label{tab:imagenet1k_opticsbench_trained}
    \footnotesize{
    \begin{tabular}{@{}l@{} llllll|l|l@{}}
    Model & Val & 1 & 2 & 3 & 4 & 5 & $\Sigma \uparrow$ & Rel. $\downarrow$\\ 
    \hline 
    ResNet50 & 76.1 & 54.6 & 45.3 & 29.8 & 20.6 & 15.6 & 33.2 & 42.9 \\   
    +AugMix & \textbf{77.5} & 60.3 & 53.6 & 40.1 & 28.7 & 21.3 & 40.8 & 36.7 \\
    +OpticsAug. \ & 74.2 & 59.5 & 54.8 & 42.3 & 29.5 & 22.1 & \textbf{41.7} & \textbf{32.5}
     \end{tabular}
    }
\end{table}

\begin{figure*}[ht]
    \centering 
    \includegraphics[trim=0cm .9cm 0cm 0cm,clip,width=.9\linewidth]{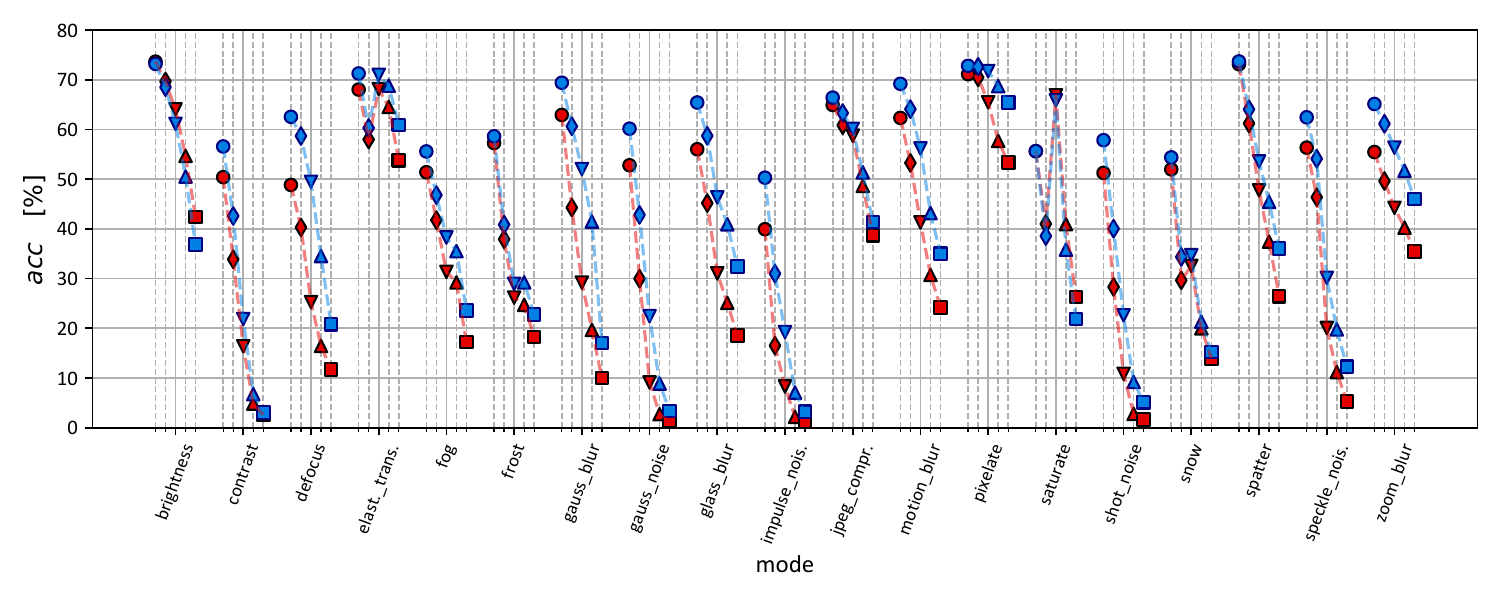}
    \caption{Accuracy for ResNeXt50 evaluated on ImageNet-100-C 2D common corruptions w/wo OpticsAugment training and all severities 1-5 (circle, diamond, triangles and square markers) at each corruption. \textbf{OpticsAugment (blue)} accuracy compared to the conventionally trained DNN (red).}
    \label{fig:common2d_modes_resnext50}
\end{figure*}

\paragraph{2D common corruptions}
As an example, we discuss here the results for ResNeXt50 with and without OpticsAugment in Fig.~\ref{fig:common2d_modes_resnext50}. 
Fig.~\ref{fig:common2d_modes_resnext50} compares the accuracies for ResNeXt50 solely trained on ImageNet-100 (red) and augmented with OpticsAugment (blue) respectively.
For most of the corruptions the augmentation is beneficial, including blur, weather corruptions (fog, frost) and pixelation. However, some corruptions are not compensated by OpticsAugment, \eg~JPEG compression and contrast. Noise robustness may be improved by chaining OpticsAugment with AugMix.

In addition, cascading AugMix and OpticsAugment (\ie executing both types of augmentations one after another) further improves robustness  on 2D Common corruptions (especially to Gaussian and speckle noise, cf. Table~\ref{tab:tab:imagenet100c_corruptions_EfficientNet}), but reduces the accuracy on OpticsBench. Table~\ref{tab:cascading_overview} shows the average improvement for a cascaded application of AugMix \& OpticsAugment and the different severities. The accuracy of EfficientNet increases by an additional $3.4\%$ points on average for 2D common corruptions. 
\begin{table}[t]
\caption{Cascading AugMix (AM) and OpticsAugment (OA). Additional average improvement \deleted{for an evaluation} on ImageNet-100\deleted{-c 2D} common corruptions~\cite{hendrycks_benchmarking_2019}.}
\centering\begin{tabular}{llllll}
\footnotesize{DNN} & \footnotesize{$1$} & \footnotesize{$2$} & \footnotesize{$3$} & \footnotesize{$4$} & \footnotesize{$5$} 
\\\hline\footnotesize{EfficientNet (AM + OA)} & \footnotesize{+2.79} & \footnotesize{+3.89} & \footnotesize{+4.34} & \footnotesize{+3.43} & \footnotesize{+2.61}\\
\footnotesize{MobileNet (AM + OA)} & \footnotesize{+1.59} & \footnotesize{+2.27} & \footnotesize{+1.75} & \footnotesize{+0.57} & \footnotesize{-0.32}
\end{tabular}
\label{tab:cascading_overview}
\end{table}

\paragraph{Adversarial robustness}
Table~\ref{tab:app:adversarial_robustness} shows results on adversarial robustness for different ImageNet-100 trained DNNs. Our models use the OpticsAugment training scheme and are compared to a conventionally trained DNN on the same dataset. To allow for evaluation on ImageNet's validation dataset, the train set is split into a validation and train split. 
To reduce the computational resources needed for the computation, 1000 validation images are randomly selected and saved as a test dataset for adversarial robustness. The attacks have been lowered to $\ell_2$ and $\epsilon=4/255$ to avoid exclusively successful attacks. The batch size is $32$ and we use $5$ restarts on the $1000$ ImageNet-100 images.
However, there is no clear trend with this setting, the overall robustness to the attacks is low, but on average OpticsAugment does not lower adversarial robustness compared to a conventionally trained DNN. The evaluation for each DNN takes several hours on a NVIDIA GeForce 3080Ti 12GB VRAM GPU. 
\begin{table}[h]
    \centering
    \caption{Adversarial robustness in \% to adversarial attacks from AutoAttack\deleted{using }\added{ (}APGD-CE, APGD-DLR\added{)}~\cite{croce_reliable_2020}, $\ell_2$ and $\epsilon=4/255$.}   
\label{tab:app:adversarial_robustness} \footnotesize
    \begin{tabular}{l l l l}
       DNN  & Robust Acc & APGD-CE & APGD-DLR \\
       \hline
       DenseNet  & 5.2 & 13.9 & 5.7  \\ 
       DenseNet (ours)  & 6.4 & 14.5 & 6.4   \\ 
       EfficientNet  & 2.1 & 8.8 & 2.4 \\ 
       EfficientNet (ours)  & 1.7 & 8.4 & 1.9 \\ 
       MobileNet  & 1.2 & 6.2 & 1.6 \\ 
       MobileNet (ours)  & 1.8 & 7.6 & 2.2 \\ 
       ResNeXt50  & 1.2 & 11.3 & 1.6 \\ 
       ResNeXt50 (ours)  & 3.1 & 8.9 & 3.6  \\ 
    \end{tabular}
\end{table}

\subsection{Experiments on MSCOCO - OpticsAugment} 
We have trained a Faster R-CNN with ResNet50 backbone with Feature Pyramid Network  (FPN)~\cite{lin_feature_2017} for 26 epochs on MS COCO~\cite{fleet_microsoft_2014} with and without OpticsAugment~\cite{mueller_opticsbench2023} to improve robustness to primary aberrations. 
The last three backbone layers of the non-robust pre-trained ResNet50 are not frozen. We also train with robust backbones (AugMix, OpticsAugment) and  fixed-feature transfer~\cite{yamada_does_2022,vasconcelos_proper_2022}. Compared to the baseline model, OpticsAugment improved by an average of +7.7\% for mAP across all severities and corruptions. Details can be found in the supplementary material~\ref{app:training}.

\section{Testing robustness to lens aberrations - \textit{LensCorruptions}}
The previous results investigated base vectors from the coefficient space in Fig.~\ref{fig:coefficient_space}. Following this perspective, we now study realistic lens kernels which are distributed in this coefficient space.
As shown in Figs.~\ref{fig:coefficient_space}  
 and~\ref{fig:lens_quality} the lens quality varies with each individual lens and field position, leading to diverse blur corruptions. Here, we investigate the impact of the various lens blur corruptions to classification and object detection using our curated dataset with lens blur kernels and apply these kernels to images using the 
 method described in 
Section~\ref{sec:lens_corruptions_construction}.
We evaluate several diverse pre-trained models for image classification and object detection respectively. 

\subsection{Experiments on ImageNet-100}
\label{subsec:classification_lens_corruptions}
For image classification we conduct experiments on the ImageNet-100~\cite{tian_contrastive_2020} validation set to which we apply our 100 lens corruptions each evaluated at five field positions. ImageNet-100 is a smaller dataset of 100 randomly selected classes from ImageNet-1k with a validation set of \num{5000} images.~\cite{tian_contrastive_2020} 
We analyze a diverse set of 12 pre-trained DNNs including CNNs and transformer models at different scales on the corrupted and unmodified dataset.  

Tab.~\ref{tab:classification_lens_corruptions_overview} shows the DNNs used and their corresponding accuracies (top-1) achieved on the ImageNet-100 validation dataset together with their average performance on LensCorruptions and OpticsBench~\cite{mueller_opticsbench2023}. 
The models achieve an average accuracy of \SI{81.0}{\percent} on the clean dataset and \SI{67.8}{\percent} on the corrupted dataset. 
The average effect of the lenses can be seen as a decrease by \SI{13.2}{\percent} accuracy on ImageNet-100 across field positions and corruptions. The models have an average accuracy of \SI{48.3}{\percent} on OpticsBench.
ConvNeXt\_large performs best and is also most robust with an average reduction by \ensuremath{\SI{9.3}{\percent}} compared to the validation accuracy. Besides ConvNeXt\_small, EfficientNet\_b3 has high robustness.
\begin{table}[t]
  \centering
  \caption{Accuracies on ImageNet-100 for different pre-trained classification models on Validation (clean), averaged lens corruption (LC) and OpticsBench (OB).}
\begin{tabular}{l c c c}
    DNN & Clean & LC & OB \\
    \hline     
    ConvNeXt\_large ~\cite{liu2022convnet} & \textit{83.10} & \textbf{73.84} & 54.78 \\
        ConvNeXt\_small~\cite{liu2022convnet} & 82.16 & 71.86 & 51.83 \\
    DenseNet\_169~\cite{huang_densely_2017} & 78.70 & 63.13 & 41.31 \\
    DenseNet\_201~\cite{huang_densely_2017} & 78.64 & 63.76 & 42.95 \\    
    EfficientNet\_b0~\cite{tan_efficientnet_2019} & 79.28 & 62.21 & 41.74 \\
    EfficientNet\_b3~\cite{tan_efficientnet_2019} & \textbf{83.26} & \textit{72.47} & 52.71 \\ 
    MobileNet\_v3\_l~\cite{howard_mobilenets_2017} & 77.26 & 61.55 & 36.19 \\
    ResNet50~\cite{he_deep_2016} & 81.38 & 68.65 & 46.42 \\    
    ResNeXt50~\cite{xie_aggregated_2017} & 82.42 & 69.87 & 46.92 \\
    Swin\_v2\_base~\cite{liu_swin_2022}& 82.86 & 71.81 & \textit{54.93} \\
    ViT\_base~\cite{ranftl_vision_2021} & 80.98 & 67.59 & 53.04 \\
    ViT\_large~\cite{ranftl_vision_2021} & 81.88 & 69.36 & \textbf{56.13} \\
    \hline
    $\Sigma$ & 81.00 & 67.81 & 48.25 \\
\end{tabular}     \label{tab:classification_lens_corruptions_overview}
\end{table}

\begin{figure}
    \centering
        \includegraphics[width=0.9\linewidth]{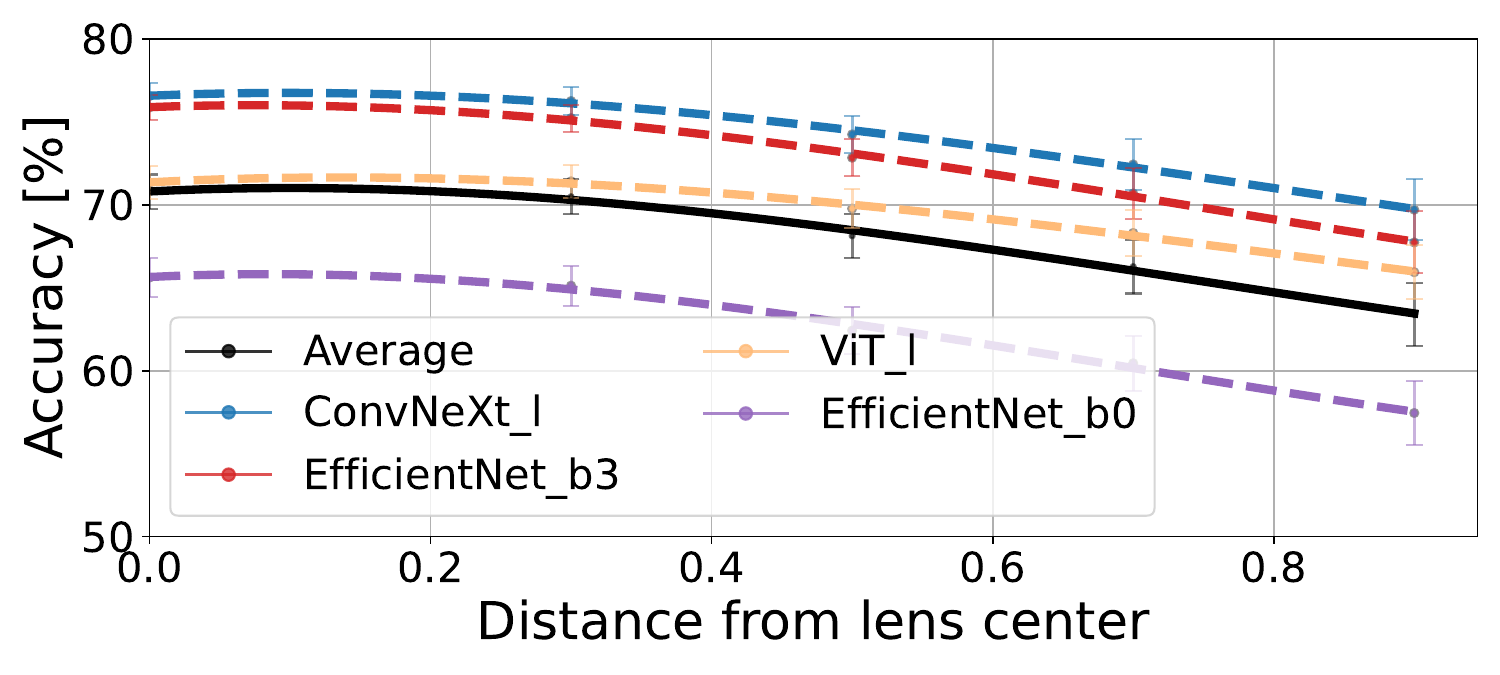}\\
        \includegraphics[width=0.9\linewidth]{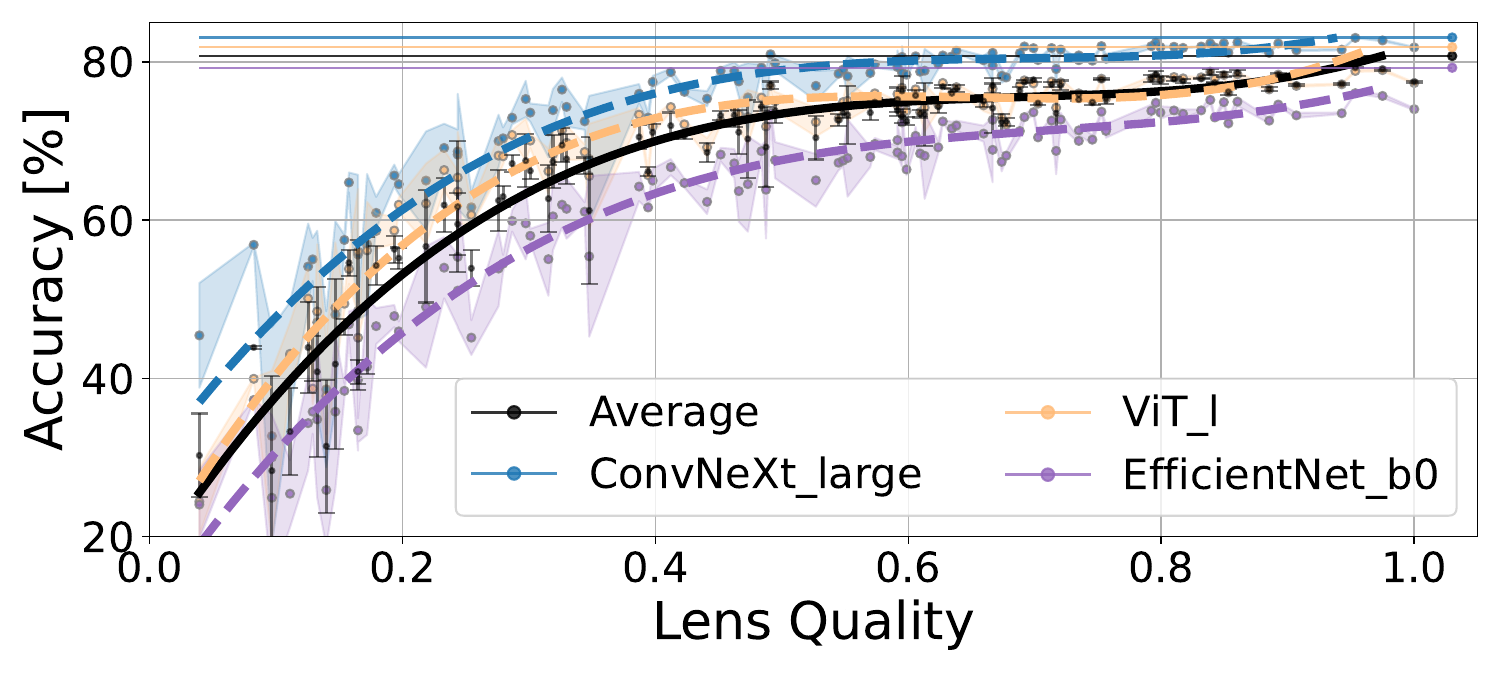}        
    \caption{
    Classification on ImageNet-100 lens corruptions for high and low performing DNNs and the average over all 12 DNNs. The subfigures visualize the ``trends" of the lens corruptions. (top) Distance and accuracy are correlated.
    (bottom) Lens quality and accuracy are very strongly correlated. 
    The higher the lens quality, the higher the accuracy. Error bars and envelope show the standard error over the corruptions for the field average. 
    The spread per lens varies from lens to lens. Higher quality lenses tend to have a smaller spread.}\label{fig:distance_and_quality_classification_vs_accuracy}
\end{figure}

The LensCorruptions dataset depends on two parameters namely the lens and where the lens is sampled, \ie the distance from the lens center. To resolve for these parameters, 
Fig.~\ref{fig:distance_and_quality_classification_vs_accuracy} shows the dependence of the accuracy on lens quality and field for high and low performing DNNs on the ImageNet-100 lens corruptions dataset. The average curves are drawn from all 12 available models.

Fig.~\ref{fig:distance_and_quality_classification_vs_accuracy} (top) displays the field dependence of the corruptions for the average (black curve) and selected high and low performing models. 
The average accuracy decreases by \SI{10.2}{\percent} for the central field and by a further \SI{7.4}{\percent} towards the edge of the lens. So, the classification task becomes more difficult as the field height increases, which is expected behavior. 
However, there are a few defocused lenses with poorer optical quality in the center of the lens \added{that} reduce this effect. Furthermore, the different DNNs' accuracies in Fig.~\ref{fig:distance_and_quality_classification_vs_accuracy} (top) are similar within a distance that is mainly determined by their validation accuracy. In addition, the large ConvNeXt\_l (blue) has the highest average accuracy over the field positions, while the EfficientNet\_b3  (red) has similar accuracy near the center of the lens, but decreases more towards the edge. Conversely, the large ViT\_l has significantly lower accuracy at the center of the lens, which levels out towards the edge of the lens. 

Fig.~\ref{fig:distance_and_quality_classification_vs_accuracy} (bottom) shows the accuracy for the selected models with averaged field versus lens quality. Each data point represents the accuracy of a single lens corruption. The horizontal lines show the validation accuracy without any corruption applied. The error bars and envelopes refer to the average in accuracy over field positions and the regression curve is drawn by a third-order polynomial.
As expected, the average of all models (black curve) is very strongly correlated with the lens quality measure (MTF50 average) with a Pearson correlation coefficient $\rho\!=\!+0.86$, Kendall Tau rank coefficient $\tau\!=\!+0.81$ and p-values close to zero. Furthermore, each lens has its own characteristic scatter. Lens corruptions with a large standard error are usually below the regression curve and tend to be more pronounced with lower lens quality and accuracy.  
Although the models have similar clean accuracy, their robustness depends on lens quality.

\subsection{Experiments on MSCOCO}
\label{subsec:detection_lens_corruptions}
We also evaluate the impact of lens corruptions for object detection models. 
To keep the experiments feasible, we generate a subset of the MSCOCO validation dataset with \num{1000} randomly selected images and predefined seed for reproducibility. We compare the full validation dataset consisting of \num{5000} images with the subset in the supplementary material~\ref{app:mscoco_subset}. 

Table~\ref{tab:detection_robustness_lenses_overview} lists the evaluated detection models together with their mean Average Precision (mAP) on the MSCOCO subset without any corruption applied, the average LensCorruptions and OpticsBench mAPs. 
The detection models have an average mAP of \ensuremath{\SI{47.0}{\percent}} that drops to an average \ensuremath{\SI{38.6}{\percent}} on  LensCorruptions. 
While the large and slow DINO model has the absolute best average mAP on \added{both} the clean dataset and on LensCorruptions,  YOLO-X is the most robust DNN with respect to its initial validation mAP. \begin{table}[t]
  \centering
  \caption{Detection results overview (mAP). If not further specified a ResNet50 is used as backbone. Clean= unmodified validation dataset, LC= Lens corruptions, OB=OpticsBench. All values are given as percentage values.}
  \begin{tabular}{l c c c}
    DNN / Backbone & Clean & LC & OB \\
    \hline
    Cascade R-CNN~\cite{cai_cascade_2021} & 40.7 & 31.4 & 18.4\\ 
    Casc. Mask R-CNN~\cite{cai_cascade_2021} / ConvNeXt & 51.9 & 43.2 & 29.4\\
    Deformable DETR~\cite{zhu_deformable_2020} & 46.6 & 36.8  & 21.9 \\
    DINO~\cite{caron_emerging_2021} / Swin\_l & \textbf{58.9} & \textbf{50.4} & \textbf{38.0} \\
    Faster R-CNN~\cite{lin_feature_2017} & 37.4 & 29.1 & 16.8\\
    Mask R-CNN~\cite{he_mask_2017} / Swin\_s & 48.5 & 40.1 & 27.7\\
    RetinaNet~\cite{lin_focal_2017} / Eff.Net\_b3 & 41.0 & 34.4 & 23.9\\
    YOLO-X\_x~\cite{ge_yolox_2021} / DarkNet & 50.6 & 43.8 & 30.6\\
    \hline
    $\Sigma$ & 47.0 & 38.6 & 25.7
  \end{tabular}
  \label{tab:detection_robustness_lenses_overview}
\end{table}

Table~\ref{tab:detection_field_dependence} lists the mAP values for the field dependence of lens blur corruptions for two high performers, one low performer, and the average of all eight models. The average field dependence is similar to the classification task: 
Moving away from the lens center, mAP decreases. Also, while the average mAP is relatively constant up to 0.3 lens quality, it decreases more rapidly from 0.5 towards the edge. 
\begin{table}[]
    \centering
    \caption{Field dependence of selected models. From the center (0.0) to the edge (0.9), mAP decreases more and more.}
    \label{tab:detection_field_dependence}
    \begin{tabular}{l llllll}
     DNN & clean & 0.0 & 0.3 & 0.5 & 0.7 & 0.9 \\
     \hline
    DINO & 58.9 & 52.8 & 52.1 & 50.5 & 49.3 & 47.4 \\
    YOLO{-}X  & 50.6 & 46.2 & 45.5 & 43.9 & 42.6 & 40.6 \\
    Faster R{-}CNN & 37.4 & 31.2 & 30.6 & 29.1 & 28.0 & 26.3 \\
    \hline
    $\Sigma$ & 46.9 & 40.9 & 40.3 & 38.7 & 37.6 & 35.7
    \end{tabular}
\end{table}

\begin{figure}
    \centering
    \includegraphics[width=\linewidth]{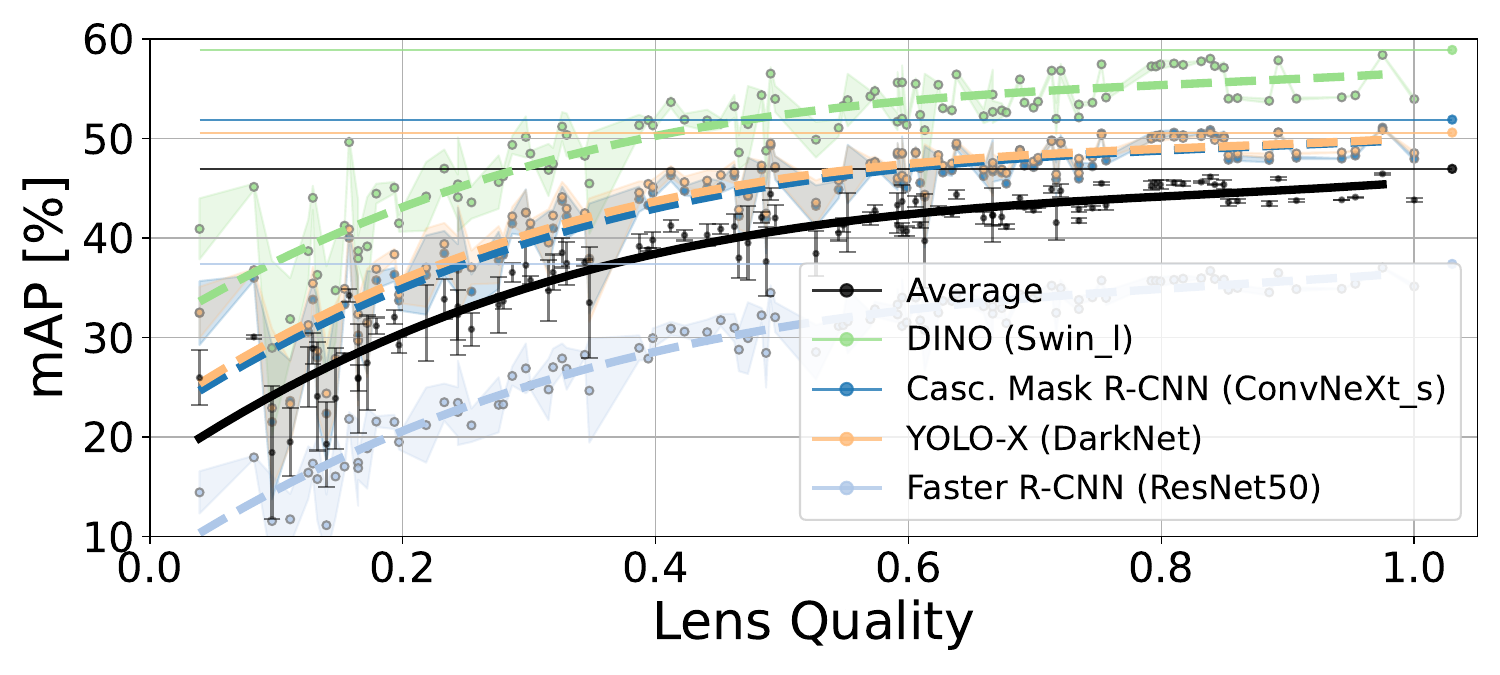}
    \caption{
    Detection on the MSCOCO LensCorruptions for two high performing, a low performing DNN, and the average over all 8 DNNs. 
    mAP versus lens quality for averaged field positions.  
    Lens quality and mAP correlate strongly (Pearson correlation coefficient $\rho\!=\!+0.84$\added{,} \deleted{and} Kendall Tau $\tau\!=\!0.65$\added{,} \deleted{with} p-values below \num{e-22}): the higher the lens quality, the higher the accuracy.}
\label{fig:distance_and_quality_detection_vs_accuracy}
\end{figure}

Fig.~\ref{fig:distance_and_quality_detection_vs_accuracy} shows the mAP in dependence on lens quality for all lens corruptions and selected high- and low-performance models. The horizontal lines refer to the clean mAP without any corruption applied. Similar to the classification results, the correlation between the lens quality and mAP is very strong ($\rho\!=\!0.89$, $\tau\!=\!78.8$, $p\!\approx\!0$) \added{on average} (black). The spread over the distance from the lens center decreases with higher mAP and lens quality, while the average mAP ranges from \SI{20}{\percent} to \SI{45}{\percent}. 
DINO (green) has consistently the highest mAP and YOLO-X is more robust than Cascade Mask R-CNN. 

\section{Limitations}
\label{sec:discussion}
OpticsBench and LensCorruptions can be used to analyze the model robustness to optical aberrations, and OpticsAugment \emph{increases} the robustness to optical aberrations. More specifically, the analysis with respect to architectural backbones allows for the selection of methods, but we do not see a stringent picture here: for example, model training has a strong influence on robustness and rather the combination of model architecture and training is crucial.

\section{Conclusion}
This article considered various types of optical aberrations and their impact on computer vision tasks of classification and object detection, with the goal to provide the community with a thorough evaluation framework for more realistic optical blurs. The main experiments are divided into image corruptions obtained from aberration base vectors (OpticsBench) and linear combinations of base vectors from real lens prescriptions (LensCorruptions). The lens corruptions from lenses of a wide range of qualities show that lens quality and accuracy or mAP correlate and that the difficulty increases with field height. OpticsBench corruptions compare to a single aberration and show that different aberration types may have different impact on DNNs. Their impact and difference to standard kernels were discussed. In addition to that, 
OpticsAugment demonstrates that such individual aberrations can be compensated  using data augmentation. This may allow \eg for matching lens and computer vision task in specific applications. 
We conclude that corruptions modeling specific blur types should be taken into account when optimizing computer vision systems.
\printbibliography
\begin{IEEEbiographynophoto}{Patrick M\"uller}
 is a postdoctoral researcher at University of Siegen  with a focus on out-of-distribution robustness of computer vision algorithms to optical aberrations. He received his M.Sc. from the Hochschule Düsseldorf, University of Applied Sciences, in Electrical Engineering and Information Technology and his PhD degree from University of Siegen under the supervision of Prof. Margret Keuper and Prof. Alexander Braun.
\end{IEEEbiographynophoto}
\begin{IEEEbiographynophoto}{Alexander Braun}
Alexander Braun received his diploma in physics and laser fluorescence spectroscopy from the University of Göttingen in 2001. His PhD research in quantum optics and quantum computers was carried out at the Universities of Hamburg and Siegen in 2007. He started working as an optical designer for camera-based ADAS with the company Kostal, and later became responsible for the optical quality of the series mass production. He is now researching optical metrology, optical models for simulation and the link between optical quality and computer vision performance at the University of Applied Sciences Düsseldorf.
\end{IEEEbiographynophoto}
\begin{IEEEbiographynophoto}{Margret Keuper}  is a full professor for Machine Learning with focus on computer vision at the University of Mannheim. She is also an affiliated research leader at the  Max Planck Institute for Informatics, Saarbrücken. She received her PhD degree from the University of Freiburg under the supervision of Prof Thomas Brox with her thesis entitled "Segmentation of Cells and Sub-cellular Structures from Microscopic Recordings". She worked as a postdoctoral researcher at the  University of Freiburg working on topics related to motion estimation, segmentation, and grouping. She is also an ELLIS fellow.
\end{IEEEbiographynophoto}
\vfill
\newpage 
\appendices
\appendices
\section*{Supplementary material}
This appendix provides additional details on the generation of our LensCorruptions and OpticsBench image corruptions and extends the analysis to more datasets and models. The content is structured as follows: 
\begin{itemize}
    \item Appendix~\ref{app:datasets} gives detailed information about the different datasets used on which we apply our blur corruptions.
    \item Appendix~\ref{app:models} lists the pre-trained models for image classification and object detection.
    \item Appendix~\ref{app:optical_aberrations} gives further insight of the PSF simulation. This includes the definition of Zernike polynomials and the generation of PSFs from wavefront aberration $W_\lambda$.
    \item Appendix~\ref{app:dataset_curation_details}  describes the LensCorruptions dataset.  We provide detailed insight into the dataset generation process and add additional visualizations of the dataset. In addition, we show a toleranced PSF.
    \item Appendix~\ref{app:optics_bench} describes details of OpticsBench. The kernel generation is outlined and then the matching process to the baseline blur kernel discussed. Also, the different OpticsBench kernels are visualized and compared to the baseline and the kernel alignment process is validated. 
    \item Appendix~\ref{app:evaluation_and_tables} extends the analysis on OpticsBench and OpticsAugment given in the main paper for image classification and object detection. It shows results for more severities and contains detailed tables to various aspects.
    \item Appendix~\ref{app:additional_detection_datasets} shows OpticsBench for the object detection dataset NuImages~\cite{caesar_nuscenes_2020}.
    \item Appendix~\ref{app:additional_lc} provides additional discussion of the results on LensCorruptions.
\end{itemize}

\section{Ground truth datasets}
\label{app:datasets}
This appendix gives more details about the datasets used to apply the various LensCorruptions and OpticsBench image blur corruptions. Standard vision datasets are used for both types of image corruption.
\subsection{ImageNet}
For image classification experiments we use both ImageNet-1k~\cite{russakovsky_imagenet_2015,deng_imagenet_2009} with \num[group-separator={,}]{1000} classes and \num[group-separator={,}]{50000} validation images and ImageNet-100~\cite{tian_contrastive_2020}, which is a subset of ImageNet-1k, with 100 classes. The 10 times smaller dataset having \num[group-separator={,}]{5000} validation images allows us to conduct more experiments with limited computational resources. Also, we follow~\cite{hendrycks_benchmarking_2019} to scale the validation images to the standard input resolution of $224\times224$ such that the relative blur kernel size remains constant.

\subsection{MS COCO}
\label{app:mscoco_subset}
For object detection, we use the standard image dataset MSCOCO~\cite{fleet_microsoft_2014}, which contains \num[group-separator={,}]{5000} larger validation images of different sizes and 80 classes with multiple annotations per image. The dataset contains a large number of persons as well as of a large number of diverse classes. The evaluated pre-trained DNNs are trained on the train split of MSCOCO, which contains \num[group-separator={,}]{118000} images. 

\begin{figure}
    \centering
    \begin{mysubfigure}{\linewidth}{
    \includegraphics[trim=0cm 0.4cm 0cm 0cm,clip,width=\linewidth]{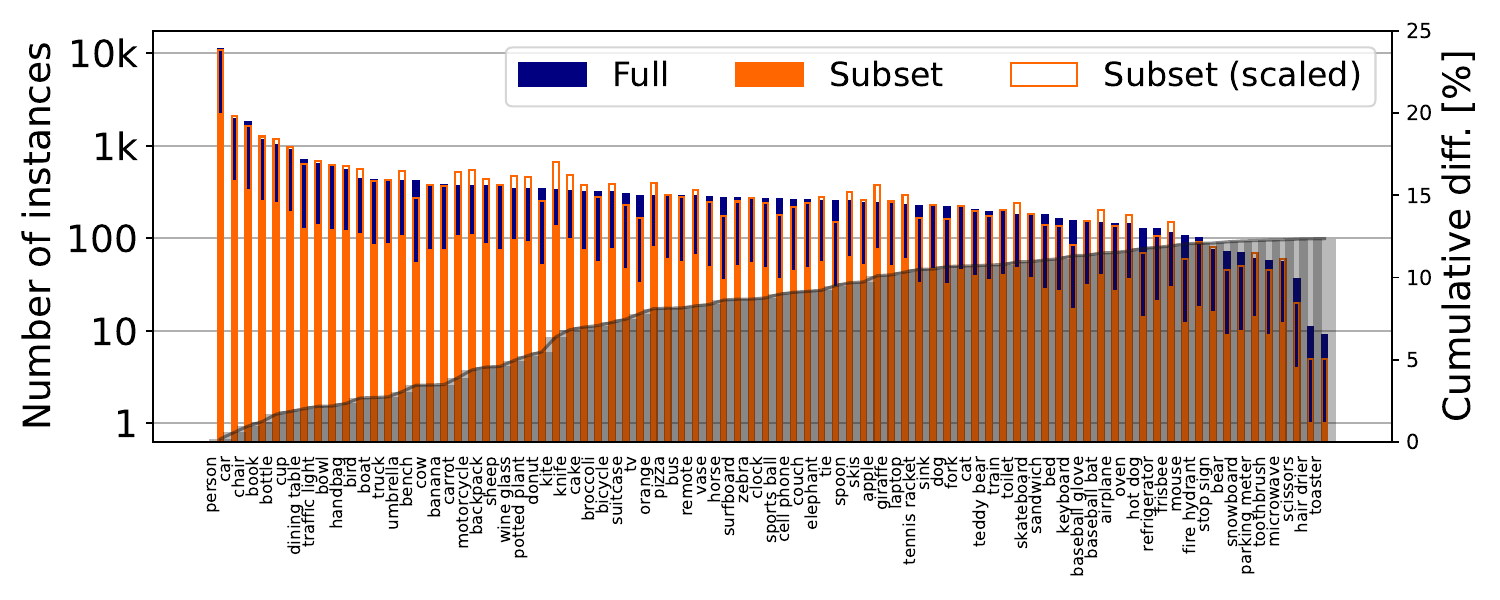}}{(a)}\label{fig:app_class_distribution}
    \end{mysubfigure}
    \begin{mysubfigure}{\linewidth}{
    \includegraphics[trim=0cm 0.25cm 0cm 0cm,clip,width=\linewidth]{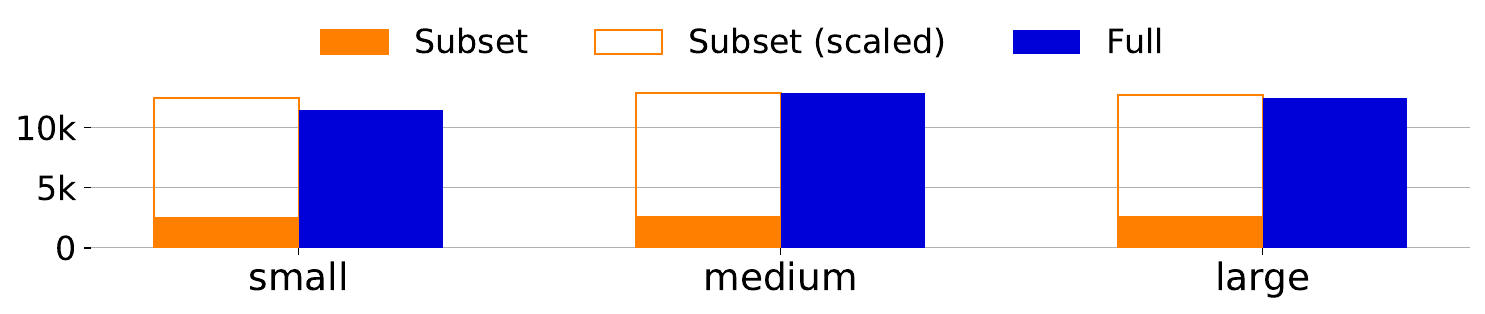}}{(b)}\label{fig:app_bbox_distribution}        
    \end{mysubfigure}
    \caption{To evaluate the difficulty of the MSCOCO subset used, we report the class distribution and the bounding box area distribution for the subset and the full set consisting of \num{1000} and \num{5000} images, respectively. For better comparability, we scale the subset by a factor of 5 - the ratio of the number of images from the full set and the subset. 
    The class distribution (a) for the subset (orange) is similar to the full MSCOCO validation dataset. The cumulative absolute deviation from the MSCOCO validation dataset shows a total difference of 12\% from the full validation set. The largest class person deviates by 0.7\%.
    The bounding box area sizes in (b) are very similar (linear scaling) for the subset compared to the full MSCOCO validation dataset. For comparability, we also scale the annotations from 1000 to 5000 images.}
    \label{fig:app_detection_subset}
\end{figure}

To keep the experiments on LensCorruptions feasible, we also generate a subset of the MSCOCO validation dataset with \num[group-separator={,}]{1000} randomly selected images and predefined seed for reproducibility. We compare the full validation dataset consisting of \num[group-separator={,}]{5000} images with the subset in Fig.~\ref{fig:app_detection_subset}. The dataset difficulty is here defined by bounding box size and class distribution. Both properties are similar, although there are slightly more small bounding boxes in the subset increasing difficulty. The cumulative deviation in class distribution differs by \SI{12}{\percent}. 
To further investigate the impact of the subset, we compare the mAP for the full validation dataset and our subset in Table~\ref{tab:difficulty_comparison_mAP}. The subset is very similar with an average increase in mAP of \SI{0.2}{\percent} points, so we conclude that the mAP on the subset is predictive to the full MSCOCO validation dataset.
\begin{table}
    \centering
    \caption{Validation mAP on the full set and the subset. The mAP values are very similar and differ by only \SI{0.2}{\percent} on average.}
    \label{tab:difficulty_comparison_mAP}
\begin{tabular}{@{}lll lll@{}}
    Model & Full & Subset & Model & Full & Subset\\
    \hline
    Cascade R-CNN & 40.3 & 40.7 &
    Faster R-CNN & 37.4 & 37.4\\
    Deform. DETR & 46.8 & 46.6 &
    RetinaNet & 40.5 & 41.0 \\
    Mask R-CNN/Swin & 48.2 & 48.5 & $\mathbf{\Sigma}$ & \textbf{42.6} & \textbf{42.8} \\
\end{tabular}
\end{table}

\section{Details on Classification \& Detection pre-trained models and backbones}
\label{app:models}
This appendix gives more detailed information about the used models for testing on the different datasets. 

\subsection{Image classification models}
\label{app:classification_models}
The last decade has been mainly dominated by Convolutional Neural Networks (CNN), with breakthroughs such as the ResNet~\cite{he_deep_2016} architecture and improvements such as EfficientNet~\cite{tan_efficientnet_2019} and ConvNeXt~\cite{liu2022convnet} leading to groundbreaking results on standard benchmarks such as ImageNet~\cite{russakovsky_imagenet_2015}. These significant improvements in model architecture are sometimes accomplished with the help of advanced methods such as Neural Architecture Search (NAS) and Compound Scaling~\cite{tan_efficientnet_2019}, which scales depth, resolution and width simultaneously. 
However, in recent years, transformer-based architectures such as Vision Transformer~\cite{dosovitskiy_image_2021} (ViT) and Swin\_v2~\cite{liu_swin_2022} have reached or exceeded the current state of the art at the cost of large models and long model training with the need for large amounts of training data. Furthermore, Smith et al.~\cite{smith_convnets_2023} show that CNN models, if trained with comparable compute budgets to vision transformers, can be on par with transformer-based architectures. CNNs remain therefore an important means of solving computer vision tasks. Therefore, we use DNNs of all types for a more comprehensive picture. 

We use pre-trained image classification models with default weights from torchvision 0.15.2 together with pytorch 2.x and python 3.11.x for all experiments on OpticsBench and LensCorruptions. Besides the architectures DenseNet, AlexNet, Inception, SqueezeNet and EfficientNet, the pre-trained models from pytorch were trained using an updated longer training schedule with AutoAugment~\cite{cubuk_autoaugment_2019}, MixUp~\cite{zhang_mixup_2018} and CutMix~\cite{yun_cutmix_2019} as data augmentations. This includes the ResNet50 model. Unfortunately, there is no publicly available training recipe for the RobustBench ResNet50 models. It is therefore difficult to determine the source of any gain or loss in accuracy compared to other models. All model inputs are scaled according to the corresponding recipe, which \eg scales the EfficientNets with increasing $\phi$.

The training recipe of OpticsAugment is based on the ImageNet\_v1 weights recipe with 90 epochs and no data augmentation as discussed in Sec.~\ref{app:training} to tackle limited computational ressources. Thus, our ResNet50 baseline model is having expectedly less rich and robust features, but still outperforms many other models on OpticsBench, including the strong pre-trained ResNet50. 

\subsection{Image classification models for fine-grained datasets Flowers102 and Stanford Cars}
\label{app:classification_models_fine_grained_datasets}
\added{
We fine-tune several pre-trained image classification models on Flowers102~\cite{nilsbackAutomatedFlowerClassification2008} and Stanford Cars~\cite{krause3DObjectRepresentations2013}. Flowers-102 consists of 102 different flower species from England, where we use an 80/20 split for training and testing. Stanford Cars consists of 196 different car makes with a roughly 50/50 split for training and testing. 
In both cases, we rescale and then center crop the images to the standard ImageNet resolution of $224 \times 224$ and apply the same data input normalization as for ImageNet. 
During training, we additionally apply basic data augmentations such as affine transformations, color jitter and flipping. In a first phase, we freeze all layers except the model head and train only this  to adapt to the fine-grained dataset. In a second phase, we optimize all model weights on Flowers102 with a layer-wise exponentially decaying learning rate with increased distance to the model head. For Stanford Cars training all model weights without layer-wise exponential learning rate decay resulted in significantly higher test accuracy. We anticipate that further hyperparameter optimization would increase the accuracy of the models further, which we cannot test with limited computational resources. Also, using more specialized models would improve the accuracy, but we want to stick to standard architectures for comparability.} 

\subsection{Object detection models}
\label{app:detection_models}
In object detection, the above described DNNs are used as the backbone of a larger model, where the model head outputs the instance localization and class. For example, the standard Faster R-CNN~\cite{ren_faster_2015} uses a ResNet backbone. However, the final bounding boxes and classes are obtained in a two-stage process: a region proposal network (RPN) finds candidate regions and then passes these to a refinement and classification stage. 
In addition to such a two-stage object detection architecture, there exist single-stage architectures like YOLO~\cite{redmon_you_2016}, which predict class probabilities and bounding boxes in a single pass through the network from the image divided into grid cells.
These primitives have been largely extended in recent years and supplemented by transformer based approaches like Deformable DETR~\cite{zhu_deformable_2020} and DINO~\cite{zhang_dino_2022}.
Another trend we are seeing is the development of non-specialised foundation models for vision like GLIP~\cite{cheng_yolo-world_2024}, 
which can solve various downstream tasks. 

Here, we use pre-trained models on MSCOCO 
from mmdetection~\cite{mmdetection} for testing on the different object detection datasets with LensCorruptions and the OpticsBench corruptions. A variety of architectures and backbones were considered in the model selection. We list all models used in Table~\ref{tab:app_dnn_overview_sup} and discuss the different settings. 
Besides R-CNN derivatives such as Faster R-CNN and Mask R-CNN we include other approaches based on CNNs include \eg YOLO (YOLOX-x, YOLOv3), RetinaNet, Fovea and FCOS. DETR and its extension Deformable DETR combine Vision Transformer and CNNs. Also, we use a Swin transformer model with DINO pre-training.
\begin{table}[h]
    \centering
    \small
    \caption{Evaluated Models on different object detection datasets.}  
    \label{tab:app_dnn_overview_sup}
    \begin{tabular}{@{}ll@{}}
        Model & Backbone \\
        \hline \\ 
        Faster R-CNN~\cite{ren_faster_2015} (FPN~\cite{lin_feature_2017}) & ResNet50 / ResNeXt101~\cite{he_deep_2016,xie_aggregated_2017} 
        \\
        Cascade R-CNN~\cite{cai_cascade_2021} & ResNet50 / ResNeXt101~\cite{he_deep_2016,xie_aggregated_2017} 
        \\
        Grid R-CNN~\cite{lu_grid_2019} & ResNeXt101~\cite{xie_aggregated_2017} 
        \\
        Mask R-CNN~\cite{he_mask_2017} & Swin-T/ Swin-S~\cite{liu_swin_2021}
        \\
        Casc. Mask R-CNN~\cite{cai_cascade_2021} & ConvNeXt-~\cite{smith_convnets_2023} \\
        YoloX (x/s) ~\cite{redmon_yolov3_2018} / YoloV3 & CSPDarknet53~\cite{redmon_yolov3_2018} 
        \\
        RetinaNet~\cite{lin_focal_2017} & EfficientNet\_b3~\cite{tan_efficientnet_2019} 
        \\
        Hybrid Task Cascade~\cite{chen_hybrid_2019} & ResNet50~\cite{he_deep_2016} 
        \\
        DETR / Def. DETR~\cite{carion_end--end_2020,zhu_deformable_2020} & ResNet50~\cite{he_deep_2016} 
         \\
         Fovea~\cite{kong_foveabox_2020} & ResNet50~\cite{he_deep_2016}
         \\
         FCOS~\cite{tian2019fcos} & ResNet101~\cite{he_deep_2016}
         \\
         ATSS~\cite{zhang_bridging_2020} & ResNet50~\cite{he_deep_2016}
        \\
        DINO~\cite{caron_emerging_2021} & Swin~\cite{liu_swin_2021}\\
        
        \\ 
    \end{tabular}
\end{table}

\section{PSF simulation}
\label{app:optical_aberrations}
This appendix discusses the PSF simulation in more detail.
OpticsBench and OpticsAugment rely on synthesized kernels obtained from Zernike polynomials, named after Frits Zernike~\cite{born_principles_1999,noauthor_handbook_2005}. We briefly describe the synthesizing process and application to images here. 
The kernel shapes that represent real optical aberrations use the linear system model for diffraction and aberration as in \cite{goodman_introduction_2017,born_principles_1999}, and define the wavefront aberrations by Zernike polynomials~\cite{born_principles_1999,zernike_beugungstheorie_1934,lakshminarayanan_zernike_2011}: 
\begin{equation}
W_{\lambda}(\rho,\varphi,\lambda) = \lambda \cdot \sum_{n,m} A_n^m(\lambda) \cdot Z_n^m(\rho,\varphi)
\label{eq:zernike_expansion_sup}
\end{equation}
The wavefront $W_\lambda$ is expanded into a complete and orthogonal set of polynomials $Z_n^m$, decomposed into a radial $R_m^n(\rho)$ and azimuthal factor depending on the sign of $m$:
\begin{equation}
    Z_n^m(\rho,\varphi) = R_n^m(\rho)  \cdot
\begin{cases}
    \mathrm{cos}(m\varphi), & \text{if } m < 0 \\
    \mathrm{sin}(m\varphi), & \text{if } m > 0
\end{cases}
\end{equation}
The radial polynomial $R_n^m(\rho)$ is defined as~\cite{born_principles_1999}:
\begin{equation}
    R^{\pm m}_n(\rho) = \sum_{s=0}^{\tfrac{n-m}{2}} \frac{(-1)^s\,(n-s)!}{s!\left (\tfrac{n+m}{2}-s \right )! \left (\tfrac{n-m}{2}-s \right)!} \;\rho^{n-2s}
\end{equation}
Each coefficient $A_n^m$ in multiples of the  wavelengths $\lambda_i$ represents the contribution of a particular type of aberration and therefore different aspects such as the amount of coma, astigmatism or defocus can be turned off or on. The coefficient $A_n^m$ is often also represented with a single index. $A_i$ is then obtained from a specific ordering, such as the Fringe indexing scheme, which we use here. Table.~\ref{tab:app_37coefficients} lists the first 37 Fringe coefficients and its optical equivalent.
 \begin{table}\caption{Zernike Fringe coefficients and corresponding aberration. The coefficients 1-3 represent an offset (Piston) and shifts (Tilt X, Y), which we exclude here.}\scriptsize\centering\begin{tabular}{lll}\\4: Defocus & 5: Astigmatism (straight) & 6: Astigmatism (oblique)\\7: Coma (horizontal) & 8: Coma (vertical) & 9: Primary Spherical\\10: Trefoil (horiz.) & 11: Trefoil (vertical) & 12: 2nd Astig. (straight)\\13: 2nd Astig. (oblique) & 14: 2nd Coma (horiz.) & 15: 2nd Coma (vertical)\\16: 2nd Spherical & 17: Tetrafoil (straight) & 18: Tetrafoil (oblique)\\19: 2nd Trefoil (horiz.) & 20: 2nd Trefoil (vertical) & 21: 3rd Astig. (straight)\\22: 3rd Astig. (oblique) & 23: 3rd Coma (horiz.) & 24: 3rd Coma (vertical)\\25: 3rd Spherical & 26: Pentafoil (horiz.) & 27: Pentafoil (vertical)\\28: 2nd Tetrafoil (straight) & 29: 2nd Tetrafoil (oblique) & 30: 3rd Trefoil (horiz.)\\31: 3rd Trefoil (vertical) & 32: 4th Astig. (straight) & 33: 4th Astig. (oblique)\\34: 4th Coma (horiz.) & 35: 4th Coma (vertical) & 36: 4th Spherical\\37: 5th Spherical \end{tabular} 
 \label{tab:app_37coefficients}
 \end{table}
In this article, we choose the Zernike Fringe Polynomials 4-11, including primary aberrations and elliptical coma (trefoil). Tilt x and tilt y are not considered here. 
This wavefront simulation can then be used to generate a PSF: The effect of a non-ideal lens on a point source of wavelength $\lambda$ can be compactly summarized by a Fourier transform $\mathcal{F}$ over a circular region that propagates the aberrated wave from pupil space to the image space. 
Fig.~\ref{fig:image_processing} visualizes this process. An ideal spherical wave passes through a circular pupil, which is then deformed according to the properties of the optical system. If there were no aberrations, the phase would be flat. 
\newcommand{\processSz}{0.15\linewidth}
\begin{figure}[h]
    \centering
    \begin{mysubfigure}{}{
        \includegraphics[trim=1cm 0.8cm 2cm 1.4cm,clip,width=\processSz]{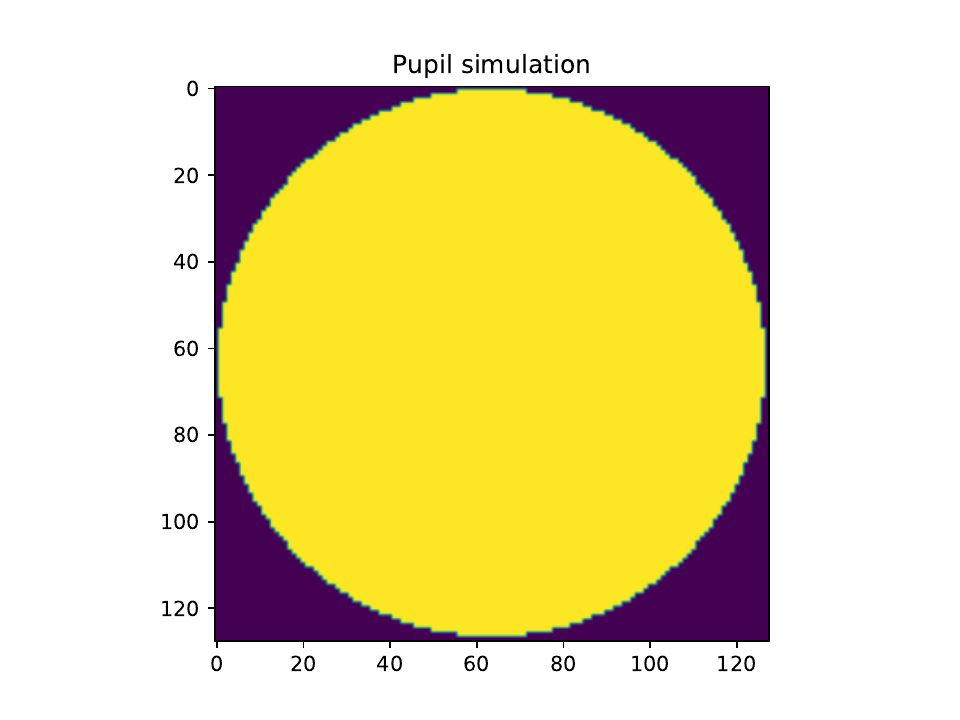}
        }{(a)}
    \end{mysubfigure}
    \begin{mysubfigure}{}{
        \includegraphics[trim=1cm 0.8cm 2cm 1.4cm,clip,width=\processSz]{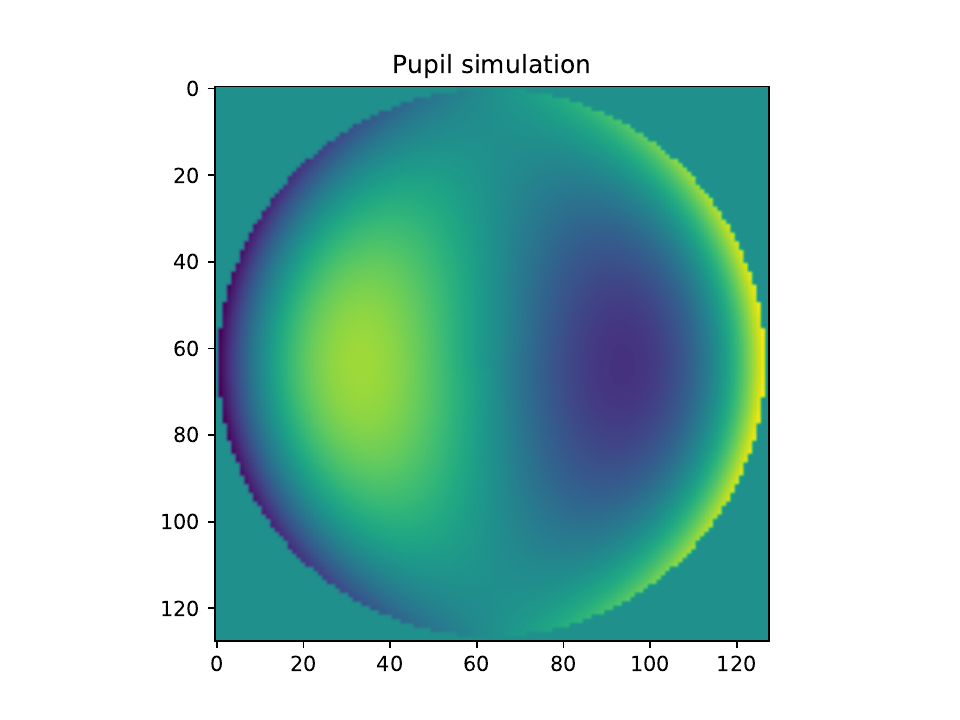}
        }{(b)}
    \end{mysubfigure}
    \begin{mysubfigure}{}{
        \includegraphics[trim=1cm 0.8cm 2cm 1.4cm,clip,width=\processSz]{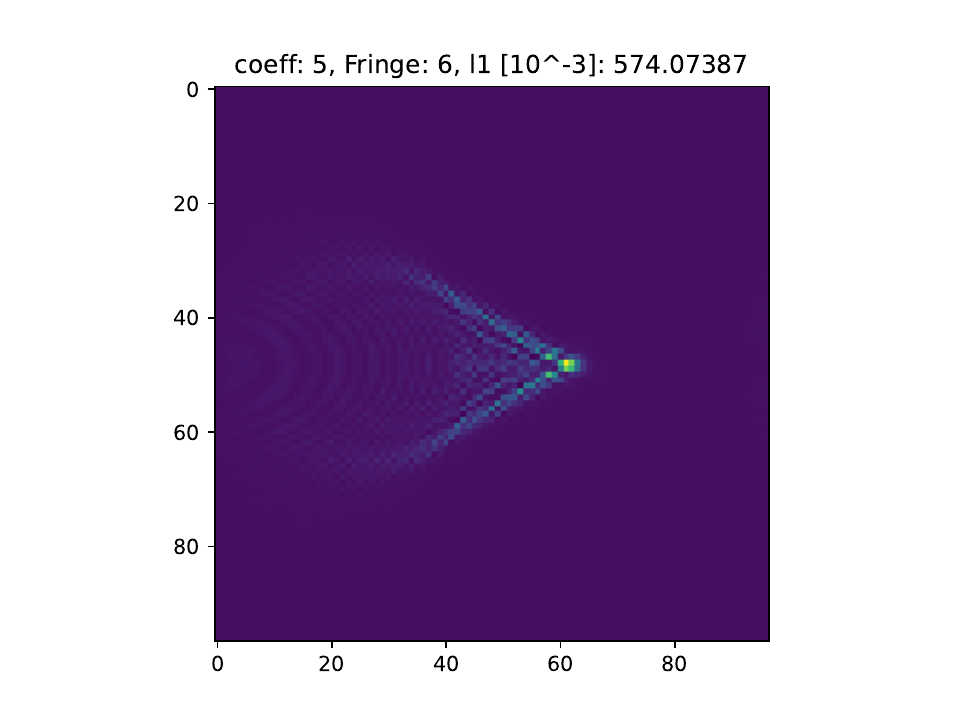}}{(c)}
    \end{mysubfigure} 
    \begin{mysubfigure}{}{
        \includegraphics[width=0.13\linewidth]{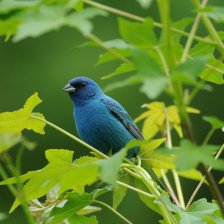}}
        {(d)}
    \end{mysubfigure}
    \begin{mysubfigure}{}{
        \includegraphics[width=0.13\linewidth]{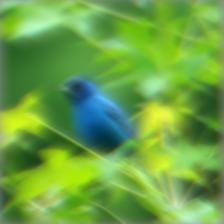}}{(e)}
    \end{mysubfigure}
    \caption{Image processing scheme: The circular pupil (a)  contains an aberrated phase (b), which is mapped into image space by a 2D Fourier Transform, yielding a PSF (c). This PSF is convolved with an image (d) to produce a blurred image (e). The process is repeated for each color channel.}
    \label{fig:image_processing}
\end{figure}
The complex phase factor describing the optical path difference $W_\lambda$ is transformed to image space at $z_i$ to yield a PSF for wavelength $\lambda$: \cite{goodman_introduction_2017}
\begin{equation}
    h(u,v,\lambda) = \vert \mathcal{F}\bigl\{(Circ(x,y) \cdot e^{-j\frac{2\pi}{\lambda z_i} \mathbf{W}_\lambda(x,y,\lambda)}\bigr\}\vert^2
    . 
    \label{eq:propagation}
\end{equation}
Convolving the kernel (c) and the image (d) results in the blurred image (e).

\section{Lens dataset curation \& imaging model}
\label{app:dataset_curation_details}
Here, we describe the process of generating the optical lens blur dataset used to create the various lens corruptions in more detail. 

\subsection{Lens dataset curation}
The collected lens dataset is based on publicly available photographic nominal lens files from the public domain database of the optical engineer Daniel J.  Reiley.~\cite{noauthor_navigation_nodate} The lenses are modeled in optical design software from patent literature and other public domain sources~\cite{noauthor_navigation_nodate}. 
There are several stages required to obtain our final lens blur kernel dataset: after initial dataset cleaning, we export a multitude of file types and select a subset of the large database. We then provide insight into the dataset and conclude with a discussion regarding the obtained image blur model. 

\paragraph*{Dataset cleaning}
The dataset is filtered for Zemax files. We then select lenses with available annotations and manually add the lens type, if it is missing.
Lenses with incorrect annotations were removed or where the lens type could not be determined. 
In addition, extender (teleconverter, adapter) elements for magnification, which are placed between the camera lens and sensor, were removed. Elements for which no valid glass catalogue could be found or raytracing were not possible are removed including afocal lenses. Also, six duplicate lenses were removed. In total 718 lenses are left from this initial dataset cleaning. 

\paragraph*{Export of metadata and PSF data}
From these lens samples Huygens PSFs are evaluated at various field heights for three different wavelengths at a data range of $256 \times 256$ and \SI{1}{\um} sampling density. The wavelengths are set to the standard Fraunhofer visible light wavelengths (F, d, C) to represent blue, green and red color channels. All PSFs are exported at five different field heights (field=\SIlist[list-units=single,list-final-separator = {, }, list-pair-separator= {, }]
  {0;30;50;70;90}{\%}) of the maximum Field of View and for off-axis positions additionally at three different orientations with an azimuth of (0, 45, 90) degrees to resolve the space-variance of a lens. The field height is measured here in percentage of the maximum Field of View (FoV) with 0\% representing the intersection with the optical axis, the imager center.
Other orientations would be obtained by rotating the PSFs or images respectively. 
We also collect metadata such as the system properties (EFL, FoV, F/\#, \dots) for various analyses, the MTF and Fringe Zernike coefficient data at two different fits (37, 12) at all field positions. 

\begin{figure}[h]
    \centering
    \includegraphics[width=\linewidth]{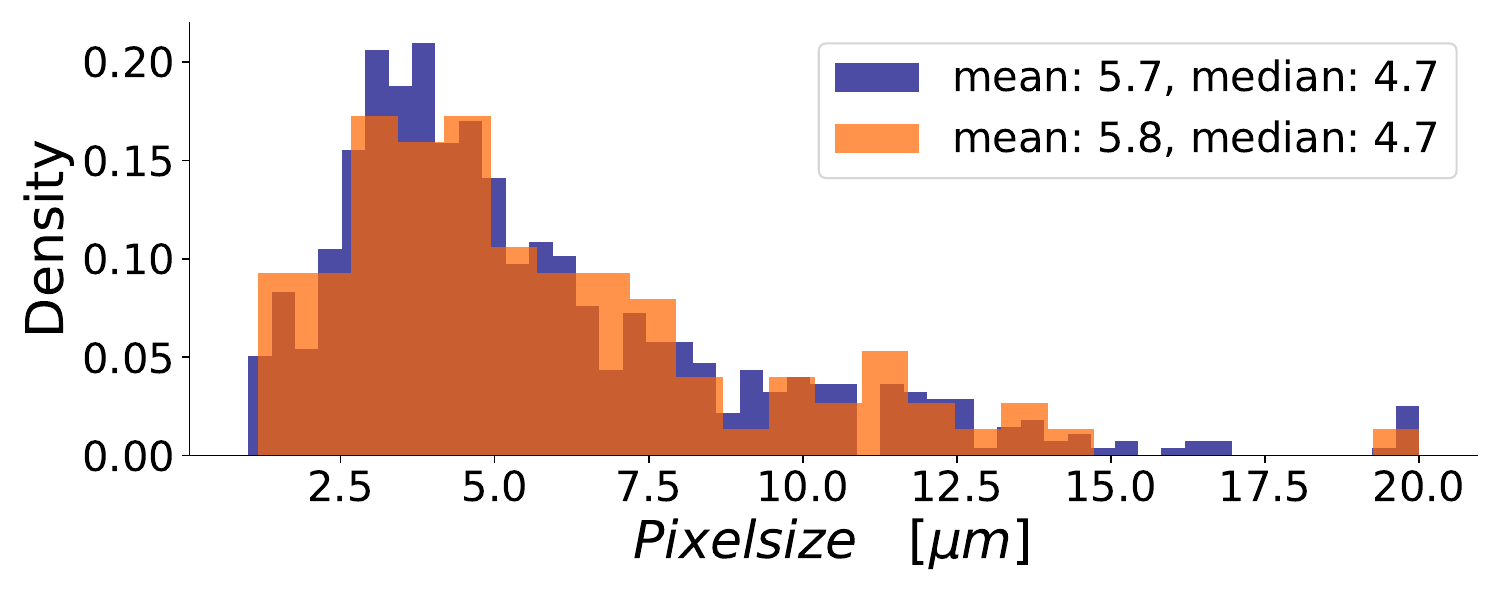}
    \caption{Distribution of virtual pixel sizes for all lenses (blue) and the subset of 100 lenses (orange). The pixel size of the selection ranges from \SIrange{1.0}{20.0}{\mu\metre} and the mean is at \SI{5.8}{\mu\metre}, the median at \SI{4.7}{\mu\metre}. These are common pixel sizes for CCD and CMOS image sensors.
    }
    \label{fig:app_pixel_sizes}
\end{figure}

\paragraph*{Scaling the PSF}
The PSFs of different wavelengths are first stacked to obtain RGB kernels with float precision. Their centroids are aligned with the array centre to avoid pixel shifts. These high-resolution kernels are then downsampled with a bicubic interpolation and an anti-aliasing filter to represent sampling with a virtual pixel size: during lens design, the lens PSF and camera pixel pitch are typically matched following a MTF-based criterion. 
The MTF directly shows how well fine details are reproduced and allows to balance resolution and sampling.~\cite{smith_modern_2000,loffler-mang_handbuch_2020}
The Nyquist criterion states that the maximum resolved frequency $f_{max}$ needs to be resolved by at least two samples: $f_{max} \leq 2f_{s} $.~\cite{gonzalez_digital_2009,noauthor_handbook_2005,loffler-mang_handbuch_2020}
However, assigning the Nyquist frequency to a lens is a trade-off between lens-limited or camera-limited resolution. A too small sampling frequency would waste the lens' performance capabilities and results in aliasing, but a high sampling frequency results in high data processing rates, more noise caused by then smaller pixels and other effects.
Here, we relate the sensor's Nyquist frequency $f_N$ to an established figure of merit in industry: The MTF20 value measures the frequency at which still $20\%$ contrast can be reproduced and relates to vanishing resolution.~\cite{garcia-villena_3d-printed_2021,loffler-mang_handbuch_2020} For all field positions we obtained the MTF20 value. We averaged these MTF20 values across wavelengths and field positions, but excluded values close to the edge as this resulted in a too low Nyquist frequency wasting lens quality. %
With this process, we obtain pixel sizes common to CMOS and CCD sensors between \SIlist{1;20}{\micro\metre}~\cite{loffler-mang_handbuch_2020}. The distribution of the pixel sizes is displayed in Fig.~\ref{fig:app_pixel_sizes}. 
A stricter criterion would result in unrealistically small pixel sizes and large array sizes.
Each PSF is scaled by the corresponding pixel size with bicubic downsampling and anti-aliasing from the original \SI{1}{\um} sampling rate. 
The resulting PSFs are then cropped using a criterion preserving $99.5\%$ of the encircled energy in a rectangular area to avoid unnecessary zero padding. 
The processing steps result in different tensor sizes across lenses. 
We then align the center of mass of the processed PSFs to their array centers and save for each lens a PyTorch tensor with float precision. 

\paragraph*{Lens selection}
From the remaining lenses, a subset $(n=100)$ is selected that covers a wide range of optical qualities.  We define the lens quality as the average of the field dependent MTF50 value, measured in cycles/px, obtained from the MTFs retrieved from the downsampled PSFs.
We use the downsampled data because it correlates with the predicted image quality and therefore the observable effect strength.
First, all lenses are sorted by average quality over the imaging field. This sorting is then sampled equidistantly by quality and the nearest neighbour lens sample is selected. 

\FloatBarrier
Fig.~\ref{fig:app_pixel_sizes} shows the distribution of pixel sizes assigned to the lens point spread functions (PSF). The selected lenses are assigned an average of \SI{5.8}{\micro \metre} with a median pixel size of \SI{4.7}{\micro \metre}. The pixel sizes range from \SIrange{1.0}{20.0}{\micro \metre}.

\begin{figure}
    \centering
    \begin{tikzpicture}
    \node (img) at (0,0) {    \includegraphics[width=\linewidth]{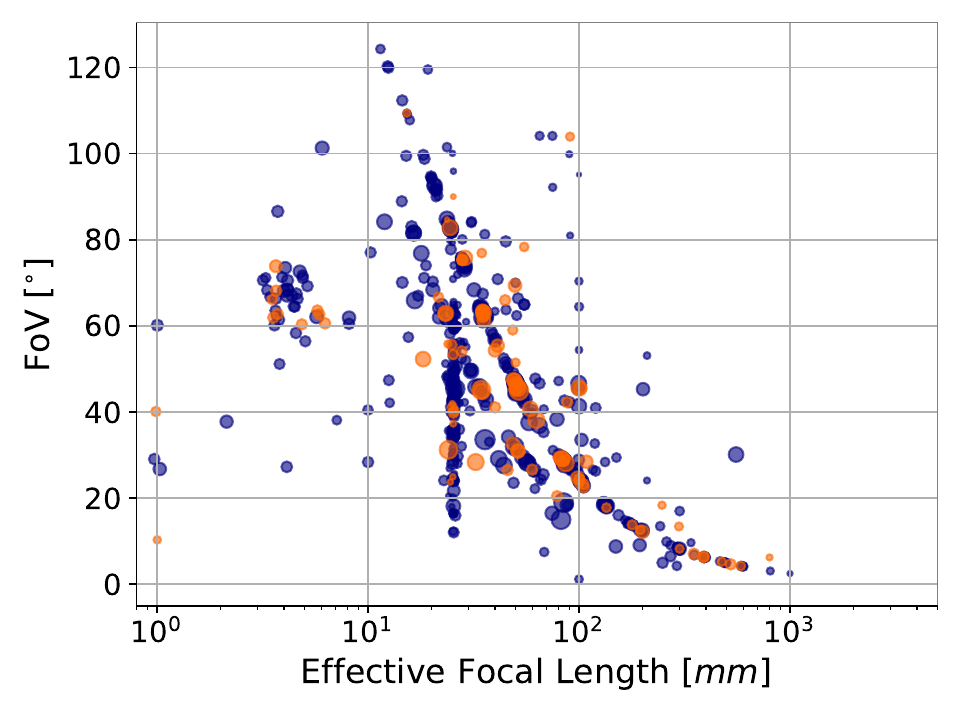}};
    \end{tikzpicture}
    \caption{Distribution of general camera parameters:  Effective Focal Length (EFL) and Field of View (FoV) in degrees  for focal lenses from~\cite{noauthor_navigation_nodate}. Spot size refers to aperture size, with larger circles indicating larger apertures, i.e. smaller f/\#. The orange dots represent the 100 lenses, which are included in the subset. The blue dots show lenses with available annotations, but not included in the subset. Intense colors result from overlapping data values. 
    As one can see, the subset picks lenses from a wide range of EFL and FoV.
    } 
    \label{fig:lens_files_efl_distribution}
\end{figure}

\paragraph*{Lens properties \& coefficient space} 
To verify the selection, we report here general lens properties and the diversity of the picked aberrations. The most characterizing system properties of a lens are the Field of View (FoV) and aperture, as well as the Effective Focal Length (EFL). Fig.~\ref{fig:lens_files_efl_distribution} shows the distribution of these properties in the full set (blue dots) and the subset (orange dots) used to define lens corruptions. The lenses can be categorized by FoV: The lenses well below \SI{40}{\degree}  represent telephoto types, above are standard types and from about \SI{60}{\degree} represent wide angle lenses.~\cite{noauthor_handbook_2008} 
The lens selection returns samples from all types and various EFL.

\begin{figure*}
    \centering
    \subfloat[\label{fig:app_coefficient_space_1}]{\includegraphics[width=0.33\linewidth]{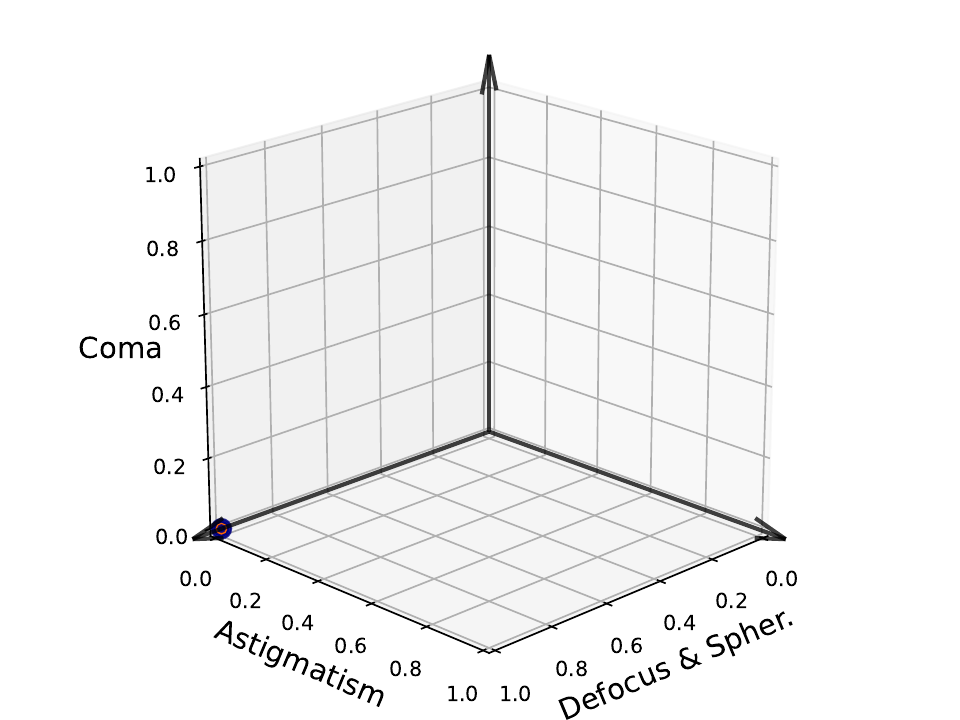}}
    \subfloat[]{ \includegraphics[width=0.33\linewidth]{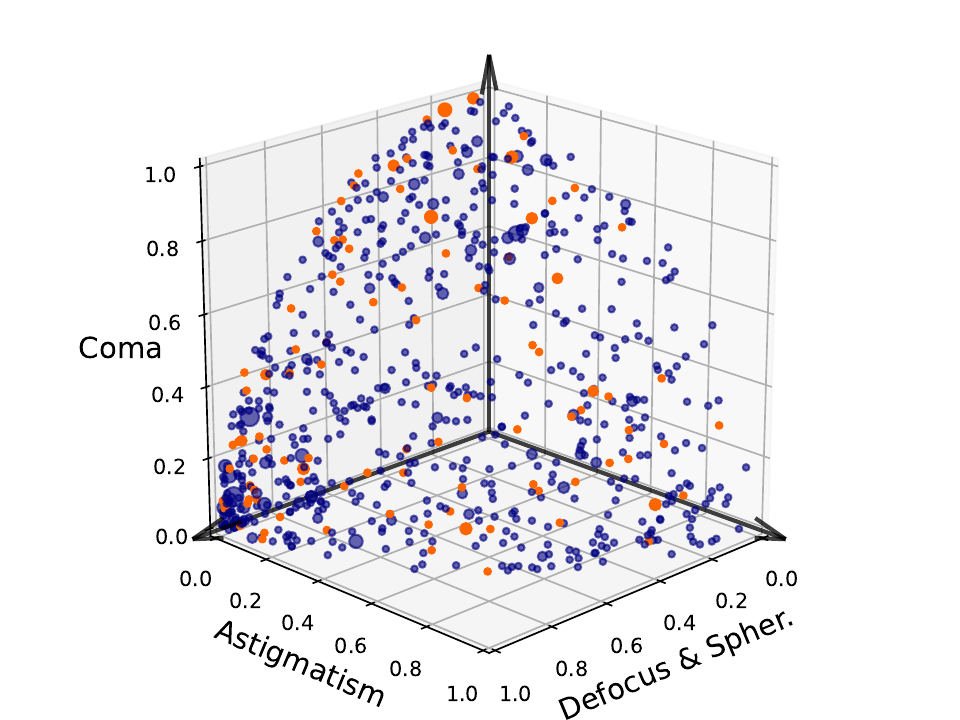}}
    \subfloat[]{\includegraphics[width=0.33\linewidth]{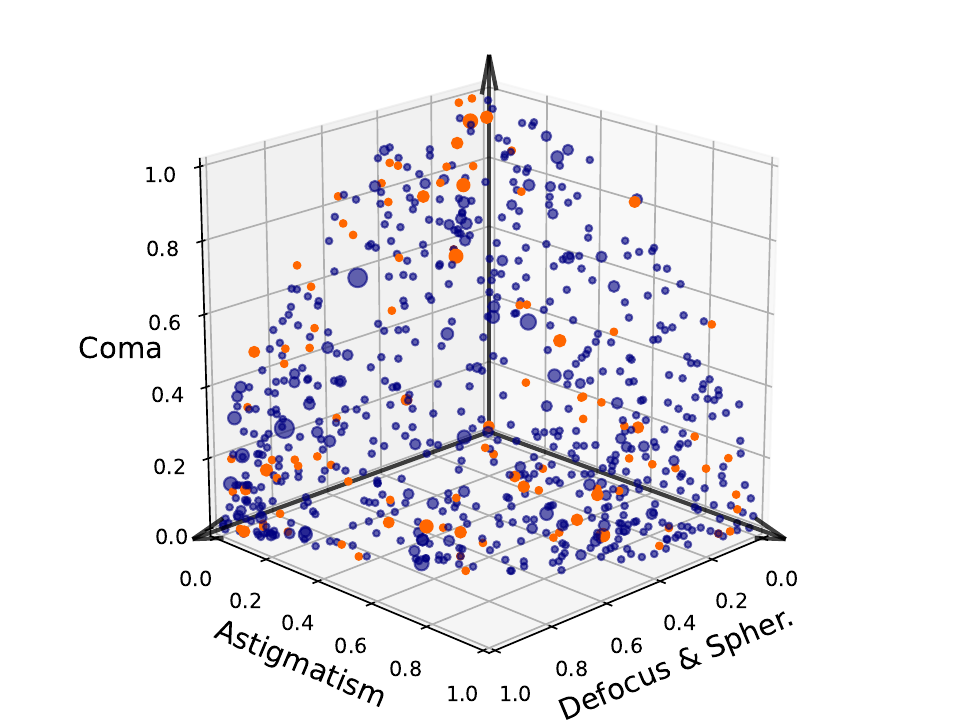}}\\
    \subfloat[]{\includegraphics[width=0.33\linewidth]{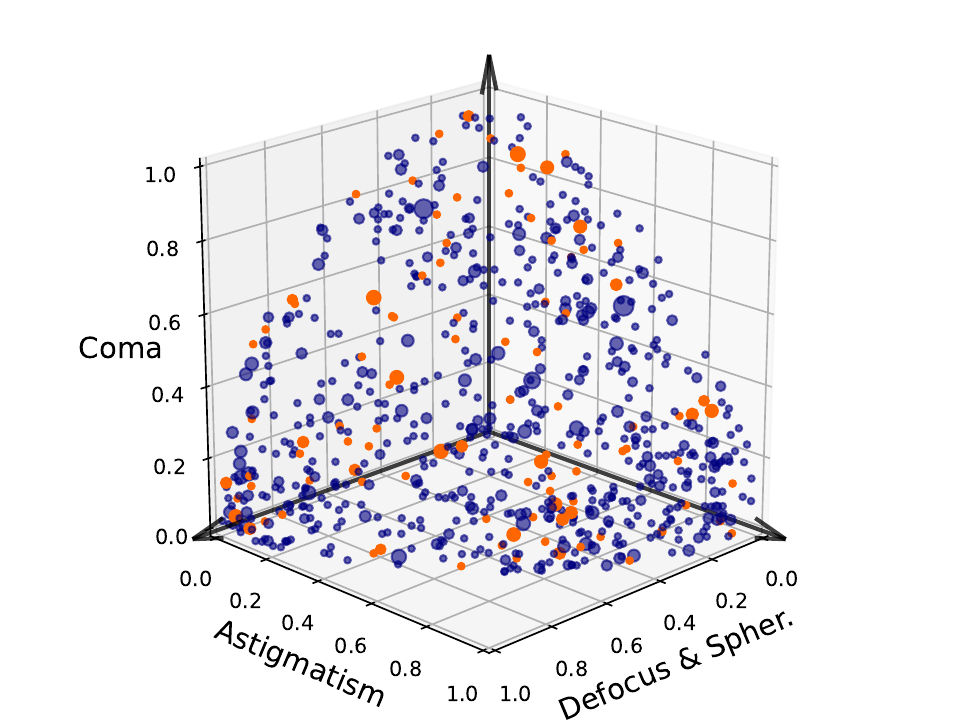}}
\subfloat[\label{fig:app_coefficient_space_5}]{ \includegraphics[width=0.33\linewidth]{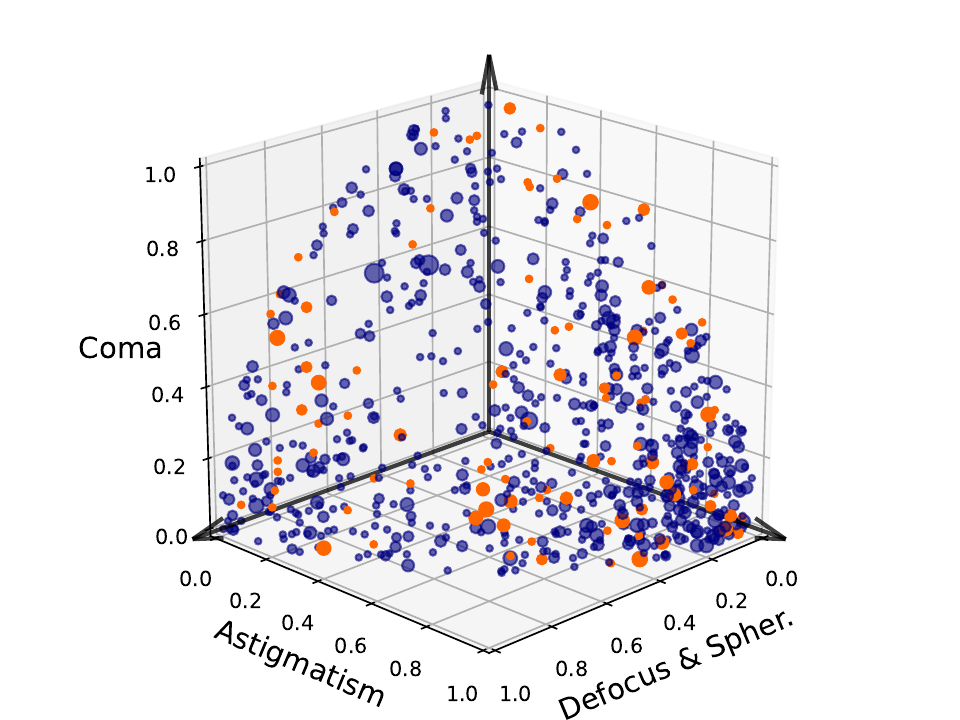}}
    \centering
    \caption{3D Zernike coefficient-space for different field positions illustrating the diversity of the lens aberrations. (a) 0.0, (b) 0.3, (c) 0.5\added{,} (d) 0.7 \added{and (e) 0.9} of the maximum Field of View (FoV). The orange dots represent the 100 selected lenses, the blue ones the excluded ones. As one can see, lenses are chosen all over the wide-spread coefficient-space. The dot size is controlled by the logarithmic vector norm. 
    Similarly to OpticsBench we choose three Zernike coefficient categories (defocus \& spherical, astigmatism and coma) derived from coefficient pairs. The corresponding orthogonal basis vectors define a 3D coefficient vector space into which we then project all lenses.
    To get these vectors, after taking the average over the wavelength, we map the coefficients $c$ from $c=12 \to 3$ coefficients by taking the \emph{absolute maximum value} for each coefficient pair. Then we select the dominant coefficient category from the different orientations and for a specific field. This yields a 3D unit vector after normalization pointing into the dominant direction. Here we show the resulting scatter plot for a medium field position. The plots for other field positions look similar except the coefficient vector distribution for the central field: here, almost all coefficient vectors are mapped to the defocus category basis vector, which is expected behaviour.} 
    \label{fig:app_coefficient_space}
\end{figure*}

In addition, we provide a visualization of the lens dataset using the Zernike coefficients. Fig.~\ref{fig:app_coefficient_space} shows the distribution of all lenses (blue dots) and the chosen subset (orange dots) in a 3D Zernike coefficient space for the different field positions. For the visualization, we select three primary aberrations  (defocus \& spherical aberration, astigmatism, coma) as base vectors and map for each lens a normalized Zernike coefficient vector into this space showing the dominant category. The dot size refers to the logarithmic vector's magnitude.

To obtain the positions, the following process is carried out for each lens. After taking the color average, the coefficient data source is available for different field positions and 12 coefficients. Mapping to a 3D coefficient category space requires the 12D coefficient vector to be reduced to just three dimensions. Similar to OpticsBench, we create pairs of coefficients: here, we combine straight and oblique astigmatism, horizontal and vertical coma, and defocus with primary spherical aberration. For these pairs, the absolute maximum value is selected per pair resulting in 3D coefficient vectors dependent on field position. We then choose the dominant coefficient category from the different azimuth values (r,x,y) and normalize the resulting vector.

The lens vectors in Fig.~\ref{fig:app_coefficient_space} show a variety of aberration combinations, with our selection spread across this coefficient category space. The base vectors were defined by the OpticsBench kernels. At the center of the lens shown in Fig.~\ref{fig:app_coefficient_space_1} all vectors are mapped to the defocus \& spherical base vector, which is the expected behaviour for on-axis PSFs. This also represents rotationally symmetric PSFs, which are sometimes defocused. This represents circular PSFs.
For off-axis positions, the dominant coefficient mixes with the other aberration base vectors and is distributed throughout the vector space. From an initial concentration of aberration vectors on the defocus \& spherical axis, the lens aberrations become stronger and shift to a dominant astigmatism for Fig.~\ref{fig:app_coefficient_space_5}.

\paragraph*{Imaging model}
The kernels from the dataset are applied to images using the convolution blur model from Sec.~\ref{sec:optical_aberrations}.
For the simulation, we assume that a small image with negligible spatial variance is a section in a larger imager shifted according to the field position. We argue, that this is a reasonable assumption since ImageNet or MSCOCO images of less than one megapixel are very small compared to standard multi-megapixel imagers.

\subsection{LensCorruptions dataset}
\label{app:validation_dataset}
\begin{figure}
    \centering
    \includegraphics[trim=0cm 0.5cm 0cm 0.5cm,clip,width=\linewidth]{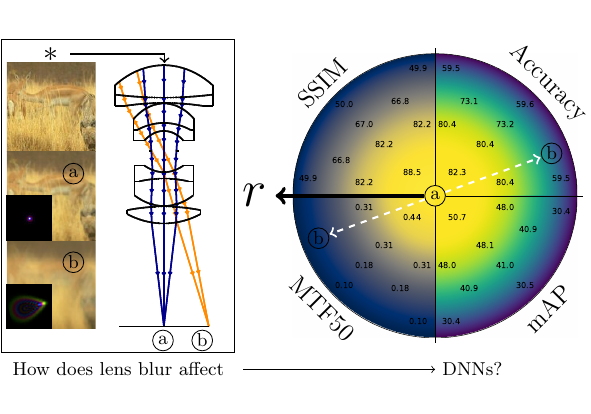}
    \caption{Lens blur sampled at different distances from the lens center, \eg at (a, b), its effect on images and the robustness of different computer vision algorithms. The left part of the figure shows how we generate our lens blur corruptions: an
    input image from ImageNet-100 at the top is convolved ($*$) with blur obtained from an optical simulation at positions a and b. In zoomed image b, the red deer is no longer visible. 
    The right part of the figure shows how this lens blur affects different quality metrics with distance from the lens center in four stacked rotationally symmetric maps. Each quadrant shows a different metric: optical (MTF50) and image (SSIM) quality and their correlation with the classification accuracy and object detection mAP. Yellowish and bluish colors indicate high and low quality. Higher values indicate higher quality.
    The quadrants are as follows: I) Classification accuracy on ImageNet100 for ConvNeXt, II) Image quality using SSIM between original images and the lens corruption, III) Optical quality using MTF50 relative to the Nyquist frequency, IV) Object detection mAP on MSCOCO subset and Cascade Mask R-CNN / ConvNeXt.}
    \label{fig:app_teaser2}
\end{figure}

The LensCorruptions dataset consists of 100 selected characteristic blurs obtained from the above lens dataset. For each corruption the lens is sampled at different positions and gives five different blurs with increasing distance to the lens center. 
These kernels are convolved with any image dataset and mimics the effect as if the image was recorded with the particular lens. This does not take into account geometric distortion or magnification, but concentrates on the characteristic lens blur.
Fig.~\ref{fig:app_teaser2} shows the idea of LensCorruptions. The lens blur is sampled at different positions for the lens, where each position gives a specific blur kernel, which can be used to blur an image. Again, we assume that the input image region of interest is shifted accordingly to obtain images (a) and (b) in Fig.~\ref{fig:app_teaser2}.
These images can then be inferenced by a DNN for classification or object detection.

\subsection{Lens PSF under production tolerance}
\label{app:production_tolerances}
To give an insight how these tolerances could affect a PSF we model simple tolerances for a lens having similar properties as an automotive lens:
five elements, f/\# 2.0, focal length \SI{5.8}{\mm}, hFoV \SI{48}{\degree} and total track length \SI{15}{\mm}. To show possible issues during mass production, we exemplarily decenter the second lens element by \SI{10}{\um} in Zemax and find by wavefront analysis that the PSF at 0.75 field height now has increased astigmatism and coma, one can see the PSF is rotated.
Such effects can occur during lens assembly.
\begin{figure}[h]
    \centering
    \begin{mysubfigure}{\linewidth}
    {\begin{tikzpicture}
    \node[anchor=south west,inner sep=0] (image) at (0,0){ \includegraphics[width=0.2\linewidth]{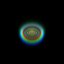}};
    \begin{scope}[x={(image.south east)},y={(image.north west)}]
    \node[white] at (0.5,0.2) {\tiny{\SI{50}{\um}}};
    \draw[white] (0.225,0.3)--(0.775,0.3);  
    \end{scope}    
    \end{tikzpicture}}
    {}
    \end{mysubfigure}
    \begin{mysubfigure}{\linewidth}{ \includegraphics[width=0.2\linewidth]{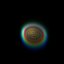}}{}
    \end{mysubfigure}
    \caption{High resolution real PSF (64x64) of a lens with similar properties to automotive lenses at 0.75 field height. (a) nominal prescription and (b) decentred element by \SI{10}{\um} causing rotation and deformation (superposition of coma and astigmatism). PSF size is approximately \SI{50}{\um}. The shape in (a) has concentric intensity rings depending on color, (b) deforms slightly.} 
    \label{fig:tolerances}
\end{figure}

\section{OpticsBench}
\label{app:optics_bench}
This appendix first describes the OpticsBench kernel generation in more detail, then describes the kernel matching process to the baseline kernels, discusses the kernels' center of mass alignment to the array center and concludes with a  visualization of OpticsBench image corruptions. 
\subsection{Kernels}
\label{app:optics_bench_kernels}
All kernels share a baseline optical wavefront model, which is adapted from~\cite{muller_simulating_2022} and evaluated at the center, i.e. at field 0° with little aberrations, but non-zero to ensure chromatic aberrations. The baseline model consists of the wavefront description displayed in Tab.~\ref{tab:baseline_wf_model}. We consider chromatic aberration an important aspect even for the study of individual Zernike coefficients. Although isolated Zernike modes are used to generate the dominating shape of the different kernels, this initialization ensures a chromatic variation of each primary effect. 
The different kernels are then generated by adding the isolated Zernike modes 4-11 from  Tab.~\ref{tab:coefficients} to the baseline wavefront model with Eq.~\ref{eq:zernike_expansion_sup} and retrieving a PSF with Eq.~\ref{eq:propagation}. This creates the kernels from Fig.~\ref{fig:kernels_cooke_rgb_iccv}. In practice, a more balanced distribution of coefficients is observed, which we investigate in detail with the \emph{LensCorruptions} benchmark. 
\begin{table}[h]
\caption{Wavefront baseline model used to produce the kernels (a,d,e) in Fig.~\ref{fig:kernels_cooke_rgb_iccv}. Other coefficients are zero, each value is in multiples of the wavelength $\lambda$ for RGB color channels red, green and blue: \SIlist{0.6563;0.5876;0.4861}{\micro\metre}. Zernike modes are in Fringe ordering, from left to right: defocus, spherical, secondary spherical and vertical quadrofoil as from~\cite{lakshminarayanan_zernike_2011}.}
\label{tab:baseline_wf_model}
\centering 
\begin{tabular}{@{}lllll@{}}
Color  &  4 & 9 & 15 & 16 \\
\hline
red 	& 	0.32671		&	  	0.088223	&  -0.061867	& -4.7631E-06 \\
green 	& 	0.11273		  & 	0.095923	&  -0.069497	&  -5.3967E-06 \\
blue 	& 	-0.41772	  & 	0.10825	    & -0.085119	&  -6.7436E-06 

\end{tabular}
\end{table}

The defocus blur kernels from~\cite{hendrycks_benchmarking_2019} are reproduced in Fig.~\ref{fig:kernels_baseline_defocus} for all severities to allow for comparison to disk-shaped kernels. As these kernels equally blur all color channels, they are grayscaled. 
\newcommand{\ksize}{0.098}
\newcommand{\ksizeBase}{0.098}
\begin{figure}[h]
\centering\begin{mysubfigure}{\ksizeBase\linewidth}{
\includegraphics[trim=3.75cm 1.5cm 3.5cm 1.5cm,clip,width=\ksizeBase\linewidth]{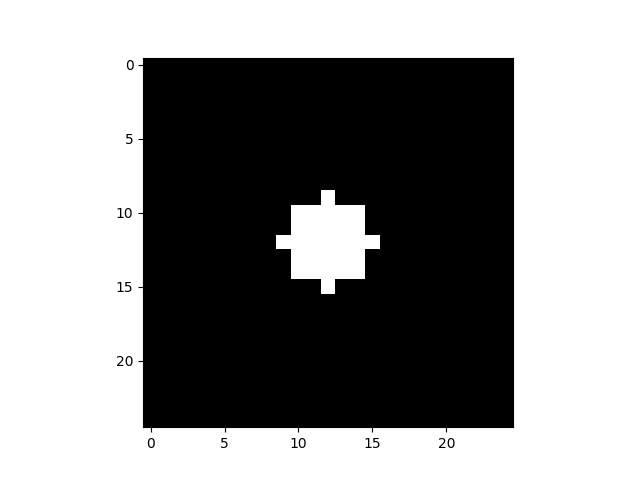}}{}
\end{mysubfigure}%
\begin{mysubfigure}{\ksizeBase\linewidth}{
\includegraphics[trim=3.75cm 1.5cm 3.5cm 1.5cm,clip,width=\ksizeBase\linewidth]{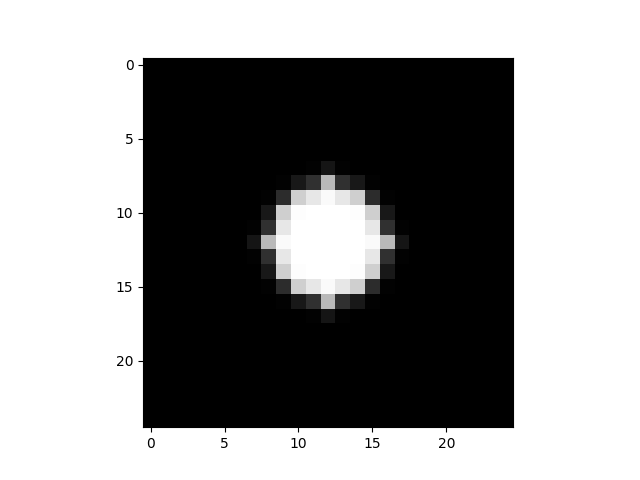}}{}
\end{mysubfigure}%
\begin{mysubfigure}{\ksizeBase\linewidth}{
\includegraphics[trim=3.75cm 1.5cm 3.5cm 1.5cm,clip,width=\ksizeBase\linewidth]{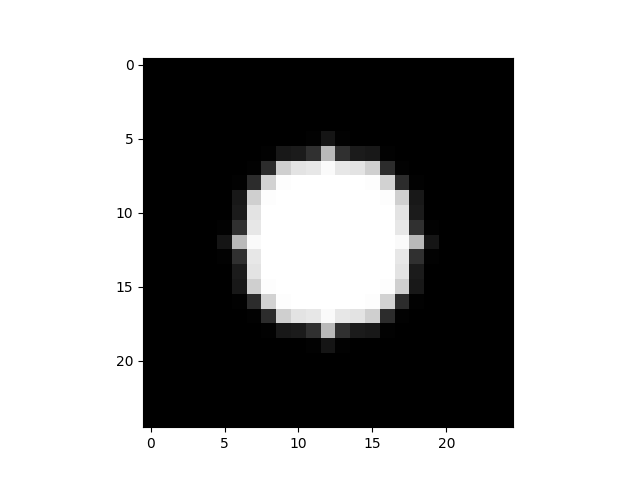}}{}
\end{mysubfigure}%
\begin{mysubfigure}{\ksizeBase\linewidth}{
\includegraphics[trim=3.75cm 1.5cm 3.5cm 1.5cm,clip,width=\ksizeBase\linewidth]{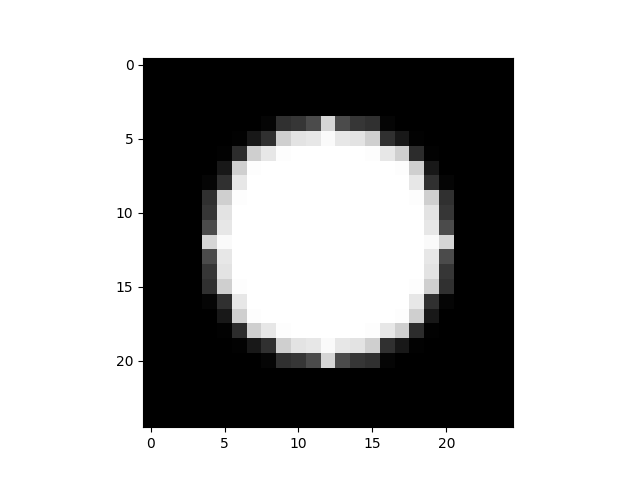}}{}
\end{mysubfigure}%
\begin{mysubfigure}{\ksizeBase\linewidth}{
\includegraphics[trim=3.75cm 1.5cm 3.5cm 1.5cm,clip,width=\ksizeBase\linewidth]{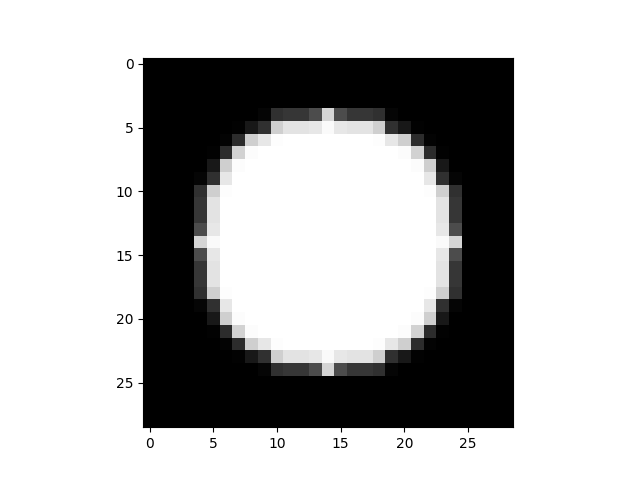}}{}
\end{mysubfigure}%
\caption{Defocus blur from~\cite{hendrycks_benchmarking_2019} for severities 1-5 used for kernel matching and as \emph{base blur type} for comparisons.}
\label{fig:kernels_baseline_defocus}
\end{figure}
Fig.~\ref{fig:kernels_cooke_rgb_iccv} shows the OpticsBench kernels used for generating the different image corruptions. 
\begin{figure}[h]
\centering
\subfloat[]{\begin{minipage}{\linewidth}
\subfloat{\includegraphics[width=\ksize\linewidth]{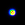}}\subfloat{\includegraphics[width=\ksize\linewidth]{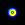}}\subfloat{\includegraphics[width=\ksize\linewidth]{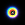}}\subfloat{\includegraphics[width=\ksize\linewidth]{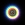}}\subfloat{\includegraphics[width=\ksize\linewidth]{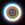}} \hspace{0.005cm} \subfloat{\includegraphics[width=\ksize\linewidth]{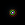}}\subfloat{\includegraphics[width=\ksize\linewidth]{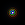}}\subfloat{\includegraphics[width=\ksize\linewidth]{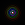}}\subfloat{\includegraphics[width=\ksize\linewidth]{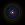}}\subfloat{\includegraphics[width=\ksize\linewidth]{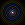}}
\end{minipage}\setcounter{subfigure}{1}}\\
\subfloat[]{\begin{minipage}{\linewidth}\centering
\subfloat{\includegraphics[width=\ksize\linewidth]{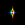}}\subfloat{\includegraphics[width=\ksize\linewidth]{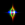}}\subfloat{\includegraphics[width=\ksize\linewidth]{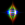}}\subfloat{\includegraphics[width=\ksize\linewidth]{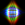}}\subfloat{\includegraphics[width=\ksize\linewidth]{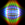}} \hspace{0.005cm} \subfloat{\includegraphics[width=\ksize\linewidth]{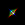}}\subfloat{\includegraphics[width=\ksize\linewidth]{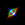}}\subfloat{\includegraphics[width=\ksize\linewidth]{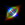}}\subfloat{\includegraphics[width=\ksize\linewidth]{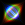}}\subfloat{\includegraphics[width=\ksize\linewidth]{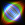}}
\end{minipage}\setcounter{subfigure}{2}}\\
\subfloat[]{
\begin{minipage}{\linewidth}\subfloat{\includegraphics[width=\ksize\linewidth]{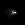}}\subfloat{\includegraphics[width=\ksize\linewidth]{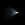}}\subfloat{\includegraphics[width=\ksize\linewidth]{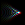}}\subfloat{\includegraphics[width=\ksize\linewidth]{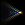}}\subfloat{\includegraphics[width=\ksize\linewidth]{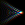}} \hspace{0.005cm} \subfloat{\includegraphics[width=\ksize\linewidth]{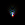}}\subfloat{\includegraphics[width=\ksize\linewidth]{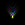}}\subfloat{\includegraphics[width=\ksize\linewidth]{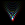}}\subfloat{\includegraphics[width=\ksize\linewidth]{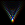}}\subfloat{\includegraphics[width=\ksize\linewidth]{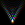}}
\end{minipage}\setcounter{subfigure}{3}}\\
\subfloat[]{\begin{minipage}{\linewidth}
\subfloat{\includegraphics[width=\ksize\linewidth]{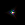}}\subfloat{\includegraphics[width=\ksize\linewidth]{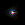}}\subfloat{\includegraphics[width=\ksize\linewidth]{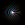}}\subfloat{\includegraphics[width=\ksize\linewidth]{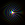}}\subfloat{\includegraphics[width=\ksize\linewidth]{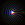}} \hspace{0.005cm} \subfloat{\includegraphics[width=\ksize\linewidth]{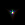}}\subfloat{\includegraphics[width=\ksize\linewidth]{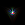}}\subfloat{\includegraphics[width=\ksize\linewidth]{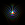}}\subfloat{\includegraphics[width=\ksize\linewidth]{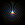}}\subfloat{\includegraphics[width=\ksize\linewidth]{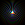}\setcounter{subfigure}{4}}
\end{minipage}}
\caption{Kernels used to generate OpticsBench. Each row contains the different severities (1-5) for a single corruption using two Zernike modes. Larger kernel sizes result in more severe blur. (a) defocus \& spherical, (b) astigmatism, (c) coma (center of mass aligned to array center), (d) trefoil. All kernels are $l_1$ normalized and therefore have the same energy. For this illustration, however, the kernels are normalized to the min-max of the colour channels, so that the different shapes become more apparent.}\label{fig:kernels_cooke_rgb_iccv}\end{figure}

\subsection{Kernel matching}
\label{app:kernel_matching}
To compare the impact of different kernel types on each other, it is crucial to have size-matched representations. As a baseline, we use the simple disk-shaped kernel prototype from Hendrycks et al.~\cite{hendrycks_benchmarking_2019} shown in Fig.~\ref{fig:kernels_baseline_defocus}. Then, with an educated initial guess of coefficient values two kernels are evaluated on different metrics and optimized by offsetting the coefficients in steps of $\pm0.1\lambda$. From this, we take the best overall fit as the kernel pair.

First, to compare two kernels, the Modulation Transfer Function (MTF) is obtained and evaluated in differently orientated slices (0°, 45°, 90°, 135°)~\cite{boreman_modulation_2001,goodman_introduction_2017}. From this, the frequency value at $50\%$ (MTF50) and the area under the curve (AUC) are obtained. The difference between these metrics provides a distance in  optical quality. Further, SSIM and PSNR are reported~\cite{wang_image_2004}. These metrics aim to compare the shape. 
Secondly, the kernels are convolved and then analyzed on established test images used in benchmarking camera image quality as another matching criterion.~\cite{phillips_camera_2018,noauthor_ieee_2017,noauthor_iso122332017_2017} 
We evaluate SSIM and PSNR on a slanted edge test chart and the scale-invariant spilled coins test chart, both at the required ImageNet target resolution of $224\times224$. The spilled coins chart, shown in Fig.~\ref{fig:spilled_coins}, consists of randomly generated disks of different sizes and is designed to measure texture loss as an image-level MTF.~\cite{mcelvain_texture-based_2010,burns_refined_2013} We report here the acutance and MTF50 value as evaluated with the algorithm of Burns~\cite{burns_refined_2013} from the spilled coins chart and compare once again between the two kernel representations. The evaluated image and optical quality on the spilled coins test chart serves here as proxy for other images from the image datasets.
\newsavebox{\imagebox}
\savebox{\imagebox}{\includegraphics[page=18,trim=0cm 8cm 21.6cm 3.5cm,clip,width=0.355\linewidth]{figures/matching/report_sheets_cooke_rgb_iccv.pdf}}%
\begin{figure}
\centering
\subfloat[\label{fig:spilled_coins}]{\centering
\raisebox{\dimexpr.5\ht\imagebox-.5\height}{ \includegraphics[width=0.175\linewidth]{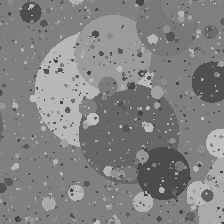}}}\hspace{.2cm}\subfloat[]{\centering\usebox{\imagebox}}\subfloat[]{\centering
\includegraphics[page=18,trim=9cm 8cm 12cm 3.5cm,clip,width=0.38\linewidth]{figures/matching/report_sheets_cooke_rgb_iccv.pdf}}
    \caption{(a) Spilled coins test chart generated with Imatest target generator tool~\cite{noauthor_imatest_nodate}. (b) MTF curves obtained from degraded versions of (a) and the corresponding PSF based MTF (c). The red curves refer to the baseline kernel at severity 3 and black to the corresponding astigmatism. Both curves are similar.}
    \label{fig:kernel_matching}
\end{figure}

Although the obtained optical kernels match also for higher severities with the defocus blur corruption~\cite{hendrycks_benchmarking_2019}, blur kernel sizes observed in the real world would be smaller, so this poses rather an upper bound on the  observable blur corruption for an entire image. However, small objects such as pedestrians in object detection datasets such as \cite{yu_bdd100k_2020} may suffer from such severe blurring as the object size decreases to tens of pixels.

\subsection{Validation of kernel alignment}
\label{app:validation}
Here, we measure the alignment of the kernels to their array center using the center of mass.
Table~\ref{tab:app_displacement} shows the average, minimum and maximum deviation of center of mass from the array center in pixels for the OpticsBench kernels. The maximum and minimum deviation are both caused by the coma corruption. Since this article also provides evaluation on object detection, where localization is important, we aligned for the kernels the center of mass to the array center, to avoid introducing a bias. \begin{table}[h]
    \centering
    \caption{Deviation of center of mass to the array's center of OpticsBench kernels. The second column shows the average and standard deviation, the third column the valley and peak deviation. The deviation hade been larger than 1 pixel only for coma.}
    \label{tab:app_displacement}
    \begin{tabular}{ccc}
        OpticsBench & (avg,std) & (min,max) \\
        \hline
        all & $(0.07,0.44)$ & $(-1.69,+1.1)$\\
        coma & $(0.38,0.68)$ & $(-1.69,+1.1)$\\
        no coma & $(0.03,0.25)$& $(-0.47,+0.25)$ 
    \end{tabular}
\end{table}

\begin{figure}[h]
    \centering
    \begin{mysubfigure}{\linewidth}{
    \includegraphics[trim=0cm 2.5cm 0cm 0cm,width=\linewidth]{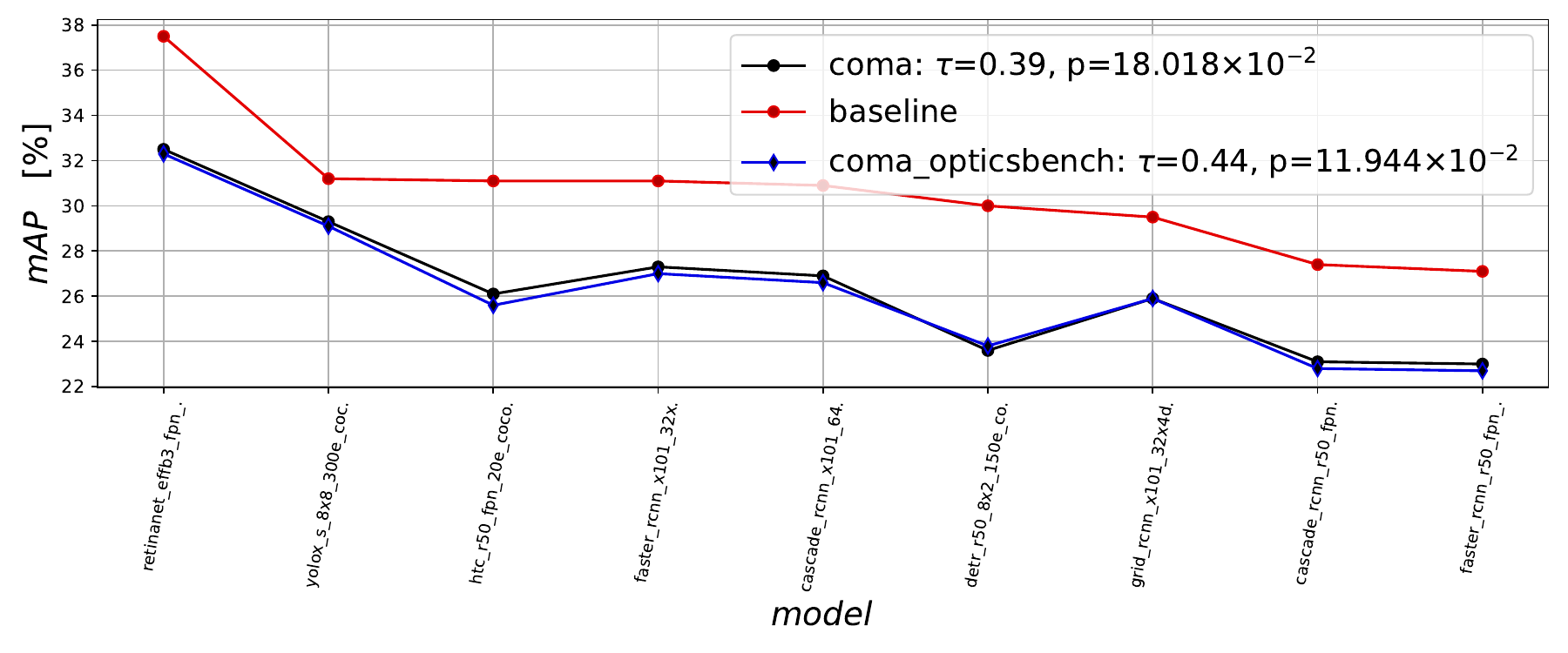}
    }{}
    \end{mysubfigure}
    \begin{mysubfigure}{\linewidth}{
    \includegraphics[trim=0cm 2.5cm 0cm 0cm, width=\linewidth]{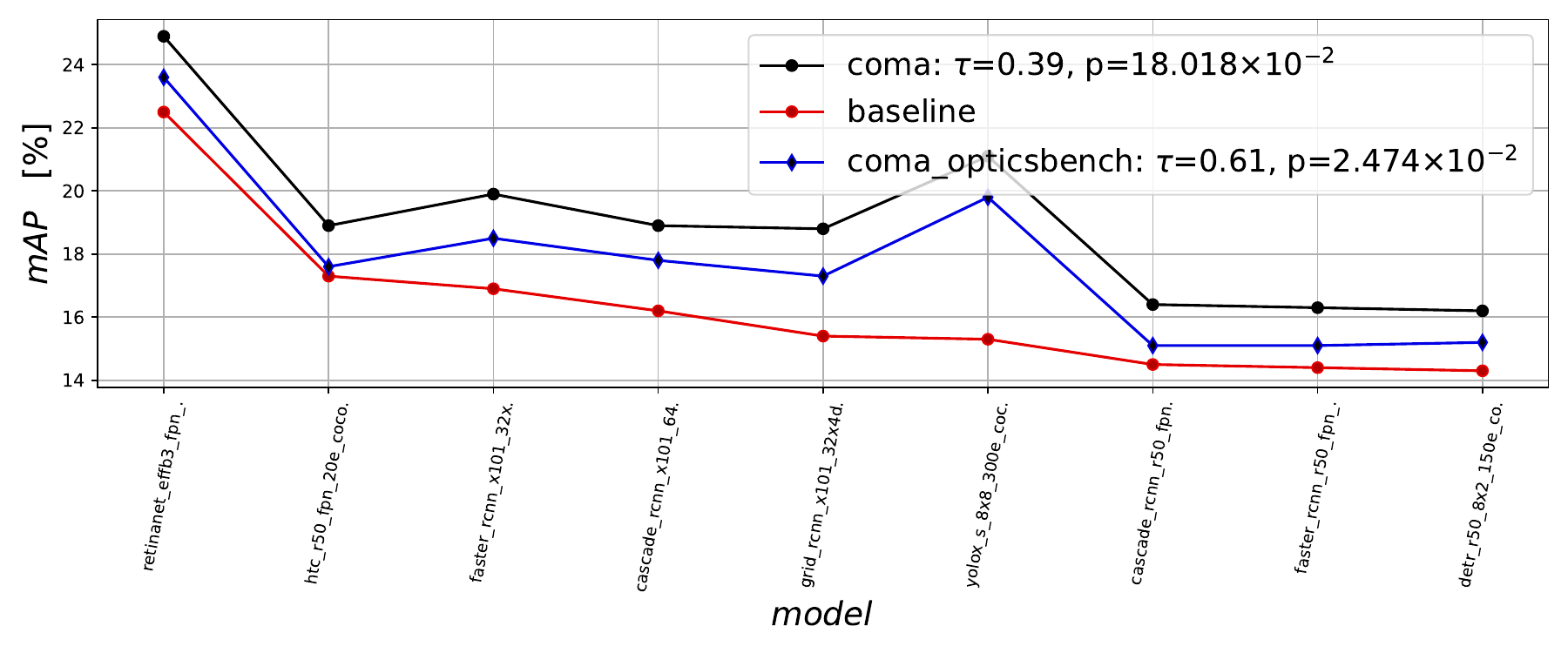}
    }{}\label{fig:kernel_alignment5}
    \end{mysubfigure}
    \caption{Comparison of coma corrruption with center of mass aligned to the array's center and the OpticsBench coma corruption evaluated on MSCOCO OpticsBench. The baseline represents defocus blur~\cite{hendrycks_benchmarking_2019}. The results are shown for severity 3 (top) and severity 5 (bottom). With increasing severity the effect of center of mass and array center misalignment becomes apparent: A systematic deviation in mAP is easily visible in (b). That's why we replaced the OpticsBench coma corruption with aligned kernels.}
    \label{fig:kernel_alignment}
\end{figure}

The comparison in Fig.~\ref{fig:kernel_alignment} shows little difference in mAP for the aligned and original kernels and severity 3, but a higher difference for severity 5. The difference in mAP observed in Fig.~\ref{fig:kernel_alignment5} appears to be systematic: the aligned version of the coma corruption (black curve) is always above the non-aligned OpticsBench coma corruption (blue curve) with a constant difference between these curves over a wide range of models. This confirms that the centre of mass of the kernels must be aligned with the centre of the array for the evaluation of object detection datasets with OpticsBench. 

\subsection{OpticsBench - Image examples}

\begin{figure*}[h]
	\centering
	
\begin{adjustbox}{width=\linewidth}
\begin{tikzpicture}

	\coordinate (ll) at (0,0);
	\coordinate (ur) at (\linewidth,\linewidth);
	
	\draw[black,dashed,opacity=.0] (ll) rectangle (ur) node[midway] (center){};
	

	\def\imSz{0.08\linewidth}
	\setlength{\tabcolsep}{3pt}	
	\renewcommand{\arraystretch}{2.}	

	\def\offx{+.025\linewidth}
	\def\offy{-.027\linewidth}
	
	\def\sepSpace{\hspace{5em}}
	\node[anchor=center,align=center] (table) at ($(center.center) + (\offx,\offy)$)
	{
	
	\begin{tabular}{cccccc cccccc}
	
	
	\includegraphics[width=\imSz]{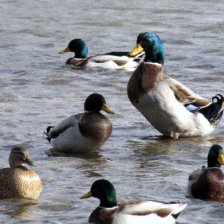} \sepSpace & 	\includegraphics[width=\imSz]{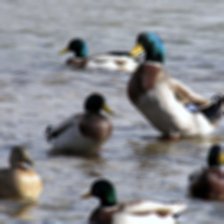}& \includegraphics[width=\imSz]{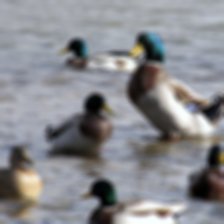}& \includegraphics[width=\imSz]{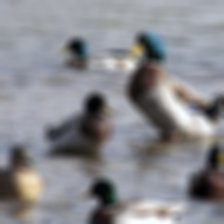}& \includegraphics[width=\imSz]{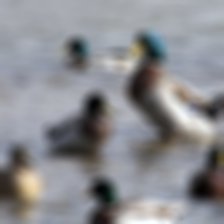}& \includegraphics[width=\imSz]{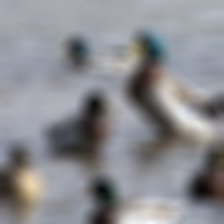}& 
	
	\includegraphics[width=\imSz]{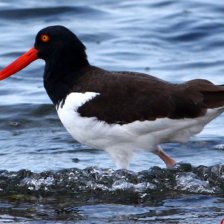} & 	\includegraphics[width=\imSz]{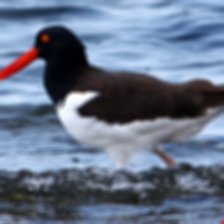}& \includegraphics[width=\imSz]{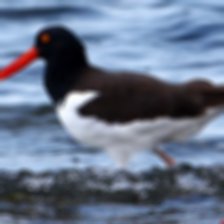}& \includegraphics[width=\imSz]{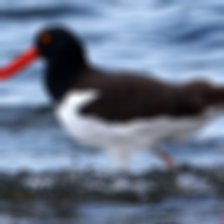}& \includegraphics[width=\imSz]{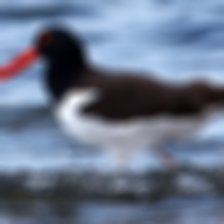}& \includegraphics[width=\imSz]{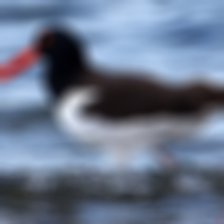}

	\\
	
	 \includegraphics[width=\imSz]{figures/dataset_tpami/clean/ILSVRC2012_val_00000415.JPEG} \sepSpace& 	 
	 \includegraphics[width=\imSz]{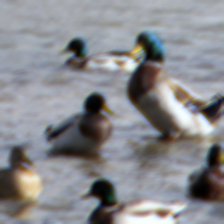}& 	 
	 \includegraphics[width=\imSz]{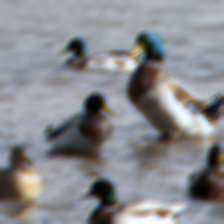}& \includegraphics[width=\imSz]{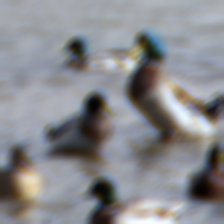}& \includegraphics[width=\imSz]{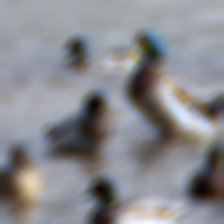}& \includegraphics[width=\imSz]{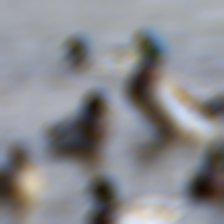}& 
	 
	 \includegraphics[width=\imSz]{figures/dataset_tpami/clean/ILSVRC2012_val_00020025.JPEG}& 	 \includegraphics[width=\imSz]{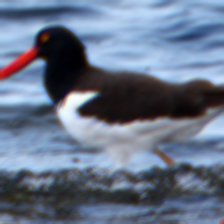}& \includegraphics[width=\imSz]{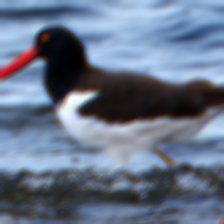}& \includegraphics[width=\imSz]{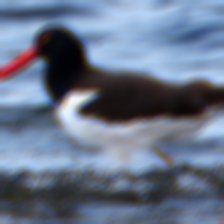}& \includegraphics[width=\imSz]{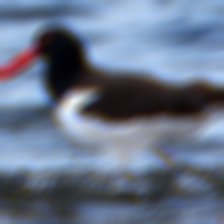}& \includegraphics[width=\imSz]{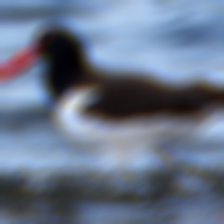}
	
	 \\ 
		 
	 \includegraphics[width=\imSz]{figures/dataset_tpami/clean/ILSVRC2012_val_00000415.JPEG} \sepSpace & 	 \includegraphics[width=\imSz]{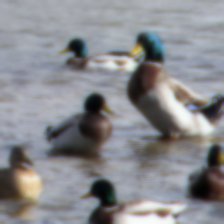}& \includegraphics[width=\imSz]{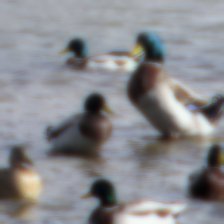}& \includegraphics[width=\imSz]{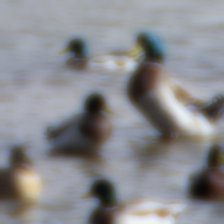}& \includegraphics[width=\imSz]{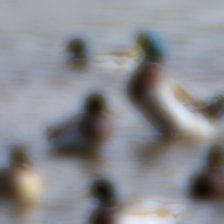}& \includegraphics[width=\imSz]{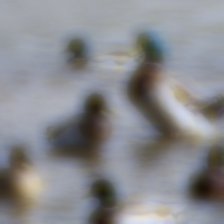}& 
	 
	 \includegraphics[width=\imSz]{figures/dataset_tpami/clean/ILSVRC2012_val_00020025.JPEG}& 
	 \includegraphics[width=\imSz]{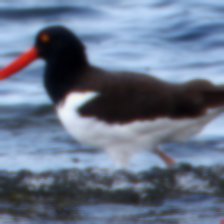}& \includegraphics[width=\imSz]{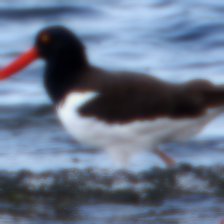}& \includegraphics[width=\imSz]{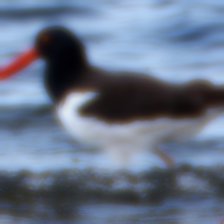}& \includegraphics[width=\imSz]{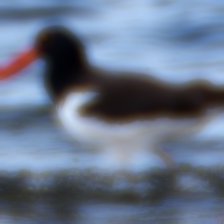}& \includegraphics[width=\imSz]{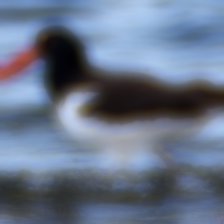}

	 \\  
	 	 
	 \includegraphics[width=\imSz]{figures/dataset_tpami/clean/ILSVRC2012_val_00000415.JPEG} \sepSpace& 	 \includegraphics[width=\imSz]{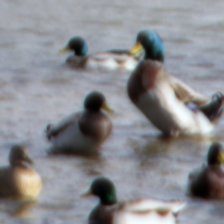}& \includegraphics[width=\imSz]{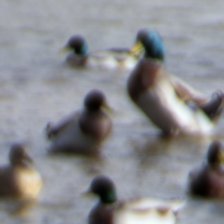}& \includegraphics[width=\imSz]{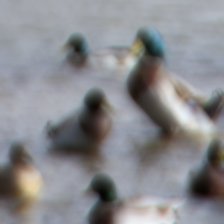}& \includegraphics[width=\imSz]{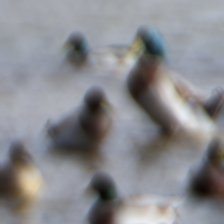}& \includegraphics[width=\imSz]{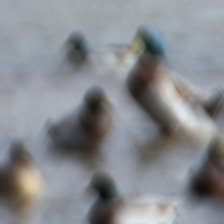}& 
	 
	 \includegraphics[width=\imSz]{figures/dataset_tpami/clean/ILSVRC2012_val_00020025.JPEG}& 	 \includegraphics[width=\imSz]{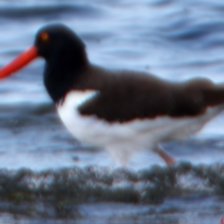}& \includegraphics[width=\imSz]{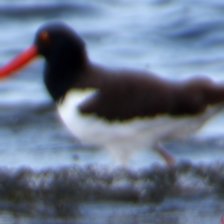}& \includegraphics[width=\imSz]{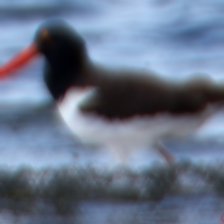}& \includegraphics[width=\imSz]{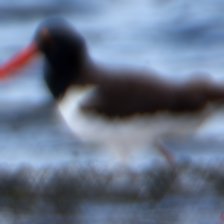}& \includegraphics[width=\imSz]{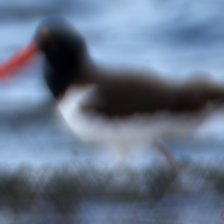}
	 
	 \\  
	 
	 \includegraphics[width=\imSz]{figures/dataset_tpami/clean/ILSVRC2012_val_00000415.JPEG} \sepSpace & 	 \includegraphics[width=\imSz]{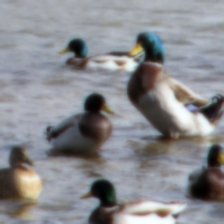}& \includegraphics[width=\imSz]{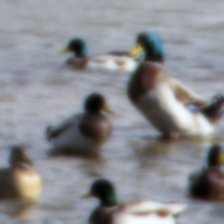}& \includegraphics[width=\imSz]{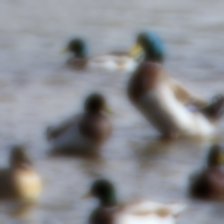}& \includegraphics[width=\imSz]{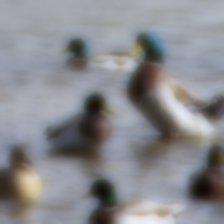}& \includegraphics[width=\imSz]{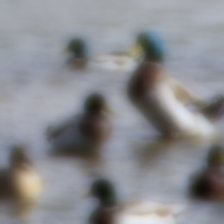}& 
	 	 
	 \includegraphics[width=\imSz]{figures/dataset_tpami/clean/ILSVRC2012_val_00020025.JPEG}& 	 \includegraphics[width=\imSz]{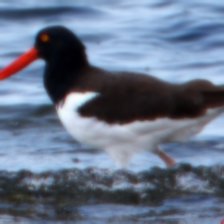}& \includegraphics[width=\imSz]{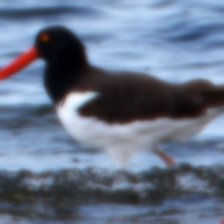}& \includegraphics[width=\imSz]{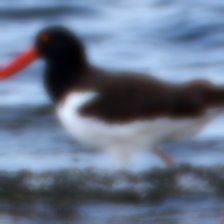}& \includegraphics[width=\imSz]{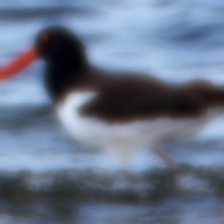}& \includegraphics[width=\imSz]{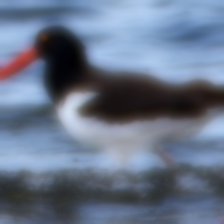}
	 
	\\
	 

	 \includegraphics[width=\imSz]{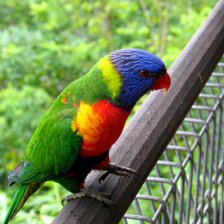} \sepSpace & 	 \includegraphics[width=\imSz]{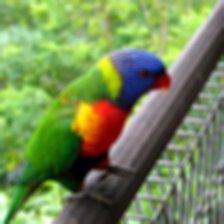}& \includegraphics[width=\imSz]{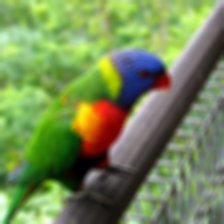}& \includegraphics[width=\imSz]{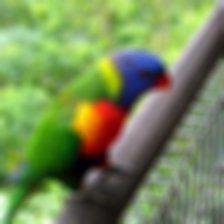}& \includegraphics[width=\imSz]{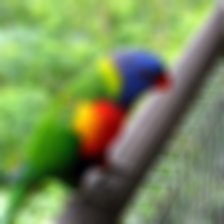}& \includegraphics[width=\imSz]{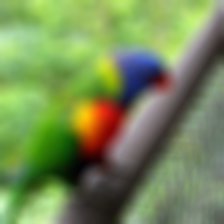}&

	 \includegraphics[width=\imSz]{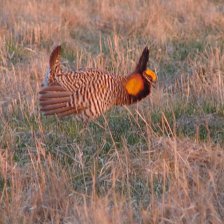}& 	 	\includegraphics[width=\imSz]{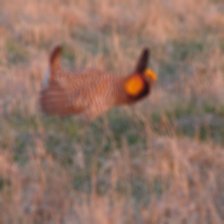}& \includegraphics[width=\imSz]{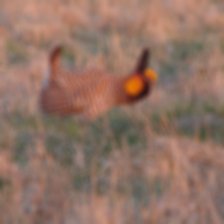}& \includegraphics[width=\imSz]{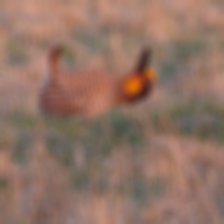}& \includegraphics[width=\imSz]{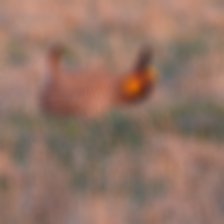}& \includegraphics[width=\imSz]{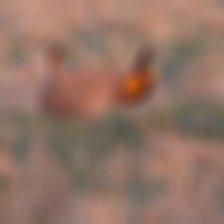}
	 	 
	\\	 	 
	 	 	 
	 \includegraphics[width=\imSz]{figures/dataset_tpami/clean/ILSVRC2012_val_00013739.JPEG} \sepSpace & 	 	 \includegraphics[width=\imSz]{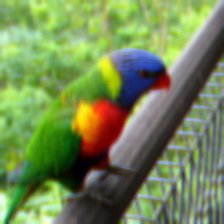}& \includegraphics[width=\imSz]{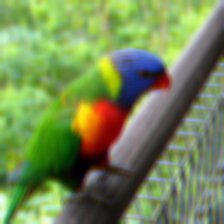}& \includegraphics[width=\imSz]{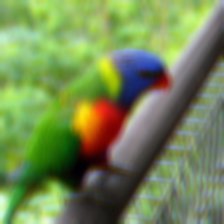}& \includegraphics[width=\imSz]{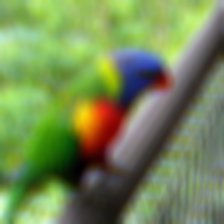}& \includegraphics[width=\imSz]{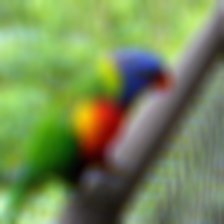}&

	 \includegraphics[width=\imSz]{figures/dataset_tpami/clean/ILSVRC2012_val_00036132.JPEG}& 	 		 \includegraphics[width=\imSz]{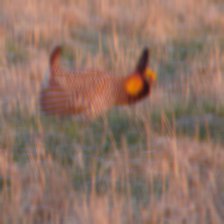}& \includegraphics[width=\imSz]{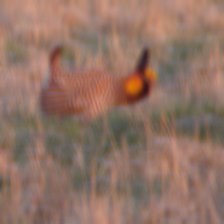}& \includegraphics[width=\imSz]{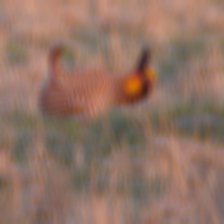}& \includegraphics[width=\imSz]{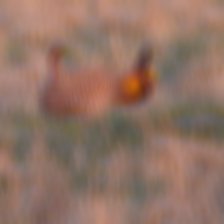}& \includegraphics[width=\imSz]{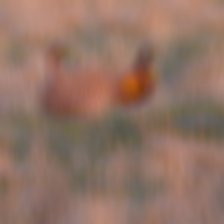}

	 \\  	 
	 
	 \includegraphics[width=\imSz]{figures/dataset_tpami/clean/ILSVRC2012_val_00013739.JPEG} \sepSpace & 	 	 \includegraphics[width=\imSz]{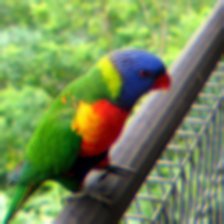}& \includegraphics[width=\imSz]{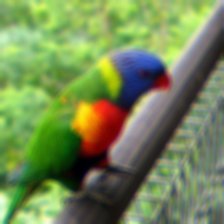}& \includegraphics[width=\imSz]{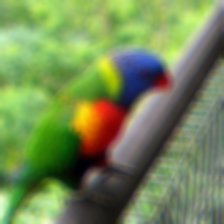}& \includegraphics[width=\imSz]{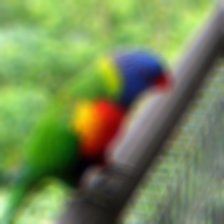}& \includegraphics[width=\imSz]{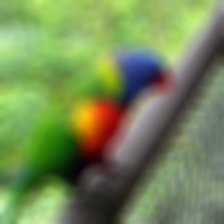}&
	 
	 \includegraphics[width=\imSz]{figures/dataset_tpami/clean/ILSVRC2012_val_00036132.JPEG}& 	 		 \includegraphics[width=\imSz]{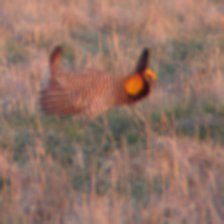}& \includegraphics[width=\imSz]{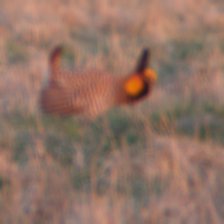}& \includegraphics[width=\imSz]{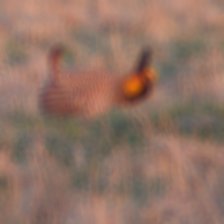}& \includegraphics[width=\imSz]{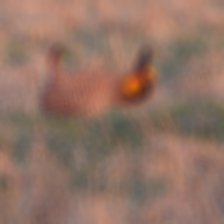}& \includegraphics[width=\imSz]{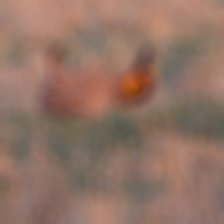}
	 
	 \\  
	 	 
	 \includegraphics[width=\imSz]{figures/dataset_tpami/clean/ILSVRC2012_val_00013739.JPEG} \sepSpace& 	 	 \includegraphics[width=\imSz]{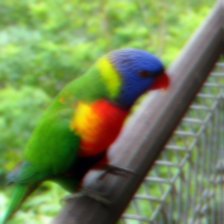}& \includegraphics[width=\imSz]{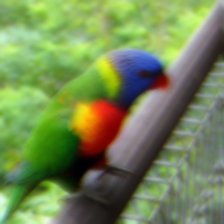}& \includegraphics[width=\imSz]{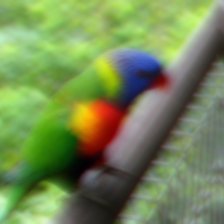}& \includegraphics[width=\imSz]{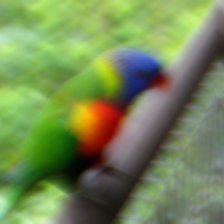}& \includegraphics[width=\imSz]{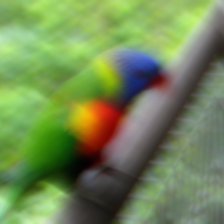}&

	 \includegraphics[width=\imSz]{figures/dataset_tpami/clean/ILSVRC2012_val_00036132.JPEG}& 	 		 \includegraphics[width=\imSz]{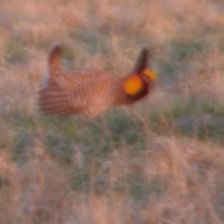}& \includegraphics[width=\imSz]{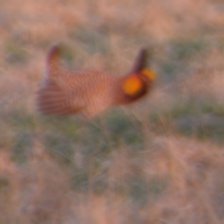}& \includegraphics[width=\imSz]{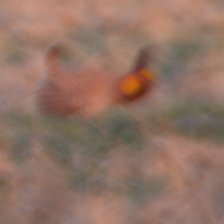}& \includegraphics[width=\imSz]{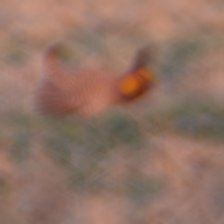}& \includegraphics[width=\imSz]{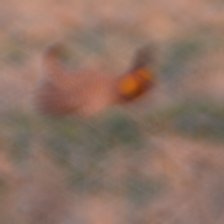}

	\\	 
	 	 
	 \includegraphics[width=\imSz]{figures/dataset_tpami/clean/ILSVRC2012_val_00013739.JPEG} \sepSpace& 	 	 \includegraphics[width=\imSz]{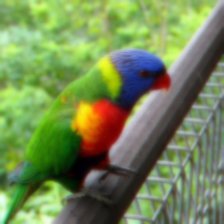}& \includegraphics[width=\imSz]{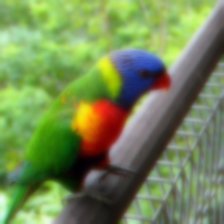}& \includegraphics[width=\imSz]{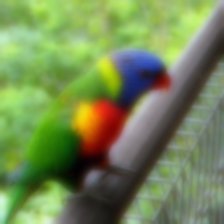}& \includegraphics[width=\imSz]{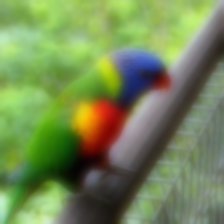}& \includegraphics[width=\imSz]{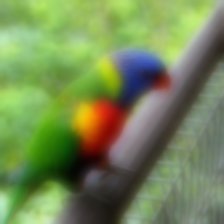}&
	 
	 \includegraphics[width=\imSz]{figures/dataset_tpami/clean/ILSVRC2012_val_00036132.JPEG}& 	 		 \includegraphics[width=\imSz]{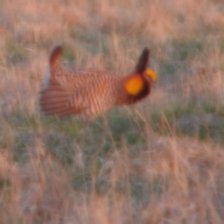}& \includegraphics[width=\imSz]{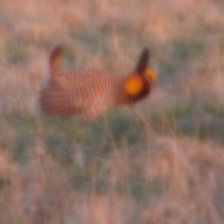}& \includegraphics[width=\imSz]{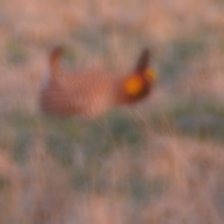}& \includegraphics[width=\imSz]{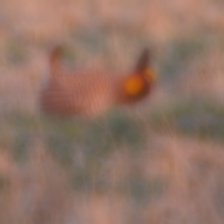}& \includegraphics[width=\imSz]{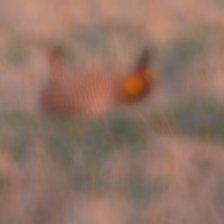}
	 	 
	\end{tabular} 

	};

	\path let \p1 = (ll), \p2 = (ur) in coordinate (ul) at (\x1,\y2);
	\def\offx{-.6} 
	\def\offy{-0.9}
	\def\szlbl{6.} 
	\draw[very thick,black,->] ($(ul) + (\offx,\offy)$) -- ($(ul) + (\offx,-\szlbl + \offy)$) node[midway,left=.25,rotate=90,anchor=center] (lbly) {aberration type};
	\draw[very thick,black,->] ($(ul) + (\offx,\offy)$) -- ($(ul) + (\offx+\szlbl,\offy)$) node[midway,above,anchor=south] (lblx) {severity};

	\def\offx{+1.15}
	\def\colorlbl{black}
	\node[\colorlbl,rotate=90] (a1) at ($(ll) !.43!(ul) + (\offx,0)$) {\scriptsize{ref.}};
	\node[\colorlbl,rotate=90] (a1) at ($(ll) !.33!(ul) + (\offx,0)$) {\scriptsize{astigm.}};
	\node[\colorlbl,rotate=90] (a2) at ($(ll) !.23!(ul) + (\offx,0)$) {\scriptsize{def.\& spher.}};
	\node[\colorlbl,rotate=90] (a3) at ($(ll) !.12!(ul) + (\offx,0)$) {\scriptsize{coma}};
	\node[\colorlbl,rotate=90] (a4) at ($(ll) !.04!(ul) + (\offx,0)$) {\scriptsize{trefoil}};

	\node[\colorlbl,rotate=90] (a1) at ($(ll) !.92!(ul) + (\offx,0)$) {\scriptsize{refer.}};
	\node[\colorlbl,rotate=90] (a1) at ($(ll) !.83!(ul) + (\offx,0)$) {\scriptsize{astigm.}};
	\node[\colorlbl,rotate=90] (a2) at ($(ll) !.72!(ul) + (\offx,0)$) {\scriptsize{def.\& spher.}};
	\node[\colorlbl,rotate=90] (a3) at ($(ll) !.62!(ul) + (\offx,0)$) {\scriptsize{coma}};
	\node[\colorlbl,rotate=90] (a4) at ($(ll) !.52!(ul) + (\offx,0)$) {\scriptsize{trefoil}};

	\end{tikzpicture}
	\end{adjustbox}
		
	\caption{\added{Four image examples from ImageNet-1k degraded by different OpticsBench blur types and severities. From left to right the severity increases from 1 to 5 and the aberration types are for each image from top to bottom reference blur~\cite{hendrycks_benchmarking_2019}, the optical blurs astigmatism, defocus \&  spherical, coma, trefoil from OpticsBench. The upper left example represents the astigmatism image corruption at severity 1. Best displayed in the digital version with zoom-in.}}\label{fig:opticsbench_in1k_image_examples}\end{figure*}
Fig.~\ref{fig:opticsbench_in1k_image_examples} shows examples for all OpticsBench image corruptions and four different images from ImageNet~\cite{deng_imagenet_2009}. 
While the differences between the image corruptions are visually subtle, the models confronted to these are challenged differently. 

\section{Additional evaluation and tables}
\label{app:evaluation_and_tables}
This appendix provides more results on OpticsBench and OpticsAugment for image classification on \added{ImageNet-100}, ImageNet-1k\added{, Flowers-102, Stanford Cars} and the extension to object detection on MSCOCO.

\subsection{OpticsBench (ImageNet-1k)}
\label{app:opticsbench_ranking}
Here, an overview is given for the average accuracy on OpticsBench on ImageNet-1k, followed by a discussion of the results for the Kendall tau $\tau_b$ rank correlation for different baselines.
\subsubsection{Overview}
\label{app:opticsbench_classification}
Table~\ref{tab:app_imagenet1k_overview} lists the accuracies on the ImageNet-1k validation dataset without image corruption and the corresponding average accuracy on OpticsBench and Table~\ref{tab:app_models_avg_rank_ob} lists the first 36/72 models for image classification ranked after their average accuracy on ImageNet-1k OpticsBench. Despite the low computational budget training regime used to train ResNet50 with OpticsAugment, which ranks the model 48/72 on the clean ImageNet dataset, the model is ranked at 20th place in Table~\ref{tab:app_models_avg_rank_ob} on OpticsBench. It is outperformed by stronger baselines such as Swin\_v2, ResNet152 or ConvNeXt and ResNet50+DeepAugment, which uses model specific information during training. Other models using data augmentation such as NoisyMix, AugMix and training on stylized ImageNet and ImageNet perform worse. 
\begin{table}[h]
    \centering
    \caption{ImageNet-1k accuracy on ImageNet-1k validation dataset and average accuracy on OpticsBench (OB).}
    \label{tab:app_imagenet1k_overview}
    \scriptsize
    \begin{tabular}{@{}c c c c@{}}
    DNN & Clean / OB & DNN & Clean / OB \\
    \hline 
    AlexNet & 52.9 / 13.8 & RegNet\_y\_3\_2gf & 78.6 / 45.8 \\ 
    ConvNeXt\_base & 80.3 / 48.5 & RegNet\_x\_400mf & 70.2 / 33.7 \\
    ConvNeXt\_large & 80.7 / 49.9 & RegNet\_x\_800mf & 73.1 / 32.8 \\
    ConvNeXt\_small & 79.4 / 47.2 & RegNet\_x\_8gf & 78.1 / 45.2 \\
    ConvNeXt\_tiny & 78.0 / 44.3 & RegNet\_y\_16gf & 79.6 / 47.0 \\
    DenseNet121 & 70.5 / 32.7 & RegNet\_y\_1\_6gf & 77.9 / 44.4 \\
    DenseNet161 & 73.8 / 38.8 & RegNet\_y\_32gf & 80.4 / 48.9 \\
    DenseNet169 & 72.5 / 36.0 & RegNet\_y\_3\_2gf & 78.6 / 45.8 \\
    DenseNet201 & 73.1 / 37.4 & RegNet\_y\_800mf & 75.7 / 39.4 \\
    EfficientNet\_b0 & 73.8 / 35.3 & RegNet\_y\_8gf & 79.5 / 47.7 \\
    EfficientNet\_b1 & 75.8 / 37.5 & ResNet101 & 78.7 / 48.3 \\
    EfficientNet\_b2 & 77.6 / 40.2 & ResNet152 & 79.3 / 50.1 \\
    EfficientNet\_b3 & 79.3 / 47.6 & ResNet18 & 66.4 / 26.3 \\
    EfficientNet\_b4 & 80.4 / 43.9 & ResNet34 & 69.9 / 33.0 \\
    EfficientNet\_b5 & 79.9 / 43.2 & ResNet50 & 76.9 / 41.0 \\
    EfficientNet\_b6 & 80.9 / 43.5 & ResNet50 (+OpticsAugm.) & 71.4 / 45.2 \\
    EfficientNet\_b7 & 80.0 / 41.1 & ResNeXt101 & 79.8 / 50.0 \\
    ResNet50 (+NoisyMix) & 74.6 / 44.1 & ResNeXt50 & 77.7 / 42.1 \\
    ResNet50 (+SIN\_IN\_IN) & 72.7 / 36.5 & ShuffleNet\_v2\_x0\_5 & 57.1 / 18.1 \\
    ResNet50 (+SIN\_IN) & 71.3 / 40.0 & ShuffleNet\_v2\_x1\_0 & 64.6 / 24.4 \\
    ResNet50 (+AugMix) & 74.4 / 43.7 & ShuffleNet\_v2\_x1\_5 & 69.0 / 31.5 \\
    ResNet50 (+DeepAug.) & 74.3 / 49.3 & ShuffleNet\_v2\_x2\_0 & 72.7 / 34.7 \\
    Inception\_v3 & 73.0 / 31.7 & SqueezeNet1\_0 & 54.5 / 13.6 \\
    MnasNet0\_5 & 63.8 / 22.6 & SqueezeNet1\_1 & 54.5 / 15.2 \\
    MnasNet0\_75 & 66.8 / 26.1 & Swin\_v2\_base & 80.3 / 49.0 \\
    MnasNet1\_0 & 68.6 / 27.1 & Vgg11 & 64.2 / 19.2 \\
    MnasNet1\_3 & 72.4 / 33.2 & Vgg11\_bn & 65.4 / 22.8 \\
    MobileNet\_v2 & 67.5 / 25.9 & Vgg13 & 64.5 / 19.4 \\
    MobileNet\_v3\_l & 71.7 / 32.3 & Vgg13\_bn & 66.2 / 22.0 \\
    MobileNet\_v3\_small & 63.7 / 25.9 & Vgg16 & 67.1 / 21.6 \\
    RegNet\_x\_16gf & 79.0 / 45.0 & Vgg16\_bn & 68.8 / 25.5 \\
    RegNet\_x\_1\_6gf & 74.8 / 37.9 & Vgg19 & 67.8 / 22.1 \\
    RegNet\_x\_32gf & 79.3 / 46.3 & Vgg19\_bn & 69.5 / 27.7 \\
    RegNet\_x\_3\_2gf & 77.4 / 43.6 & ViT\_b16 & 78.1 / 48.7 \\
    RegNet\_x\_400mf & 70.2 / 33.7 & ViT\_b & 74.7 / 45.4 \\
    RegNet\_x\_800mf & 73.1 / 32.8 & ViT\_l & 76.0 / 48.7 \\
    \hline
    $\Sigma$ & 73.0 / 36.8 \\ 
\end{tabular}
\end{table}

\begin{table}[h]
  \centering
  \caption{Models ranked according to their average accuracy on ImageNet-1k OpticsBench.}
  \label{tab:app_models_avg_rank_ob}
  \begin{tabular}{@{}c c c c@{}}
    \textbf{Rank} & \textbf{Model} & \textbf{Rank} & \textbf{Model} \\
    \hline
    1 & ResNet152 & 19 & RegNet\_x\_8gf \\
    2 & ResNeXt101 & \textit{20} & \textit{ResNet50 (+OpticsAugm.)} \\
    3 & Wide\_ResNet101\_2 & 21 & RegNet\_x\_16gf \\
    4 & ConvNeXt\_large & 22 & RegNet\_y\_1\_6gf \\
    5 & ResNet50(+DeepAug.) & 23 & ConvNeXt\_tiny \\
    6 & Swin\_v2\_base & 24 & ResNet50(+NoisyMix) \\
    7 & RegNet\_y\_32gf & 25 & EfficientNet\_b4 \\
    8 & ViT\_l & 26 & ResNet50(+AugMix) \\
    9 & ViT\_b16 & 27 & RegNet\_x\_3\_2gf \\
    10 & ConvNeXt\_base & 28 & EfficientNet\_b6 \\
    11 & ResNet101 & 29 & EfficientNet\_b5 \\
    12 & RegNet\_y\_8gf & 30 & wide\_ResNet50\_2 \\
    13 & EfficientNet\_b3 & 31 & ResNeXt50 \\
    14 & ConvNeXt\_small & 32 & EfficientNet\_b7 \\
    15 & RegNet\_y\_16gf & 33 & ResNet50 \\
    16 & RegNet\_x\_32gf & 34 & EfficientNet\_b2 \\
    17 & RegNet\_y\_3\_2gf & 35 & ResNet50(+SIN\_IN) \\
    18 & ViT\_b & 36 & RegNet\_y\_800mf \\
  \end{tabular}
\end{table}
\subsubsection{Additional analysis}
This paragraph provides a more detailed analysis of the model behaviour when confronted to optical aberrations. In Table~\ref{tab:classification_opticsbench_overview}  
Swin\_v2 and ConvNeXt have the highest clean accuracies and also the highest accuracies on OpticsBench with a reduction of \ensuremath{\SI{31.3}{\percent}} and \ensuremath{\SI{30.2}{\percent}}. Although  Swin\_v2 has the highest validation accuracy, ViT\_l has similar accuracy on OpticsBench and reduces only by \ensuremath{\SI{27.2}{\percent}}.

As shown in Fig.~\ref{fig:classification_opticsbench_imagnet1k_ranking4}, 
ResNet50+DeepAugment (g) outperforms the large ViT (a) on most of the image corruptions. Although using the DeepAugment scheme (g) results in remarkably good compensation for trefoil (green), astigmatism (light blue) and defocus \& spherical (orange), the coma image corruption remains constant. 
ResNet50+AugMix (n) and ResNet50+SIN\_IN (q) using joint training on stylized ImageNet and ImageNet improve slightly on all OpticsBench corruptions, and ResNet50+NoisyMix (k) improves on all but little on trefoil.
The WideResNet (b) and ConvNeXt (d) models have more than \ensuremath{\SI{5}{\percent}} lower  accuracy for astigmatism than the ViTs (a,c), but remain fairly stable on the others. ResNeXt50 (k) lacks in compensating for defocus \& spherical and trefoil.

The ResNet50+AugMix (i) seems to perform well on the baseline and trefoil, but SIN\_IN (i-1) compensates better for OpticsBench corruptions. Looking at the Swin transformer (g), it can compensate slightly better than ConvNeXt (h).
Wide ResNet (b) is ranked 3rd for defocus blur, but fails to compensate for the astigmatism corruption and defocus \& spherical, and EfficientNet\_b4 is less robust to astigmatism and defocus \& spherical.

When comparing OpticsBench and LensCorruptions lens aberrations, 
Large ViT is the most robust model on the ImageNet-100-OpticsBench dataset with an accuracy of \SI{56.13}{\percent}, but is clearly outperformed on LensCorruptions by many models, including the ConvNeXt models and EfficientNet\_b3. On OpticsBench ImageNet-1k, ViT\_l outperformed the ConvNeXt model at severity 4 for all OpticsBench corruptions with a baseline rank of 1 and 3 respectively, but is ranked on average after ConvNeXt\_l and Swin\_v2 in Table~\ref{tab:classification_opticsbench_overview}. 

\subsubsection{Kendall tau rank correlation}
\label{app:rank_correlation_matrix}
\begin{figure}[h]
    \centering
    \begin{tabular}{@{}c@{}c@{}}
    {\includegraphics[width=0.5\linewidth]{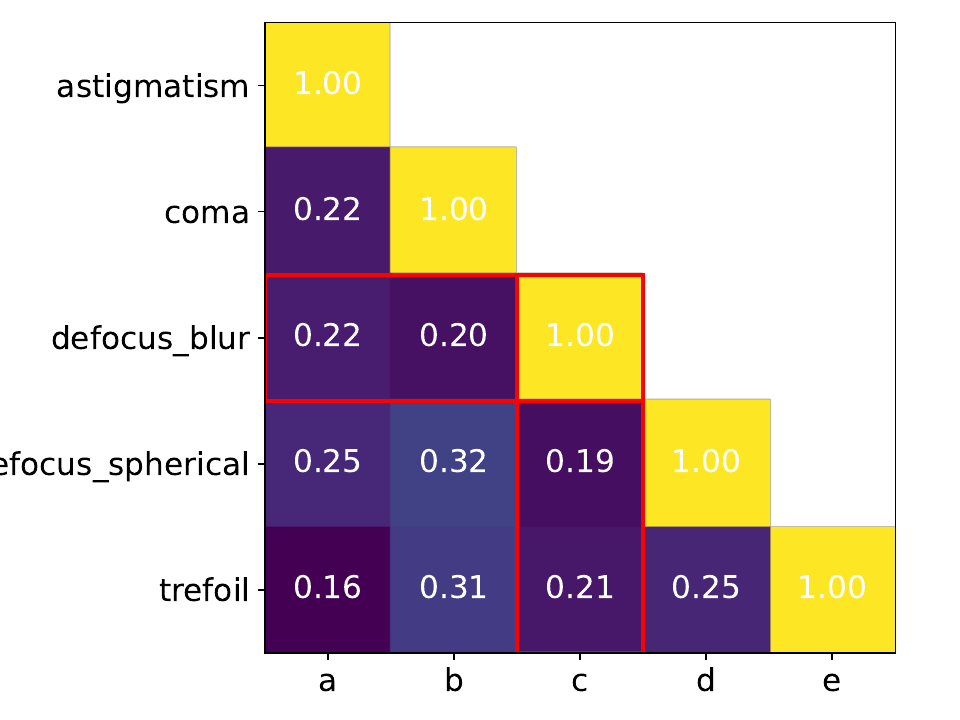}}&
     {\includegraphics[width=0.5\linewidth]{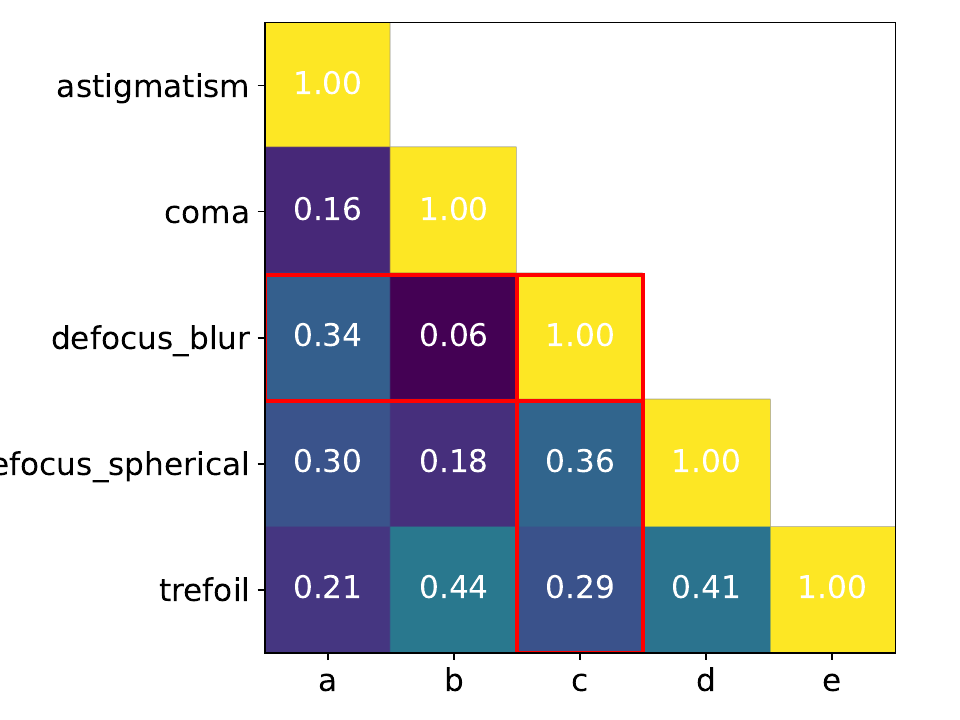}}\\
   \scriptsize{(a)} All models & \scriptsize{(b)} Robust models excluded  
    \end{tabular}
    \caption{Kendall tau rank correlation between all corruptions at severity 4. Since the ranking is symmetric, the upper triangle matrix is left blank intentionally. The corruptions are sorted in both directions in the same way.}
    \label{fig:opticsbench_baseline_corr}
\end{figure}

At first glance, the OpticsBench corruptions seem to be closely tied to each other, however this perspective is not true in general. When looking at different baselines, which is equivalent with flattening a specific corruption curve, other models will have ties to other corruptions. 
A general observation is that often, if a model is robust to a specific corruption it is as well to another. However, there is no consistent picture, which allows to use one corruption as proxy for the others. To further investigate the correlation between the models and corruptions, we report in Fig.~\ref{fig:opticsbench_baseline_corr} the various Kendall tau rank correlation coefficients, which give a measure, of how helpful a specific corruption ranking is as a proxy for the others. Each element of the matrix measures the correlation between two  image corruptions. Since the rank correlation is symmetric, we only show the lower triangular matrix. The elements on the diagonal showed auto rank correlation and have therefore $\tau_b=1$. The red labeled elements are the resulting correlations to the defocus blur baseline as reported in the main paper at Fig.~\ref{fig:classification_opticsbench_imagnet1k_ranking4}. If instead \eg trefoil is used as the baseline (last row), the correlation with astigmatism drops to 0.16 and rises to 0.31 for coma, remains at 0.21 for defocus blur due to symmetry, and rises to 0.25 for defocus \& spherical. 
Fig.\ref{fig:opticsbench_baseline_corr}b shows the correlation, when all robust models are excluded from the computation. The rank correlation rises for most of the constellations. 
However, we find the rank correlation alone not predictive, since the corruptions could proceed at a different level, leading to highly different accuracies. So, it is important to take also the accuracies into account.

\subsubsection{Ranking results for more severities}
\begin{figure*}[b]
    \centering
    \begin{tabular}{@{}l@{}r@{}}
    \rotatebox{90}{\phantom{blablibluss}{Severity 1}}&
    \includegraphics[width=0.9\linewidth]{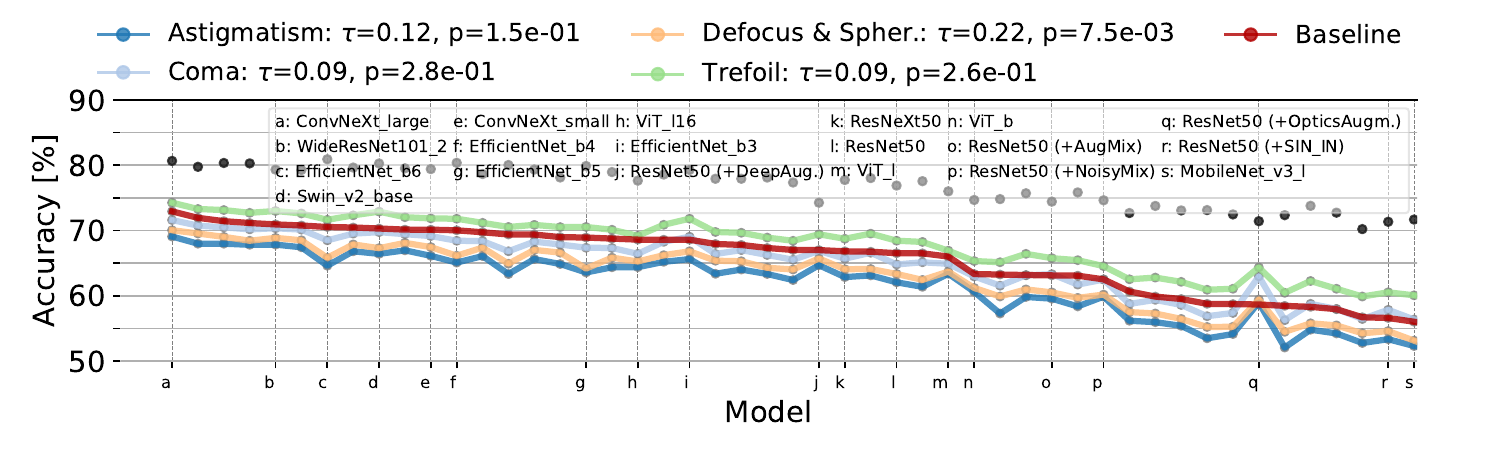}
    \end{tabular}
    \caption{Ranking on OpticsBench-ImageNet-1k and baseline defocus\_blur for severity 1. The black dots represent the validation accuracies. The validation accuracies differ in about \SI{10}{\percent} points and the accuracies on the different image corruptions for severity 1 differ by about \SI{20}{\percent}. This gap becomes larger towards higher severities.}
    \label{fig:app_ranking_plots}
\end{figure*}

\begin{figure*}
    \setcounter{subfigure}{2} 
    \centering
     \begin{tabular}{@{}l@{}r@{}}
   \rotatebox{90}{\phantom{blablibluss}{Severity 2}}&
    \includegraphics[width=0.9\linewidth]{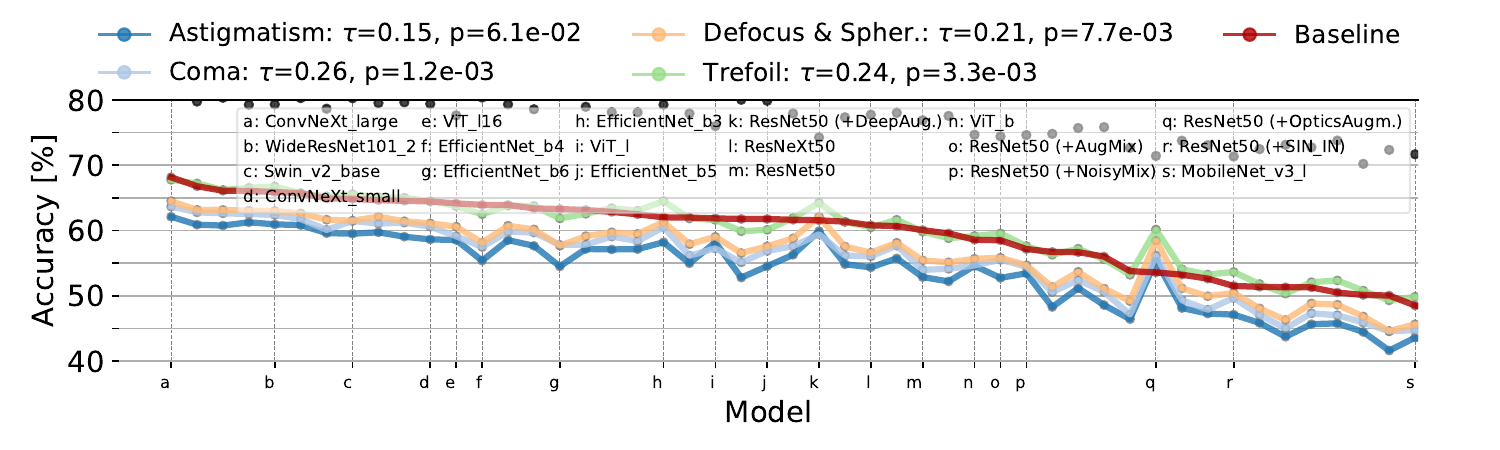}
    \\
    \rotatebox{90}{\phantom{blablibluss}{Severity 3}}&
    \includegraphics[width=0.9\linewidth]{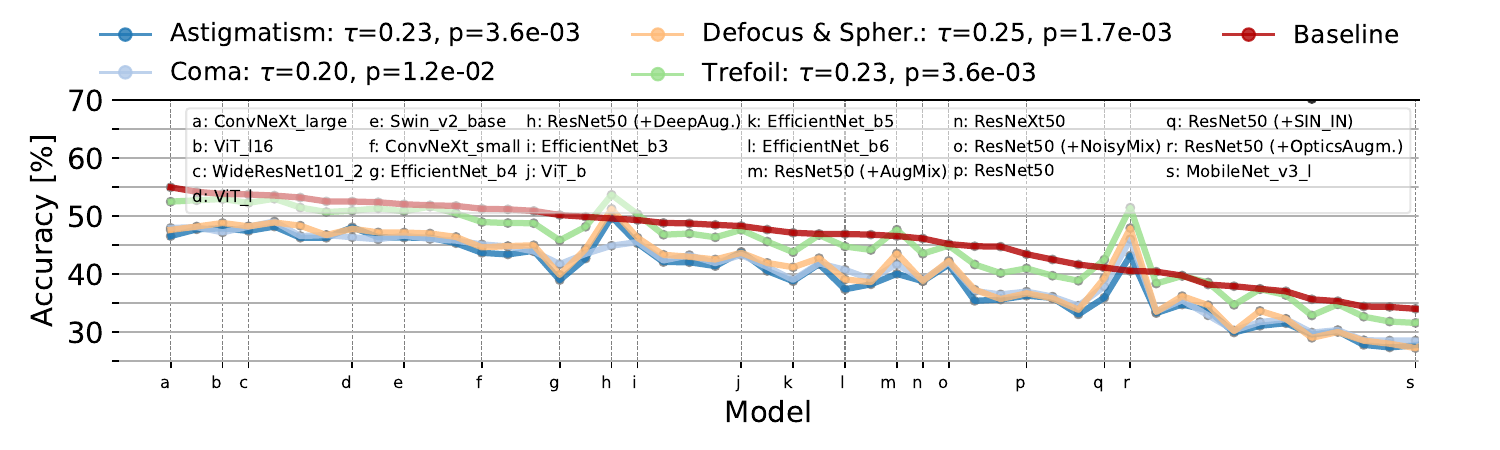}
    \\
     \rotatebox{90}{\phantom{blablibluss}{Severity 4}}&
    \includegraphics[width=0.9\linewidth]{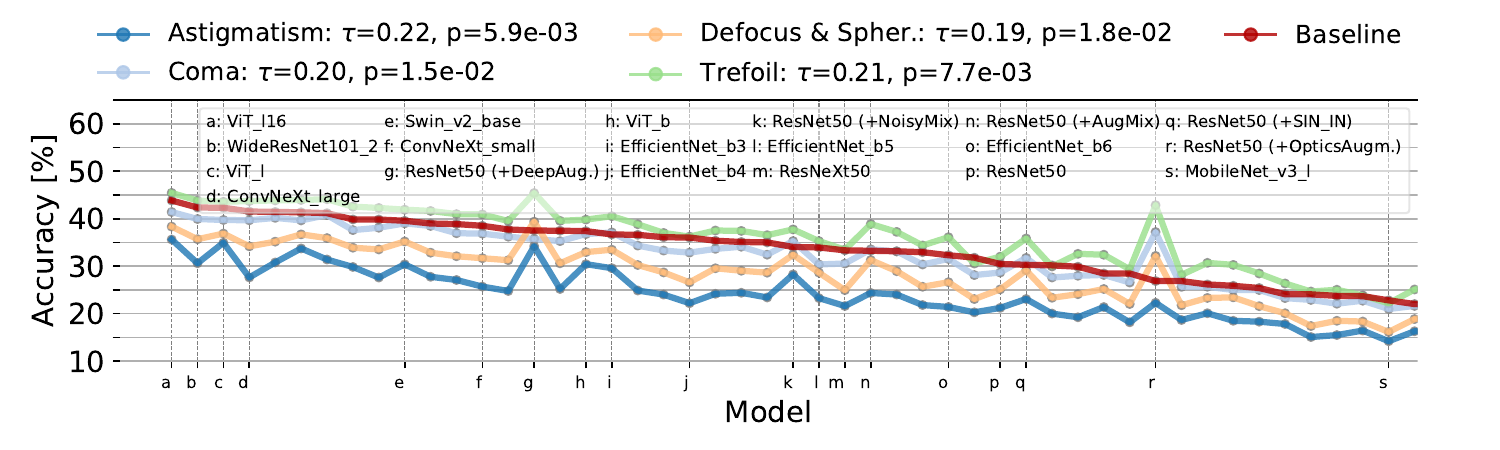}
    \\
     \rotatebox{90}{\phantom{blablibluss}{Severity 5}}&
    \includegraphics[width=0.9\linewidth]{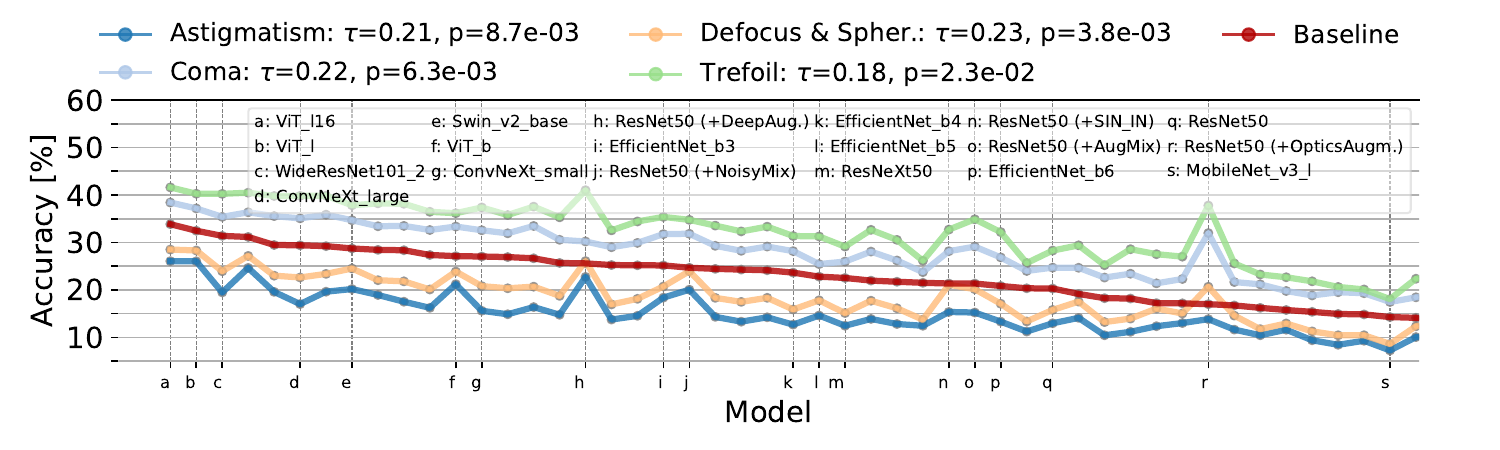}
 \end{tabular}
    \caption{Ranking on OpticsBench-ImageNet-1k and baseline defocus\_blur for all severities 2,3, 4 and 5.}
    \label{fig:app_ranking_plots_45}
\end{figure*}

Figs.~\ref{fig:app_ranking_plots} and~\ref{fig:app_ranking_plots_45} show the ranking of the 72 image classification models with the baseline defocus blur~\cite{hendrycks_benchmarking_2019} image corruption for all five severities on the ImageNet-1k OpticsBench.
In general, as the severity increases in Figs.~\ref{fig:app_ranking_plots} and~\ref{fig:app_ranking_plots_45}, the peak differences to the baseline become more pronounced for many (robust) models such as for the ResNet50+DeepAugment for severity 5 in Fig.~\ref{fig:app_ranking_plots_45} compared to severity 1 in Fig.~\ref{fig:app_ranking_plots}. Also, some of the models are re-sorted with varying severity: while ConvNeXt large is ranked at first place for severities 1-3 it is replaced by the ViT for severities 4 and 5. 

Subsequently, the ranking results for the different severities are discussed in more detail. The black dots in Fig.~\ref{fig:app_ranking_plots} show the clean accuracy on the validation dataset. In general, the initial accuracy is roughly predictive for the ranking for severities 1 and 2. However, compared to the baseline ranking there are some deviations for other image corruptions. Although EfficientNet\_b6 (c) has slightly higher validation accuracy than the neighbors including Wide ResNet101 (b) and Swin\_v2 (d), the accuracy for astigmatism and defocus \& spherical drops significantly compared to the baseline and validation accuracy. EfficientNet\_b3 (i) conversely performs better on trefoil compared to its assigned baseline rank. 

For severity 2 in Fig.~\ref{fig:app_ranking_plots_45} there are more pronounced deviations from the baseline rank compared to severity 1. Again, Efficient\_b4 (f) has lower accuracy for most of the OpticsBench corruptions compared to its neighbors. The large ViT (i) performs overly well on astigmatism and the Deep Augment (k) training of ResNet50 results in significantly higher accuracy on all OpticsBench image corruptions compared to the similarly ranked models. However, the OpticsAugment model (q) improves largely on the OpticsBench image corruptions. 
For severity 3 in Fig.~\ref{fig:app_ranking_plots_45} the ResNet50+DeepAugment (h) compensates well for most of the image corruptions, but fails to compensate for coma. OpticsAugment then outperforms the DeepAugment model for the coma and trefoil image corruption. In addition, for many models, there are variations in accuracy for the various image corruptions compared to the baseline rank. 

We also repeat severity 4 in Fig.~\ref{fig:app_ranking_plots_45} for direct comparison with the other severities. The results have already been discussed in subsection~\ref{subsec:imagenet1k_opticsbench}: ResNet50+DeepAugment (g) outperforms the large ViT (a) on most of the image corruptions. Although, using the DeepAugment scheme (g) results in  remarkably good compensation for trefoil (green), astigmatism (light blue) and defocus \& spherical (orange), the coma image corruption remains constant. ResNet50+AugMix (n) and ResNet50+SIN\_IN (q) using joint training on stylized ImageNet and ImageNet improve slightly on all OpticsBench corruptions, and ResNet50+NoisyMix (k) improves on all but little on trefoil.
The WideResNet (b) and ConvNeXt (d) models have more than \SI{5}{\percent} lower  accuracy for astigmatism than the ViTs (a,c), but remain fairly stable on the others. ResNeXt50 (k) lacks in compensating for defocus \& spherical and trefoil.

Severity 5 in Fig.~\ref{fig:app_ranking_plots_45} shows that the WideResNet101 (c) achieves high accuracies for most of the image corruptions, but does not compensate that well for astigmatism. The ResNet50+DeepAugment (h) again largely improves on most of the image corruptions, but fails in compensating for coma at the same scale. For this image corruption, it is outperformed by all higher ranked models as well as by AugMix (o) and largely by OpticsAugment (r).

\subsection{OpticsBench (Reflective Boundaries)}
\label{app:subsec:in1k_ob_reflective}
\begin{figure*}[p]
    \setcounter{subfigure}{2} 
    \centering
     \begin{tabular}{@{}l@{}r@{}}
   \rotatebox{90}{\phantom{blablibluss}{Severity 1}}&
    \includegraphics[trim=0cm .87cm 0cm 0cm,clip,width=0.93\linewidth]{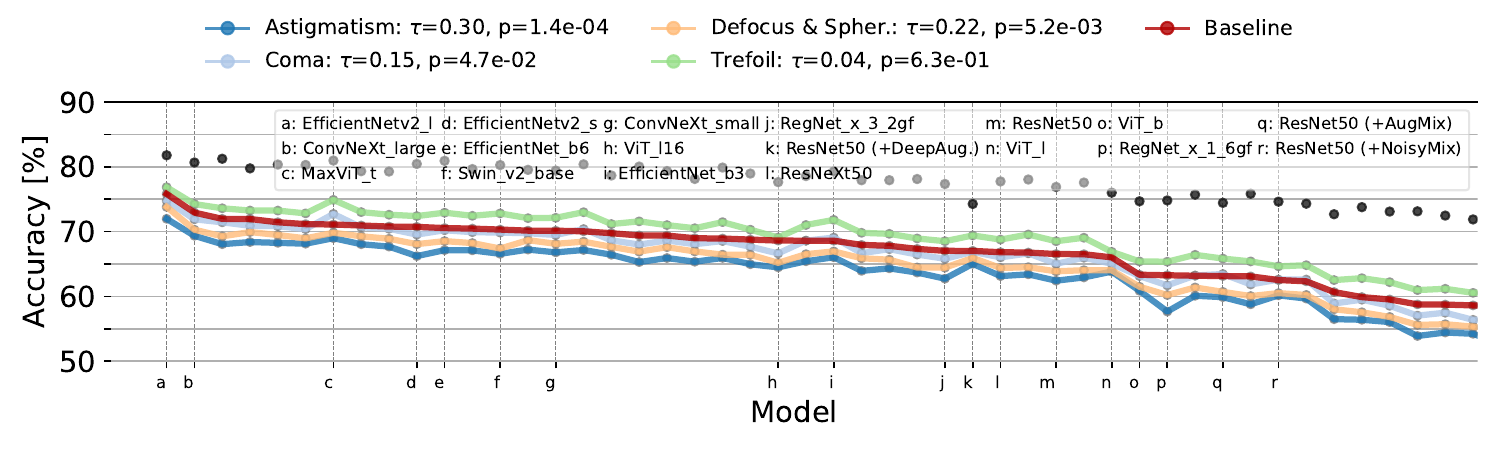}
    \\
    \rotatebox{90}{\phantom{blablibluss}{Severity 2}}&
    \includegraphics[trim=0cm .87cm 0cm 0cm,clip,width=0.93\linewidth]{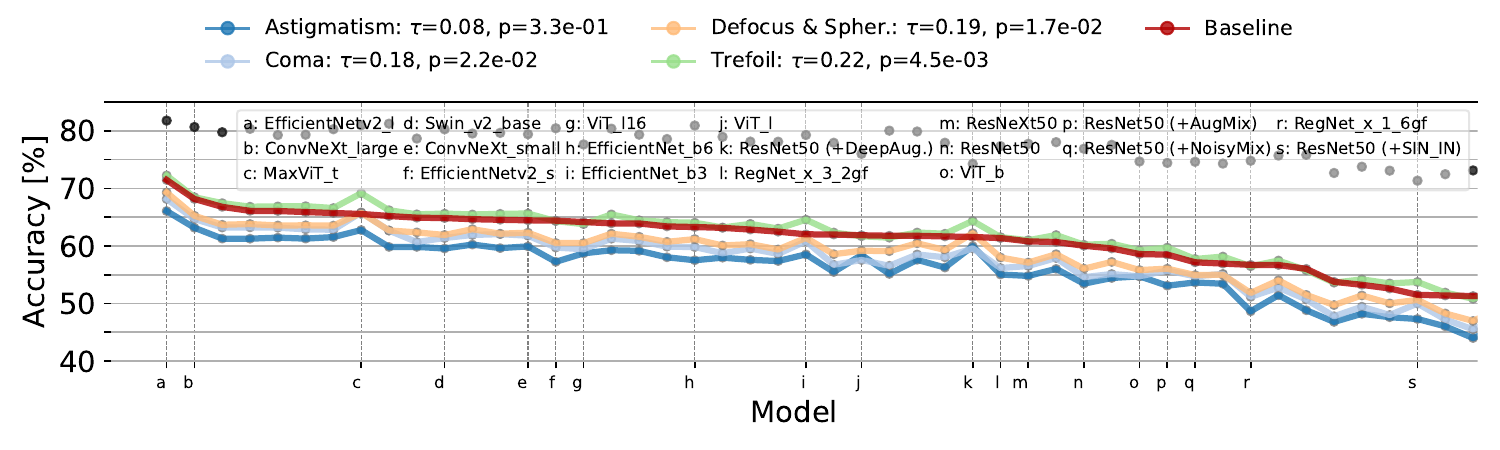}
    \\
     \rotatebox{90}{\phantom{blablibluss}{Severity 3}}&
    \includegraphics[trim=0cm .87cm 0cm 0cm,clip,width=0.93\linewidth]{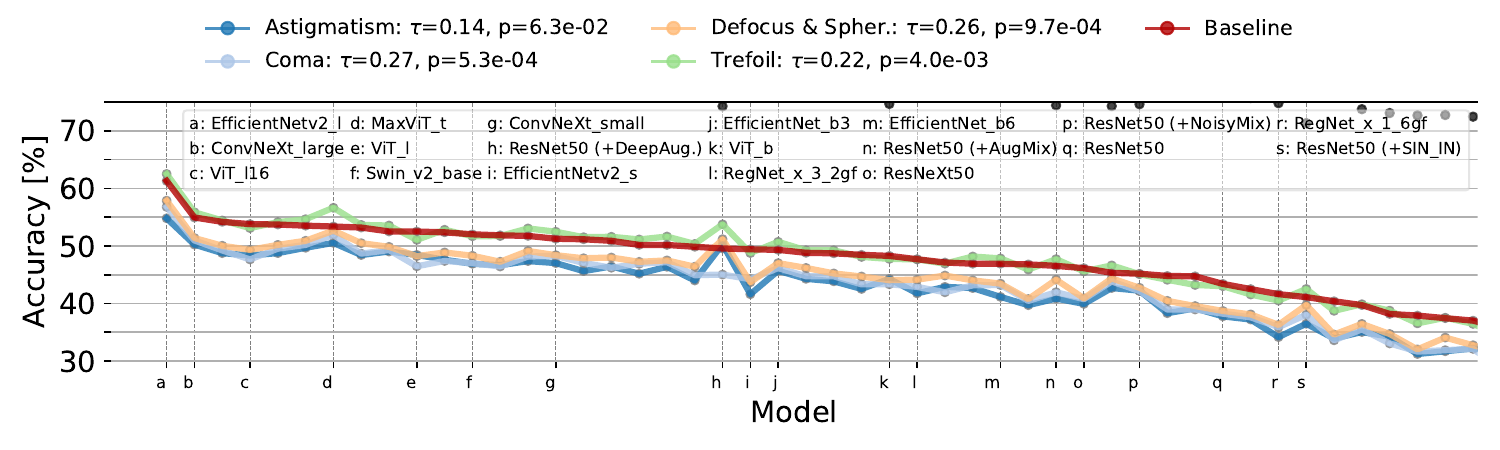}
    \\
     \rotatebox{90}{\phantom{blablibluss}{Severity 4}}&
    \includegraphics[trim=0cm .87cm 0cm 0cm,clip,width=0.93\linewidth]{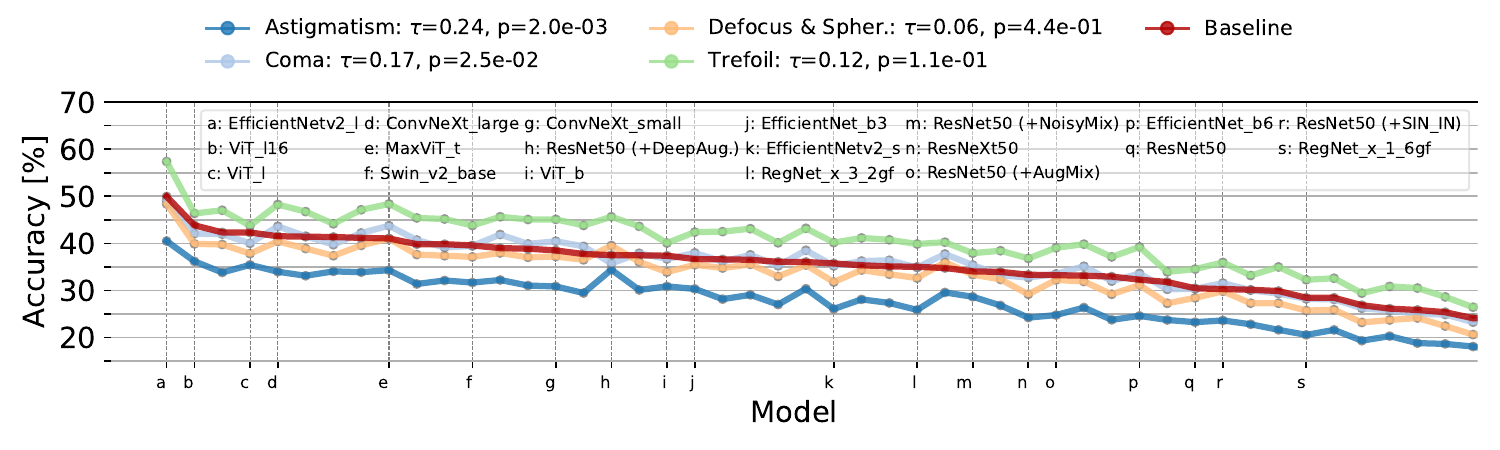}
     \\
     \rotatebox{90}{\phantom{blablibluss}{Severity 5}}&
    \includegraphics[trim=0cm .87cm 0cm 0cm,clip,width=0.93\linewidth]{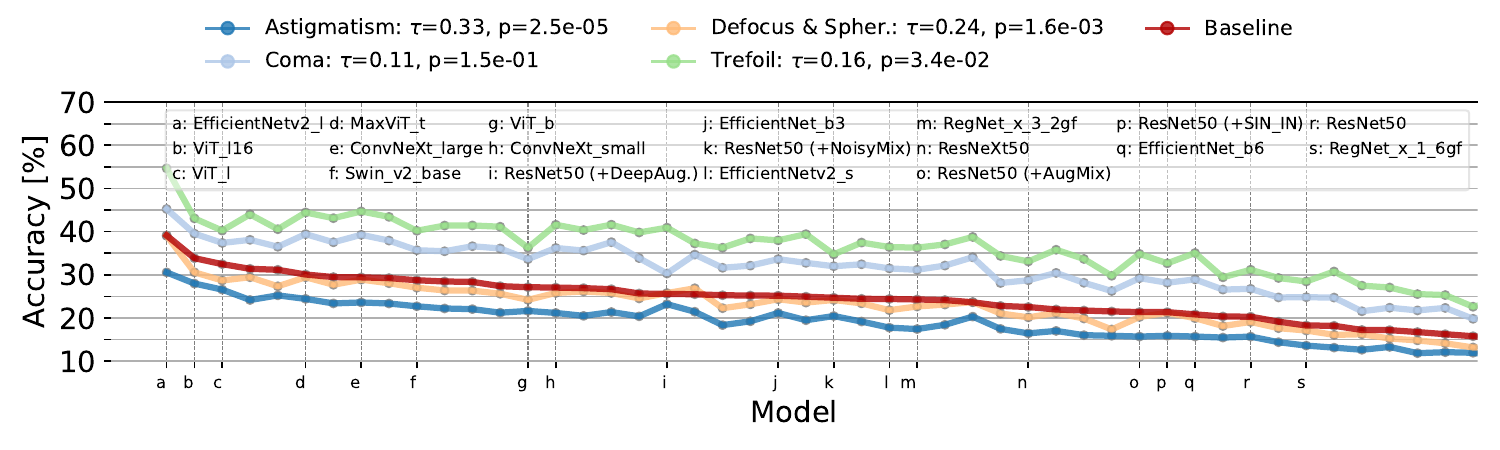}
 \end{tabular}
 \caption{Ranking on OpticsBench-ImageNet-1k (reflective boundaries) and baseline defocus\_blur for severities 1-5 in (a-e) showing 50 image classification models. There are significant deviations from the reference (red) for certain models and corruptions (\eg, MaxViT on trefoil, DeepAugment on astigmatism).}\label{fig:in1k_opticsbench_without_guides_von_neumann} 	
\end{figure*}

Here, we provide an additional analysis on OpticsBench using reflective boundaries to generate the OpticsBench image corruptions. Fig.~\ref{fig:in1k_opticsbench_without_guides_von_neumann} shows a large number (50) of image classification models evaluated on OpticsBench-ImageNet-1k and the reference blur. Each subfigure represents results for a specific severity, where the accuracies are sorted with the baseline (defocus blur~\cite{hendrycks_benchmarking_2019}). 
Similar observations as for the zero-padded version hold true. For instance, Deep Augment training still helps for most of the corruptions, except for coma. However, the intensity of the fluctuations is less pronounced. 
The EfficientNet\_v2 large model has the highest accuracy across severities. The model has roughly 118M parameters and was trained for 600 epochs with several standard augmentations. 
These factors contribute to the high overall accuracy.
\def\figsize{.93\linewidth}
\vspace{-1em}
\subsection{OpticsBench - Flowers-102 and Stanford Cars}
\label{app:subsec:results_ob_flowers_and_cars}
\begin{figure}[!htb]
    \begin{tabular}{c}
    \includegraphics[width=\linewidth]{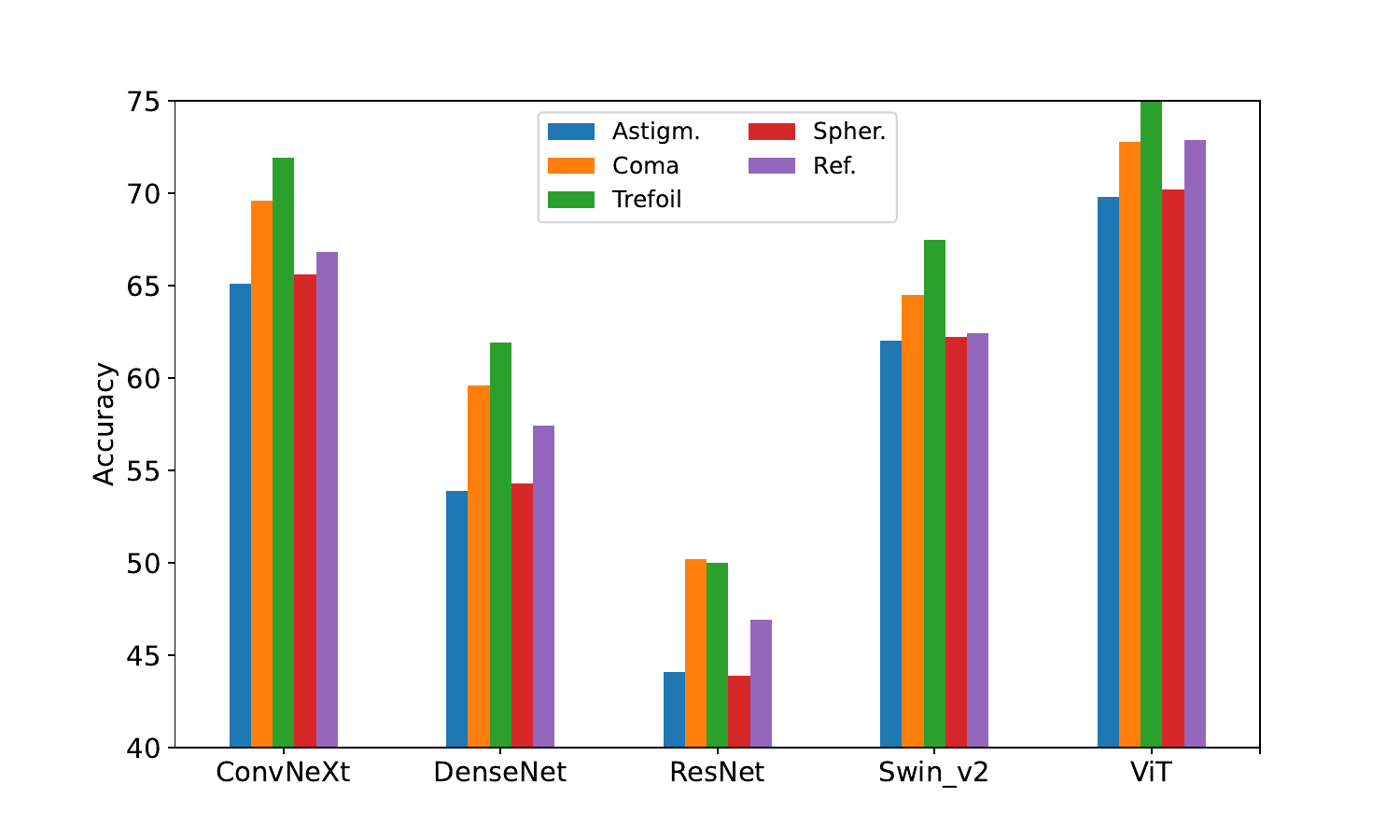}
           \\
           (a) \\ 
    \includegraphics[width=\linewidth]{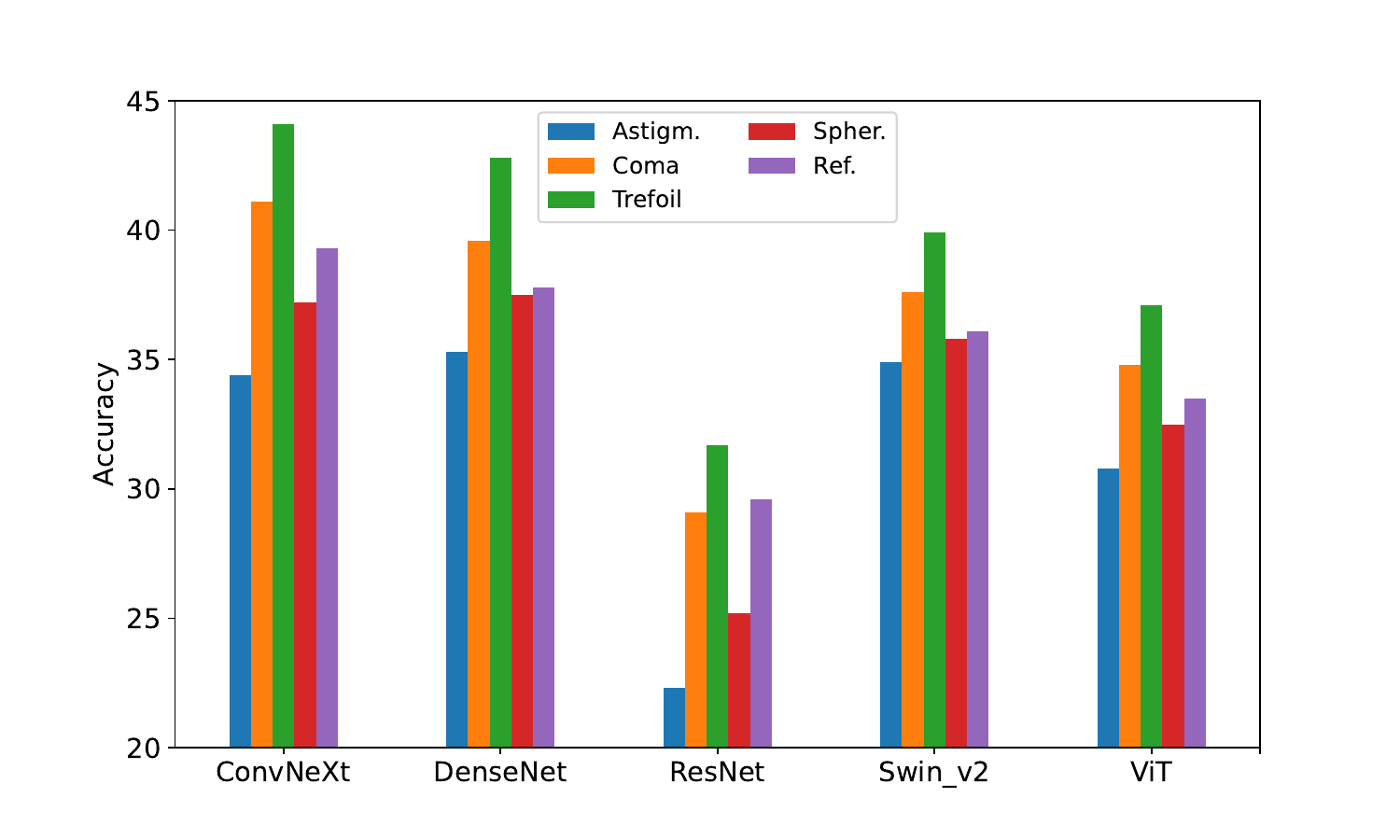}
          \\ 
          (b)
    \end{tabular}
    \centering
    \caption{\added{OpticsBench: Flowers-102 (a) and Stanford Cars (b). Average accuracy across severities for multiple fine-tuned models on fine-grained image datasets for OpticsBench image corruptions and the reference (Ref.).}} \label{fig:app_results_flowers_and_cars_vs_cor}
\end{figure}

\begin{figure}[!tb]
    \begin{tabular}{c}
    \includegraphics[width=\linewidth]{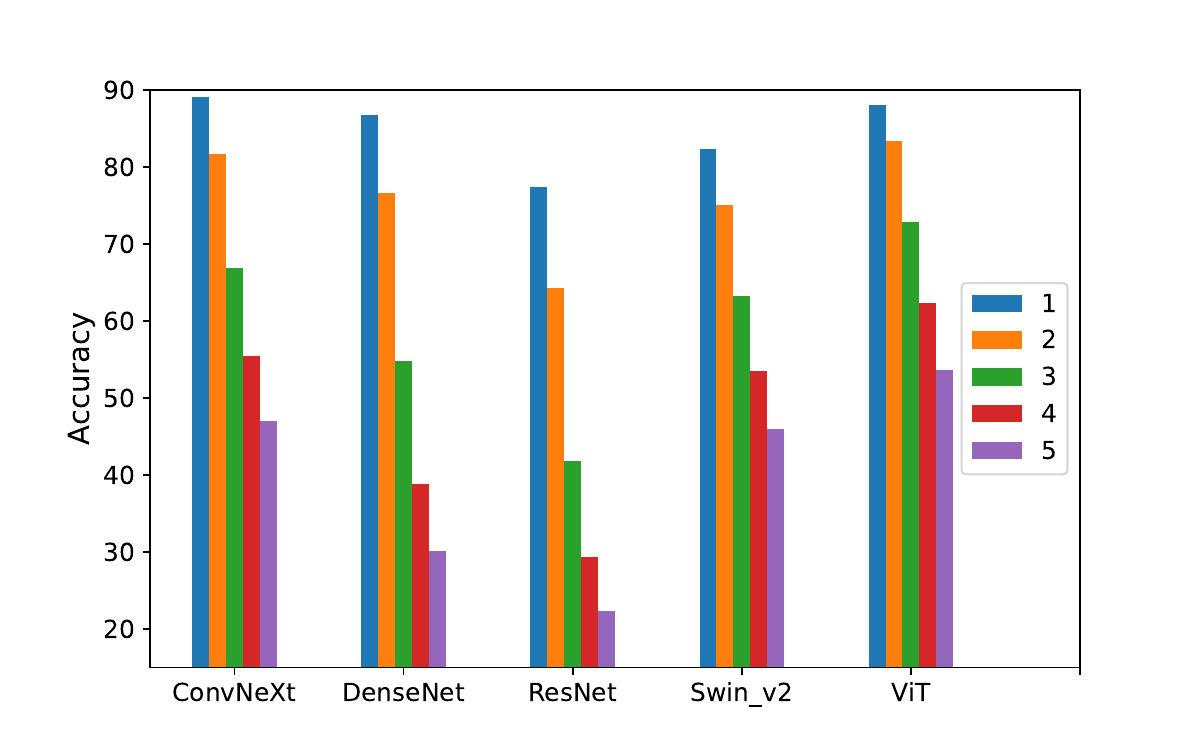}
           \\
           (a) \\ 
    \includegraphics[width=\linewidth]{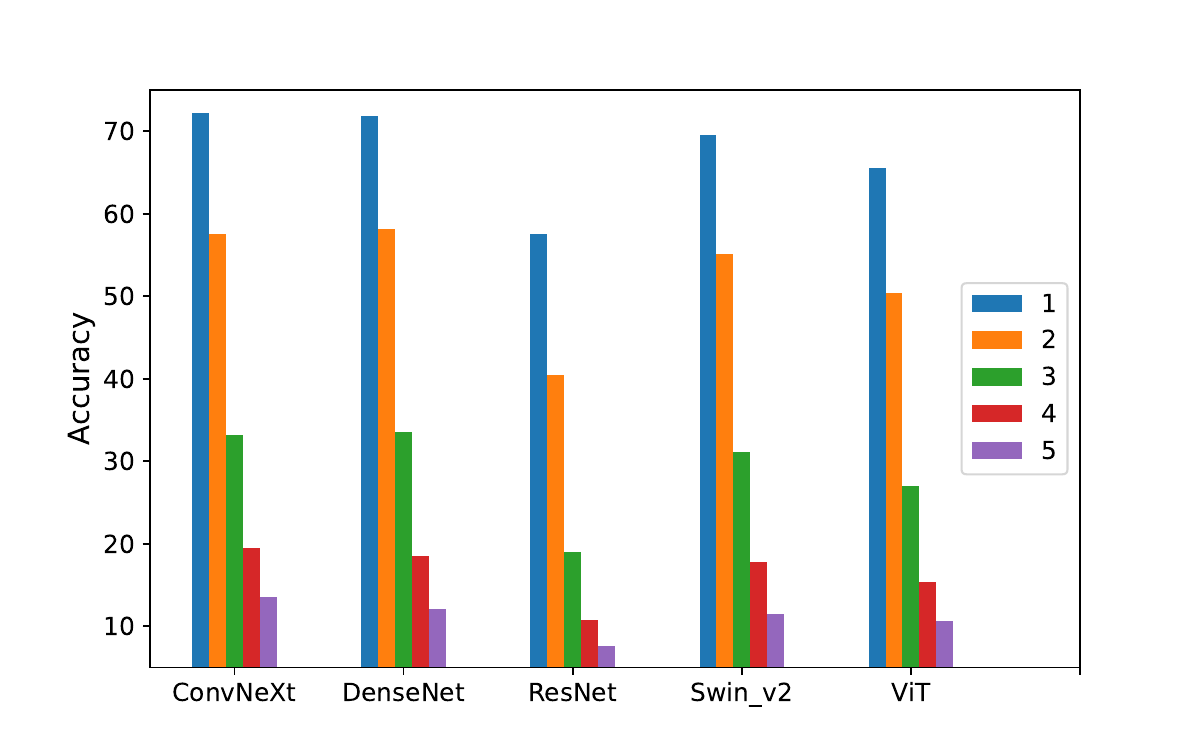}
          \\ 
          (b)
    \end{tabular}
    \centering
    \caption{\added{OpticsBench: Flowers-102 (a) and Stanford Cars (b). Average accuracy across corruptions for different severities (1-5) and multiple fine-tuned models on fine-grained image datasets.}}
    \label{fig:app_results_flowers_and_cars_vs_sev}
\end{figure}

\added{We continue here the analysis of the results on OpticsBench for the Flowers-102~\cite{nilsbackAutomatedFlowerClassification2008} and Stanford Cars~\cite{krause3DObjectRepresentations2013} datasets. Fig.~\ref{fig:app_results_flowers_and_cars_vs_cor} shows the average accuracy for OpticsBench applied to Flowers-102 (a) and Stanford Cars (b) for several diverse image classification models for each image corruption. The models are challenged differently, such that Swin\_v2 is having similar accuracy on spherical (red) and the reference (purple) image corruptions, while ResNet has significantly lower accuracy for astigmatism and spherical with respect to the reference. Fig.~\ref{fig:app_results_flowers_and_cars_vs_sev} shows for the same models and OpticsBench image corruptions the results per severity.}

\subsection{OpticsBench (MSCOCO)} 
\label{app:mscoco_rankings}
\label{app:mscoco_rankings_a}
\subsubsection{TIDE errors}
Tab.~\ref{tab:app_tide_metrics_overview} lists the six main errors as defined by Bolya et al. in~\cite{bolya_tide_2020} on the validation dataset of MSCOCO and for OpticsBench (second column) the average over all corruptions and severities. While the classification error (Cls) doubles from 2.7 to 5.3, the localization errors (Loc) decrease from an average 6.4 to 5.6.
Both and duplicate (Dupe) detection errors remain fairly constant at a low level, background (Bkg) errors halve on average from 4.0 to 2.2, and missing detections increase from 6.0 to 14.4. 
\begin{table*}[h]
    \centering
    \footnotesize
    \caption{Evaluated DNNs on MSCOCO~\cite{fleet_microsoft_2014} validation dataset (Clean) and OpticsBench (OB). We report here the different errors from evaluation with TIDE~\cite{bolya_tide_2020}. The first value is the clean value and the second an average on OpticsBench.}  
    \begin{tabular}{l c c c c c c}
        DNN / Backbone & Cls & Loc & Both & Dupe & Bkg & Miss \\
        \hline
Casc. Mask R{-}CNN (ConvNeXt\_s) & 2.3 / 5.2 & 6.2 / 6.1 & 1.1 / 1.2 & 0.1 / 0.1 & 4.2 / 2.2 & 6.3 / 18.5 \\ Casc. R{-}CNN (ResNet50) & 4.0 / 7.3 & 6.9 / 5.0 & 1.2 / 1.3 & 0.2 / 0.1 & 4.1 / 1.8 & 8.7 / 18.3 \\ Def. DETR (ResNet50) & 2.2 / 5.6 & 6.3 / 4.8 & 1.1 / 1.4 & 0.5 / 0.6 & 4.5 / 2.6 & 4.0 / 8.8 \\ DINO (Swin\_l) & 1.2 / 2.8 & 4.8 / 6.3 & 0.9 / 1.0 & 0.5 / 0.7 & 4.1 / 2.7 & 2.6 / 9.1 \\ Faster R{-}CNN (ResNet50) & 3.8 / 7.3 & 7.0 / 5.1 & 1.2 / 1.4 & 0.3 / 0.2 & 4.0 / 1.7 & 8.0 / 16.1 \\ Mask R{-}CNN (Swin\_s) & 2.1 / 4.5 & 6.7 / 6.2 & 1.1 / 1.2 & 0.1 / 0.1 & 3.8 / 1.9 & 6.8 / 19.8 \\ RetinaNet (Eff.Net\_b3) & 2.7 / 4.5 & 7.0 / 5.5 & 1.2 / 1.5 & 0.3 / 0.2 & 3.7 / 2.2 & 5.4 / 9.8 \\ YOLOX\_x (DarkNet) & 3.2 / 5.6 & 6.4 / 5.8 & 1.1 / 1.4 & 0.2 / 0.2 & 3.9 / 2.5 & 6.4 / 14.8 \\ 
\hline
$\Sigma$ & 2.7 / 5.3 & 6.4 / 5.6 & 1.1 / 1.3 & 0.3 / 0.3 & 4.0 / 2.2 & 6.0 / 14.4 
    \end{tabular}
    \label{tab:app_tide_metrics_overview}
\end{table*}

\subsubsection{Additional analysis}
In Fig.~\ref{fig:cls_and_loc} (top) for some models the Cls error depends on the image corruption, while for others it is quite similar. YOLO (c), Mask R-CNN (f) and in particular DINO (h) have similar Cls errors for all image corruptions. For the Faster R-CNN (a), Cascade R-CNN (b) and Deformable DETR(e) the Cls errors are more different for the various image corruptions. 
The Cls errors of Faster R-CNN (a) and Cascade R-CNN (b) are highest for astigmatism and coma (light blue), while the baseline (red) has the lowest Cls error. Compared to Deformable DETR (e), the Cascade Mask R-CNN (d) has a lower Cls error on trefoil (green) and defocus \& spherical (orange), but a higher Cls error on the baseline.

In Fig.~\ref{fig:cls_and_loc} (bottom) the coma corruption (light blue) has the highest Loc errors for half of the models (a-d), while the distance to the baseline becomes negligible for lower ranked models (e-h). The astigmatism (blue) fluctuates around the baseline (red) and tends to generate the lowest Loc errors compared to other corruptions. Trefoil (green) and defocus \& spherical (orange) have for most models the lowest Loc errors. The specific behavior depends on the model: The DINO model (a) has a high Loc error for the coma image corruption, a similar Loc error for astigmatism and the baseline and a comparably low Loc error for defocus \& spherical and trefoil. Compared to the initial Loc error of 4.8 in Table~\ref{tab:mscoco_ob_dnn_overview} there is a high increase in Loc error for the DINO model, which is especially true for the coma image corruption. The deviation from the baseline across different image corruptions is large for DINO, while the Loc errors for RetinaNet (e) follow the baseline closely.
Deformable DETR (h) has the lowest Loc errors, but a high Cls error and a low mAP in Table~\ref{tab:mscoco_ob_dnn_overview}. 
Mask R-CNN (b) and the YOLO (c) model have a higher Loc error on the baseline than on astigmatism, but YOLO has lower Loc error for trefoil and defocus \& spherical.

\subsection{OpticsAugment}
\label{app:optics_augment}
This appendix discusses more results for the OpticsAugment data augmentation on ImageNet-100 and MSCOCO image corruptions. Tables~\ref{tab:tab:imagenet100_corruptions_DenseNet_revisited} and~\ref{tab:tab:imagenet100_corruptions_ResNeXt50_revisited} list the accuracies for DenseNet and ResNeXt50 with and without OpticsAugment training on ImageNet-100 OpticsBench.
\begin{table*}[htb]
\centering
\caption{Accuracies for DenseNet w/wo OpticsAugment evaluated on ImageNet-100 OpticsBench.}\label{tab:tab:imagenet100_corruptions_DenseNet_revisited}
\begin{tabular}{llll|lll|lll|lll|lll}&\multicolumn{3}{c}{\scriptsize{1}}&\multicolumn{3}{c}{\scriptsize{2}}&\multicolumn{3}{c}{\scriptsize{3}}&\multicolumn{3}{c}{\scriptsize{4}}&\multicolumn{3}{c}{\scriptsize{5}}\\\tiny{Corruption} & \tiny{clean} & \tiny{\textbf{ours}} & \tiny{$\Delta$} & \tiny{clean} & \tiny{\textbf{ours}} & \tiny{$\Delta$} & \tiny{clean} & \tiny{\textbf{ours}} & \tiny{$\Delta$} & \tiny{clean} & \tiny{\textbf{ours}} & \tiny{$\Delta$} & \tiny{clean} & \tiny{\textbf{ours}} & \tiny{$\Delta$}\\\hline\tiny{astigmatism} & \tiny{50.26} & \textbf{\tiny{67.68}} & \tiny{17.42} & \tiny{39.54} & \textbf{\tiny{65.72}} & \tiny{26.18} & \tiny{26.80} & \textbf{\tiny{57.02}} & \tiny{30.22} & \tiny{16.22} & \textbf{\tiny{32.52}} & \tiny{16.30} & \tiny{12.34} & \textbf{\tiny{18.84}} & \tiny{6.50}\\\tiny{coma} & \tiny{54.28} & \textbf{\tiny{70.54}} & \tiny{16.26} & \tiny{43.54} & \textbf{\tiny{67.36}} & \tiny{23.82} & \tiny{29.46} & \textbf{\tiny{57.78}} & \tiny{28.32} & \tiny{23.84} & \textbf{\tiny{50.08}} & \tiny{26.24} & \tiny{20.60} & \textbf{\tiny{42.84}} & \tiny{22.24}\\\tiny{defocus\_blur} & \tiny{55.46} & \textbf{\tiny{64.80}} & \tiny{9.34} & \tiny{46.76} & \textbf{\tiny{59.84}} & \tiny{13.08} & \tiny{31.96} & \textbf{\tiny{50.44}} & \tiny{18.48} & \tiny{21.18} & \textbf{\tiny{36.00}} & \tiny{14.82} & \tiny{14.46} & \textbf{\tiny{22.40}} & \tiny{7.94}\\\tiny{defocus\_spherical} & \tiny{49.92} & \textbf{\tiny{67.14}} & \tiny{17.22} & \tiny{41.48} & \textbf{\tiny{66.20}} & \tiny{24.72} & \tiny{26.98} & \textbf{\tiny{56.68}} & \tiny{29.70} & \tiny{18.96} & \textbf{\tiny{39.32}} & \tiny{20.36} & \tiny{13.18} & \textbf{\tiny{23.52}} & \tiny{10.34}\\\tiny{trefoil} & \tiny{57.34} & \textbf{\tiny{70.96}} & \tiny{13.62} & \tiny{45.52} & \textbf{\tiny{67.54}} & \tiny{22.02} & \tiny{30.14} & \textbf{\tiny{59.74}} & \tiny{29.60} & \tiny{22.90} & \textbf{\tiny{50.06}} & \tiny{27.16} & \tiny{20.90} & \textbf{\tiny{43.06}} & \tiny{22.16}\\\tiny{{{$\Sigma$}}} & \tiny{53.45} & \textbf{\tiny{68.22}} & \tiny{14.77} & \tiny{43.37} & \textbf{\tiny{65.33}} & \tiny{21.96} & \tiny{29.07} & \textbf{\tiny{56.33}} & \tiny{27.26} & \tiny{20.62} & \textbf{\tiny{41.60}} & \tiny{20.98} & \tiny{16.30} & \textbf{\tiny{30.13}} & \tiny{13.84}\end{tabular}\end{table*}

\begin{table*}[h]
\caption{Accuracies for ResNeXt50 w/wo OpticsAugment evaluated on ImageNet-100 OpticsBench.}\label{tab:tab:imagenet100_corruptions_ResNeXt50_revisited}
\centering\begin{tabular}{llll|lll|lll|lll|lll}&\multicolumn{3}{c}{\scriptsize{1}}&\multicolumn{3}{c}{\scriptsize{2}}&\multicolumn{3}{c}{\scriptsize{3}}&\multicolumn{3}{c}{\scriptsize{4}}&\multicolumn{3}{c}{\scriptsize{5}}\\\tiny{Corruption} & \tiny{clean} & \tiny{\textbf{ours}} & \tiny{$\Delta$} & \tiny{clean} & \tiny{\textbf{ours}} & \tiny{$\Delta$} & \tiny{clean} & \tiny{\textbf{ours}} & \tiny{$\Delta$} & \tiny{clean} & \tiny{\textbf{ours}} & \tiny{$\Delta$} & \tiny{clean} & \tiny{\textbf{ours}} & \tiny{$\Delta$}\\\hline\tiny{astigmatism} & \tiny{44.08} & \textbf{\tiny{65.08}} & \tiny{21.00} & \tiny{34.44} & \textbf{\tiny{63.36}} & \tiny{28.92} & \tiny{22.98} & \textbf{\tiny{55.90}} & \tiny{32.92} & \tiny{14.24} & \textbf{\tiny{33.54}} & \tiny{19.30} & \tiny{10.26} & \textbf{\tiny{19.88}} & \tiny{9.62}\\\tiny{coma} & \tiny{49.28} & \textbf{\tiny{66.44}} & \tiny{17.16} & \tiny{39.28} & \textbf{\tiny{63.24}} & \tiny{23.96} & \tiny{26.64} & \textbf{\tiny{55.04}} & \tiny{28.40} & \tiny{21.42} & \textbf{\tiny{47.66}} & \tiny{26.24} & \tiny{18.66} & \textbf{\tiny{41.50}} & \tiny{22.84}\\\tiny{defocus\_blur} & \tiny{48.84} & \textbf{\tiny{62.54}} & \tiny{13.70} & \tiny{40.30} & \textbf{\tiny{58.68}} & \tiny{18.38} & \tiny{25.28} & \textbf{\tiny{49.46}} & \tiny{24.18} & \tiny{16.46} & \textbf{\tiny{34.54}} & \tiny{18.08} & \tiny{11.68} & \textbf{\tiny{20.82}} & \tiny{9.14}\\\tiny{defocus\_spherical} & \tiny{44.44} & \textbf{\tiny{64.06}} & \tiny{19.62} & \tiny{36.30} & \textbf{\tiny{63.26}} & \tiny{26.96} & \tiny{23.50} & \textbf{\tiny{55.00}} & \tiny{31.50} & \tiny{16.00} & \textbf{\tiny{37.80}} & \tiny{21.80} & \tiny{10.80} & \textbf{\tiny{21.26}} & \tiny{10.46}\\\tiny{trefoil} & \tiny{52.04} & \textbf{\tiny{67.58}} & \tiny{15.54} & \tiny{40.62} & \textbf{\tiny{64.84}} & \tiny{24.22} & \tiny{25.98} & \textbf{\tiny{56.78}} & \tiny{30.80} & \tiny{19.78} & \textbf{\tiny{45.94}} & \tiny{26.16} & \tiny{17.04} & \textbf{\tiny{38.78}} & \tiny{21.74}\\\tiny{{{$\Sigma$}}} & \tiny{47.74} & \textbf{\tiny{65.14}} & \tiny{17.40} & \tiny{38.19} & \textbf{\tiny{62.68}} & \tiny{24.49} & \tiny{24.88} & \textbf{\tiny{54.44}} & \tiny{29.56} & \tiny{17.58} & \textbf{\tiny{39.90}} & \tiny{22.32} & \tiny{13.69} & \textbf{\tiny{28.45}} & \tiny{14.76}\end{tabular}\end{table*}

\subsubsection{Further analysis}
The OpticsAugment trained ResNet50 on ImageNet-1k is reported to be robust to the primary aberrations. However, it only ranks 48/72 on the ImageNet-1k validation set, which is largely due to the restricted standard training recipe of 90 epochs: the pre-trained models evaluated here from Pytorch required a significantly higher workload of 600 epochs of training, learning more robust and better features. Despite these limitations, our ResNet50 model with OpticsAugment~\cite{mueller_opticsbench2023} ranks 20th on average on the OpticsBench image corruptions, outperforming other robust models such as NoisyMix~\cite{erichson_noisymix_2022} (24th), AugMix~\cite{hendrycks_augmix_2020} (26th) and  SIN\_IN~\cite{geirhos2018imagenet} (35th) as well as most of the EfficientNets (\eg 29th)~\cite{tan_efficientnet_2019}.

\subsubsection{Additional models}
\label{app:optics_augment_strong_resnets}
\begin{table}[h]
    \caption{\added{ResNet50 models trained on ImageNet-100 with different training strategies.}}
    \label{tab:additional_resnet50_models}
    \centering
    \begin{tabular}{cc|cc}
         Model & OpticsAugment & Clean $\uparrow$ & OpticsBench $\uparrow$ \\
         ResNet50 &   & 75.76 & 36.17 \\
         ResNet50 & $\checkmark$ & 78.62 & 61.35 \\
         ResNet50 (strong) & & 81.76 & 45.16 \\
         ResNet50 (strong) & $\checkmark$ & $\mathbf{83.32}$ & $\mathbf{75.36}$
    \end{tabular}
\end{table}
We here provide additional experiments on ImageNet-100 with different training strategies for ResNet50 models and report the results in Table~\ref{tab:additional_resnet50_models}. We first train the default baseline using the torchvision script as before with 90 epochs and no augmentations beyond random crop and flip. This yields a clean accuracy of 75.76 points. If we add the OpticsAugment data augmentation and train for 120 epochs, the clean accuracy increases to 78.62 points and the model is significantly more robust to the optical aberrations from OpticsBench. 
In addition, we train another baseline with a more advanced training script~\cite{grabinskiLargeItGets2024}. This includes long training for 300 epochs, compared to an initial 90 epochs, CutMix, mixup and RandAugment augmentations during training and other regularization methods and obtain the ResNet50 (strong) model. This increases the clean accuracy by 6.0 points and the robustness to OpticsBench image corruptions by 9.0 points. If we add OpticsAugment to this script, the clean accuracy \emph{increases} by 1.6 points. Also, the ResNet50 (strong) model with OpticsAugment gives very high accuracy on OpticsBench, similar to the \emph{clean} accuracy of the simple ResNet50 model, while outperforming the strong baseline on the clean validation dataset.

\subsubsection{Cascading AugMix \& OpticsAugment}
\label{app:optics_augment_cascading}
\begin{figure}
\centering
    \includegraphics[width=\linewidth]{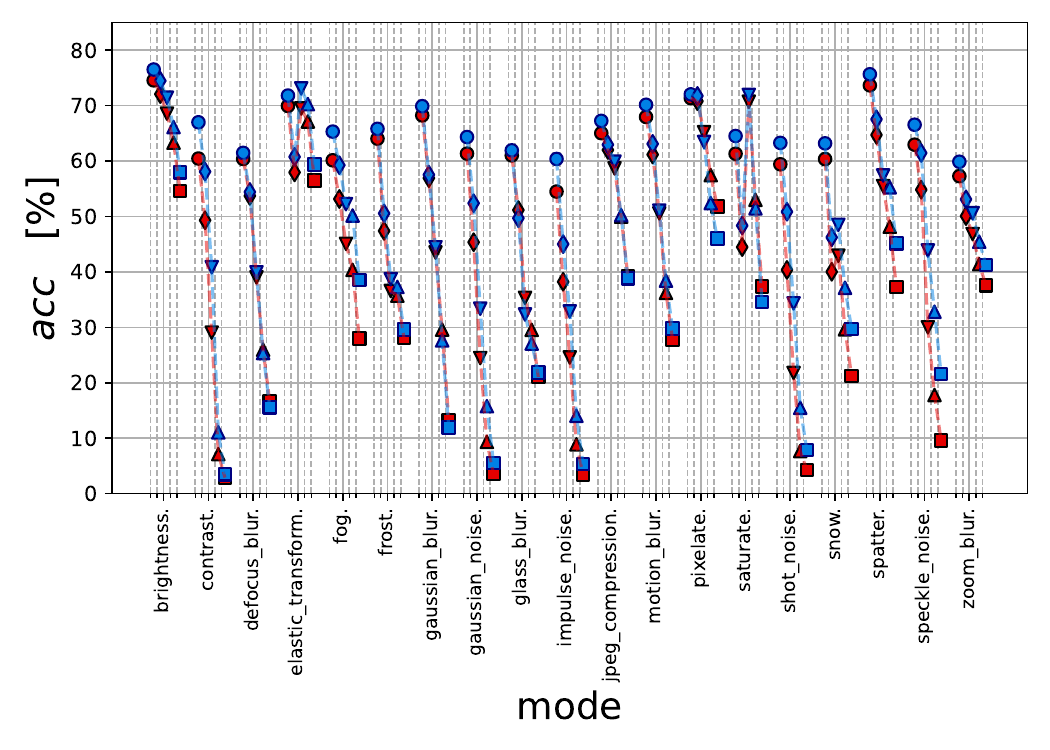}
    \caption{Cascading AugMix~\cite{hendrycks_augmix_2020} and OpticsAugment for EfficientNet: Blue represents now a cascaded application of AugMix and OpticsAugment. Red represents the OpticsAugment trained version for EfficientNet without cascading. Evaluated on 2D common corruptions on ImageNet-100-C~\cite{hendrycks_benchmarking_2019}.}
    \label{fig:cascading_common}
\end{figure}

Fig.~\ref{fig:cascading_common} shows a comparison of EfficientNet trained with OpticsAugment (red) and a cascaded application of OpticsAugment \& AugMix (blue) evaluated on 2D common corruptions~\cite{hendrycks_benchmarking_2019}. The results are also listed in Table~\ref{tab:tab:imagenet100c_corruptions_EfficientNet}. In general, the cascading improves the accuracy on 2D common corruptions~\cite{hendrycks_benchmarking_2019} across severities (last row in Table~\ref{tab:tab:imagenet100c_corruptions_EfficientNet}), with the largest average improvement of 4.3\% for severity 3. However, the improvement largely varies with the corresponding image corruption. For example, impulse noise, Gaussian noise, shot noise and speckle noise is largely improved by the combination of OpticsAugment \& AugMix. This behavior is expected since OpticsAugment uses blur kernels to mimic the various primary image aberrations, which effectively act as lowpass filters leading to image smoothing. However, the improvement on noise image corruptions varies with severity and the particular type of noise. While the gain in accuracy for the pipelined training of OpticsAugment and Augmix is 12.6\% at severity 3 and shot noise, robustness to shot noise and severity 5 is only improved by 3.6\%. Other image corruptions which largely benefit from the cascading are contrast and fog. The remaining image corruptions also benefit with few exceptions: Defocus blur, glass blur and pixelate have either the same accuracy as with OpticsAugment without AugMix or are better compensated without joint training with AugMix.

\begin{table*}[h]\centering
\caption{Accuracies for EfficientNet when cascading AugMix~\cite{hendrycks_augmix_2020} \& OpticsAugment and only OpticsAugment evaluated on ImageNet-100-c 2D common corruptions~\cite{hendrycks_benchmarking_2019}.}\label{tab:tab:imagenet100c_corruptions_EfficientNet}
\begin{tabular}{@{}l@{\,\,\,\,}l@{\,\,\,\,}l@{\,\,\,\,}l@{\,\,\,\,}|@{\,\,\,\,}l@{\,\,\,\,}l@{\,\,\,\,}l@{\,\,\,\,}|@{\,\,\,\,}l@{\,\,\,\,}l@{\,\,\,\,}l@{\,\,\,\,}|@{\,\,\,\,}l@{\,\,\,\,}l@{\,\,\,\,}l@{\,\,\,\,}|@{\,\,\,\,}l@{\,\,\,\,}l@{\,\,\,\,}l@{}}&\multicolumn{3}{c}{\scriptsize{1}}&\multicolumn{3}{c}{\scriptsize{2}}&\multicolumn{3}{c}{\scriptsize{3}}&\multicolumn{3}{c}{\scriptsize{4}}&\multicolumn{3}{c}{\scriptsize{5}}\\\tiny{Corruption} & \tiny{ours \& AugMix} & \tiny{ours} & \tiny{$\Delta$} & \tiny{ours \& AugMix} & \tiny{ours} & \tiny{$\Delta$} & \tiny{ours \& AugMix} & \tiny{ours} & \tiny{$\Delta$} & \tiny{ours \& AugMix} & \tiny{ours} & \tiny{$\Delta$} & \tiny{ours \& AugMix} & \tiny{ours} & \tiny{$\Delta$}\\\hline\tiny{brightness} & \textbf{\tiny{76.52}} & \tiny{74.56} & \tiny{-1.96} & \textbf{\tiny{74.48}} & \tiny{72.08} & \tiny{-2.40} & \textbf{\tiny{71.42}} & \tiny{68.56} & \tiny{-2.86} & \textbf{\tiny{66.14}} & \tiny{63.32} & \tiny{-2.82} & \textbf{\tiny{57.98}} & \tiny{54.60} & \tiny{-3.38}\\\tiny{contrast} & \textbf{\tiny{67.00}} & \tiny{60.46} & \tiny{-6.54} & \textbf{\tiny{58.10}} & \tiny{49.32} & \tiny{-8.78} & \textbf{\tiny{40.84}} & \tiny{29.04} & \tiny{-11.80} & \textbf{\tiny{11.06}} & \tiny{7.14} & \tiny{-3.92} & \textbf{\tiny{3.50}} & \tiny{2.90} & \tiny{-0.60}\\\tiny{defocus\_blur} & \textbf{\tiny{61.48}} & \tiny{60.38} & \tiny{-1.10} & \textbf{\tiny{54.42}} & \tiny{53.72} & \tiny{-0.70} & \textbf{\tiny{39.92}} & \tiny{39.02} & \tiny{-0.90} & \tiny{25.38} & \textbf{\tiny{25.98}} & \tiny{0.60} & \tiny{15.58} & \textbf{\tiny{16.64}} & \tiny{1.06}\\\tiny{elastic\_transform} & \textbf{\tiny{71.82}} & \tiny{69.94} & \tiny{-1.88} & \textbf{\tiny{60.76}} & \tiny{57.96} & \tiny{-2.80} & \textbf{\tiny{73.10}} & \tiny{69.48} & \tiny{-3.62} & \textbf{\tiny{70.28}} & \tiny{67.14} & \tiny{-3.14} & \textbf{\tiny{59.38}} & \tiny{56.50} & \tiny{-2.88}\\\tiny{fog} & \textbf{\tiny{65.32}} & \tiny{60.14} & \tiny{-5.18} & \textbf{\tiny{59.26}} & \tiny{53.16} & \tiny{-6.10} & \textbf{\tiny{52.22}} & \tiny{45.06} & \tiny{-7.16} & \textbf{\tiny{50.18}} & \tiny{40.38} & \tiny{-9.80} & \textbf{\tiny{38.58}} & \tiny{28.02} & \tiny{-10.56}\\\tiny{frost} & \textbf{\tiny{65.82}} & \tiny{64.04} & \tiny{-1.78} & \textbf{\tiny{50.56}} & \tiny{47.40} & \tiny{-3.16} & \textbf{\tiny{38.68}} & \tiny{36.56} & \tiny{-2.12} & \textbf{\tiny{37.30}} & \tiny{35.72} & \tiny{-1.58} & \textbf{\tiny{29.62}} & \tiny{28.08} & \tiny{-1.54}\\\tiny{gaussian\_blur} & \textbf{\tiny{69.90}} & \tiny{68.24} & \tiny{-1.66} & \textbf{\tiny{57.54}} & \tiny{56.92} & \tiny{-0.62} & \textbf{\tiny{44.44}} & \tiny{43.50} & \tiny{-0.94} & \tiny{27.66} & \textbf{\tiny{29.60}} & \tiny{1.94} & \tiny{11.90} & \textbf{\tiny{13.32}} & \tiny{1.42}\\\tiny{gaussian\_noise} & \textbf{\tiny{64.34}} & \tiny{61.34} & \tiny{-3.00} & \textbf{\tiny{52.40}} & \tiny{45.36} & \tiny{-7.04} & \textbf{\tiny{33.36}} & \tiny{24.46} & \tiny{-8.90} & \textbf{\tiny{15.78}} & \tiny{9.34} & \tiny{-6.44} & \textbf{\tiny{5.52}} & \tiny{3.56} & \tiny{-1.96}\\\tiny{glass\_blur} & \textbf{\tiny{61.94}} & \tiny{61.02} & \tiny{-0.92} & \tiny{49.70} & \textbf{\tiny{51.16}} & \tiny{1.46} & \tiny{32.32} & \textbf{\tiny{35.32}} & \tiny{3.00} & \tiny{27.08} & \textbf{\tiny{29.56}} & \tiny{2.48} & \textbf{\tiny{21.84}} & \tiny{21.04} & \tiny{-0.80}\\\tiny{impulse\_noise} & \textbf{\tiny{60.38}} & \tiny{54.50} & \tiny{-5.88} & \textbf{\tiny{45.02}} & \tiny{38.22} & \tiny{-6.80} & \textbf{\tiny{32.86}} & \tiny{24.56} & \tiny{-8.30} & \textbf{\tiny{14.06}} & \tiny{8.90} & \tiny{-5.16} & \textbf{\tiny{5.32}} & \tiny{3.46} & \tiny{-1.86}\\\tiny{jpeg\_compression} & \textbf{\tiny{67.26}} & \tiny{65.04} & \tiny{-2.22} & \textbf{\tiny{62.96}} & \tiny{62.00} & \tiny{-0.96} & \textbf{\tiny{59.88}} & \tiny{58.68} & \tiny{-1.20} & \textbf{\tiny{50.24}} & \tiny{49.98} & \tiny{-0.26} & \tiny{38.86} & \textbf{\tiny{39.30}} & \tiny{0.44}\\\tiny{motion\_blur} & \textbf{\tiny{70.16}} & \tiny{67.96} & \tiny{-2.20} & \textbf{\tiny{63.10}} & \tiny{61.18} & \tiny{-1.92} & \textbf{\tiny{51.04}} & \tiny{50.54} & \tiny{-0.50} & \textbf{\tiny{38.42}} & \tiny{36.22} & \tiny{-2.20} & \textbf{\tiny{29.78}} & \tiny{27.84} & \tiny{-1.94}\\\tiny{pixelate} & \textbf{\tiny{72.04}} & \tiny{71.36} & \tiny{-0.68} & \textbf{\tiny{71.76}} & \tiny{70.80} & \tiny{-0.96} & \tiny{63.44} & \textbf{\tiny{65.20}} & \tiny{1.76} & \tiny{52.42} & \textbf{\tiny{57.46}} & \tiny{5.04} & \tiny{46.02} & \textbf{\tiny{51.78}} & \tiny{5.76}\\\tiny{saturate} & \textbf{\tiny{64.52}} & \tiny{61.32} & \tiny{-3.20} & \textbf{\tiny{48.36}} & \tiny{44.48} & \tiny{-3.88} & \textbf{\tiny{71.92}} & \tiny{70.70} & \tiny{-1.22} & \tiny{51.48} & \textbf{\tiny{53.02}} & \tiny{1.54} & \tiny{34.58} & \textbf{\tiny{37.38}} & \tiny{2.80}\\\tiny{shot\_noise} & \textbf{\tiny{63.28}} & \tiny{59.42} & \tiny{-3.86} & \textbf{\tiny{50.86}} & \tiny{40.36} & \tiny{-10.50} & \textbf{\tiny{34.32}} & \tiny{21.76} & \tiny{-12.56} & \textbf{\tiny{15.50}} & \tiny{7.68} & \tiny{-7.82} & \textbf{\tiny{7.90}} & \tiny{4.30} & \tiny{-3.60}\\\tiny{snow} & \textbf{\tiny{63.20}} & \tiny{60.36} & \tiny{-2.84} & \textbf{\tiny{46.24}} & \tiny{40.02} & \tiny{-6.22} & \textbf{\tiny{48.48}} & \tiny{42.92} & \tiny{-5.56} & \textbf{\tiny{37.14}} & \tiny{29.66} & \tiny{-7.48} & \textbf{\tiny{29.72}} & \tiny{21.24} & \tiny{-8.48}\\\tiny{spatter} & \textbf{\tiny{75.66}} & \tiny{73.66} & \tiny{-2.00} & \textbf{\tiny{67.54}} & \tiny{64.72} & \tiny{-2.82} & \textbf{\tiny{57.40}} & \tiny{55.48} & \tiny{-1.92} & \textbf{\tiny{55.28}} & \tiny{48.16} & \tiny{-7.12} & \textbf{\tiny{45.18}} & \tiny{37.28} & \tiny{-7.90}\\\tiny{speckle\_noise} & \textbf{\tiny{66.56}} & \tiny{62.96} & \tiny{-3.60} & \textbf{\tiny{61.46}} & \tiny{54.84} & \tiny{-6.62} & \textbf{\tiny{43.86}} & \tiny{30.00} & \tiny{-13.86} & \textbf{\tiny{32.82}} & \tiny{17.80} & \tiny{-15.02} & \textbf{\tiny{21.58}} & \tiny{9.60} & \tiny{-11.98}\\\tiny{zoom\_blur} & \textbf{\tiny{59.88}} & \tiny{57.28} & \tiny{-2.60} & \textbf{\tiny{53.10}} & \tiny{50.08} & \tiny{-3.02} & \textbf{\tiny{50.58}} & \tiny{46.82} & \tiny{-3.76} & \textbf{\tiny{45.46}} & \tiny{41.40} & \tiny{-4.06} & \textbf{\tiny{41.28}} & \tiny{37.60} & \tiny{-3.68}\\\tiny{{{$\Sigma$}}} & \textbf{\tiny{66.69}} & \tiny{63.89} & \tiny{-2.79} & \textbf{\tiny{57.24}} & \tiny{53.36} & \tiny{-3.89} & \textbf{\tiny{49.48}} & \tiny{45.14} & \tiny{-4.34} & \textbf{\tiny{38.09}} & \tiny{34.66} & \tiny{-3.43} & \textbf{\tiny{28.64}} & \tiny{26.02} & \tiny{-2.61}\end{tabular}
\end{table*}

\subsubsection{Training with OpticsAugment on MSCOCO}
\label{app:training}
Table~\ref{tab:mscoco_training} lists different variants for training a Faster R-CNN on MSCOCO with OpticsAugment. 
\begin{table}[!htb]
    \centering
    \caption{Training variants of Faster R-CNN with ResNet50 backbone.}
    \label{tab:mscoco_training}
    \begin{tabular}{llc}
        Faster R-CNN & ResNet50 &\\
        \hline 
        & (frozen) &  
        \\
         & +OpticsAugment (frozen) &
        \\
        & +AugMix~\cite{hendrycks_augmix_2020,croce2021robustbench} &
        \\
        & DeepAugment~\cite{hendrycks_many_2021,croce2021robustbench} &
        \\
         & PyTorch, (partially frozen) &
        \\
         + OpticsAugment & PyTorch, (partially frozen)    &
         \\
         + OpticsAugment5 & PyTorch, (partially frozen)    &
    \end{tabular}
\end{table}
First, we train a Faster R-CNN with an ImageNet pre-trained ResNet50 backbone applied as a Feature Pyramid Network (FPN)~\cite{lin_feature_2017} as a baseline. We follow a simple training recipe~\cite{noauthor_visionreferencesdetection_082023} and train for 26 epochs on MS COCO~\cite{fleet_microsoft_2014} with batch size 2 and linear scale learning rate rule~\cite{goyal_accurate_2018} on a single A6000 GPU.
In addition, the same training recipe is used to train the Faster R-CNN using OpticsAugment~\cite{mueller_opticsbench2023} to improve robustness to primary aberrations. 
The last three backbone layers of the non-robust pre-trained ResNet50 are not frozen, so the data augmentation also affects the backbone features, making them potentially robust to primary aberrations.

Motivated by existing work~\cite{yamada_does_2022,vasconcelos_proper_2022}, we additionally used robust backbones (DeepAugment, AugMix, OpticsAugment) as initialization for fixed-feature and partial fixed-feature transfer learning, but did not find any improvement compared to a standardly trained model. 
The PyTorch pre-trained ImageNet ResNet50 backbone and data augmentation on MS COCO gave the best results: compared to a standard Faster R-CNN with ResNet50 backbone OpticsAugment improved across severities and corruptions of on average +7.7\% in mAP. 

The baseline model has 37.8\% mAP on the clean MSCOCO validation dataset in Table~\ref{tab:tab:mscoco_val2017_corruptions0} and an average mAP of 15.4\% on all image corruptions. The OpticsAugment variants have similar mAP on the clean data and clearly outperform the model on the various image corruptions of all severities by up to 7.7\% mAP on average. The largest gains for the OpticsAugment trained models are obtained for defocus \& spherical and for trefoil. The OpticsAugment models with and without aligned kernels perform similarly with an average mAP of 23.1\%. Thus, the effect of kernel alignment on training appears to be negligible. We also train a model with OpticsAugment based on severity 5, \ie stronger augmentation during training. Although, this sometimes increases the mAP at high severities, it decreases mAP at lower severities, as expected. We conclude that choosing a medium severity for training with OpticsAugment seems to be a reasonable choice.
\begin{table*}[h]\centering
\caption{Mean Average Precision evaluated on MSCOCO-OpticsBench. Our training with OpticsAugment on MSCOCO achieved an average performance gain of +7.7\% in mAP across corruptions and severities and compared to the Faster R-CNN with pretrained pytorch backbone. The largest difference is obtained for defocus \& spherical (9.33\%) and trefoil (8.8\%).}\label{tab:tab:mscoco_val2017_corruptions0}
\scriptsize
\begin{tabular}{@{}l@{\,\,\,\,}l@{\,\,\,\,}l@{\,\,\,\,}l@{\,\,\,\,}l@{\,\,\,\,}l@{\,\,\,\,}l@{\,\,\,\,}|l@{\,\,\,\,}l@{\,\,\,\,}l@{\,\,\,\,}l@{\,\,\,\,}l@{\,\,\,\,}|@{\,\,\,\,}l@{\,\,\,\,}l@{\,\,\,\,}l@{\,\,\,\,}l@{\,\,\,\,}l@{}}& & \multicolumn{5}{c}{\scriptsize{astigmatism}}&\multicolumn{5}{c}{\scriptsize{coma}}&\multicolumn{5}{c}{\scriptsize{defocus blur}}\\\tiny{Model} & \tiny{Val} & \tiny{1} & \tiny{2} & \tiny{3} & \tiny{4} & \tiny{5} & \tiny{1} & \tiny{2} & \tiny{3} & \tiny{4} & \tiny{5} & \tiny{1} & \tiny{2} & \tiny{3} & \tiny{4} & \tiny{5}\\\hline 
Faster-R-CNN (AugMix backbone) & 30.7 & 19.7 & 15.7 & 10.1 & 5.7 & 3.5
& 22.0 & 16.9 & 10.5 & 7.7 & 6.6 & 22.7 & 19.1 & 12.5 & 7.9 & 5.0 \\
Faster-R-CNN (Clean backbone) & 30.5 & 17.1 & 13.5 & 9.0 & 5.7 & 3.8 & 19.1 & 14.2 & 9.1 & 7.0 & 6.1 & 20.8 & 16.9 & 11.4 & 7.6 & 5.1 \\
Faster R-CNN (OpticsAugment backbone) & 29.9 & 21.2 & 17.1 & 9.6 & 5.9 & 3.9 
& 23.4 & 18.3 & 12.3 & 9.0 & 7.6 
& 22.1 & 18.2 & 11.3 & 7.3 & 4.9 \\
\hline 
Faster R-CNN (pretrained torch backbone) & \textbf{37.8} & 23.8 & 19.2 & 13.4 & 8.6 & 5.7 
& 25.5 & 19.6 & 13.3 & 10.3 & 9.0 
& 26.8 & 22.5 & 15.5 & 10.8 & 7.4 \\
Faster R-CNN (OpticsAugment on MSCOCO) & 37.7 & \textbf{28.8} & 26.0 & 21.1 & 14.6 & 9.9 
& \textbf{30.0} & \textbf{25.4} & 21.2 & 17.7 & 16.0 
& \textbf{28.3} & 24.8 & 19.1 & \textbf{14.2} & 9.8 \\
Faster R-CNN (OpticsAugment5 on MSCOCO) & 37.7 &27.1 & 23.9 & 19.8 & \textbf{15.2} & \textbf{11.2} 
& 28.8 & 24.2 & 20.6 & \textbf{17.8} & \textbf{16.5} 
& 27.3 & 23.6 & 17.9 & 13.8 & \textbf{9.9} \\
OpticsAugment(OpticsAugmentAligned on MSCOCO) & \textbf{37.8} & \textbf{28.8} & \textbf{26.1} & \textbf{21.3} & 14.6 & 9.8 
& 29.8 & 25.1 & \textbf{21.3} & 17.4 & 15.9 
& 28.2 & \textbf{25.0} & \textbf{19.3} & 14.0 & 9.8 
\end{tabular}\end{table*}

\begin{table*}[h]\centering
\caption{Mean Average Precision (mAP) evaluated on MSCOCO-OpticsBench for different training schemes (contd.)}\label{tab:tab:mscoco_val2017_corruptions1}
\scriptsize
\begin{tabular}{@{}llllll|lllll|l@{}}&\multicolumn{5}{c}{\scriptsize{defocus \& spherical}}&\multicolumn{5}{c}{\scriptsize{trefoil}} & $\Sigma$ \\\tiny{Model} & \tiny{1} & \tiny{2} & \tiny{3} & \tiny{4} & \tiny{5} & \tiny{1} & \tiny{2} & \tiny{3} & \tiny{4} & \tiny{5} & \\
\hline 
Faster R-CNN (Augmix backbone) & 21.0 & 18.4 & 11.6 & 7.4 & 4.9 
& 25.0 & 20.7 & 13.6 & 10.2 & 8.7 & 13.1
\\ 
Faster R-CNN (Clean) &
 18.6 & 15.1 & 9.6 & 7.0 & 5.5 
& 22.0 & 17.0 & 11.2 & 8.9 & 8.0 & 11.6
\\ 
Faster R-CNN (OpticsAugment backbone) 
& 22.9 & 20.9 & 12.9 & 8.9 & 6.6 
& 26.0 & 22.7 & 15.4 & 12.0 & 10.6 & 15.2
\\ 
\hline
Faster R-CNN (pretrained torch backbone) &
25.4 & 22.2 & 15.4 & 11.3 & 8.6 
& 29.4 & 24.0 & 17.1 & 14.0 & 12.7 & 15.4 
\\ 
Faster R-CNN (OpticsAugment on MSCOCO) & 
\textbf{30.3} & \textbf{29.7} & \textbf{27.1} & \textbf{23.7} & 18.6 
& \textbf{33.2} & \textbf{30.7} & \textbf{27.8} & \textbf{25.6} & 24.0 & \textbf{23.1}
\\
Faster R-CNN (OpticsAugment5 on MSCOCO) &
28.5 & 27.8 & 25.6 & 24.4 & \textbf{20.9} 
& 32.3 & 29.2 & 26.3 & 24.9 & \textbf{24.2} & 22.5
\\ 
Faster R-CNN (OpticsAugmentAligned on MSCOCO) & \textbf{30.3} & \textbf{29.7} & 27.0 & 23.6 & 18.5 
& \textbf{33.2} & 30.6 & 27.6 & \textbf{25.6} & 23.9 & 23.1
\end{tabular}\end{table*}

\section{More Evaluations on object detection -- NuImages}
\label{app:additional_detection_datasets}\label{app:nuimages_rankings}
Here we present evaluations with additional models on OpticsBench applied to the street scene dataset NuImages~\cite{caesar_nuscenes_2020}.
First, we show the validation mAP and model rankings for the evaluation of OpticsBench corruptions on NuImages. Changing the dataset reduces the mAP values on the NuImages validation dataset that we use for testing. The mAP for Faster R-CNN is reduced by 13.4\% from 62.1\% to 48.7\%, which is about the same as the loss of mAP when applying the MS COCO and severity 1 image corruptions. Thus, the additional application of severities 1-5 to NuImages has a greater impact on absolute mAP. Fine-tuning would reduce this gap in mAP.
\begin{figure}[h]
    \centering
    \includegraphics[width=.9\linewidth]{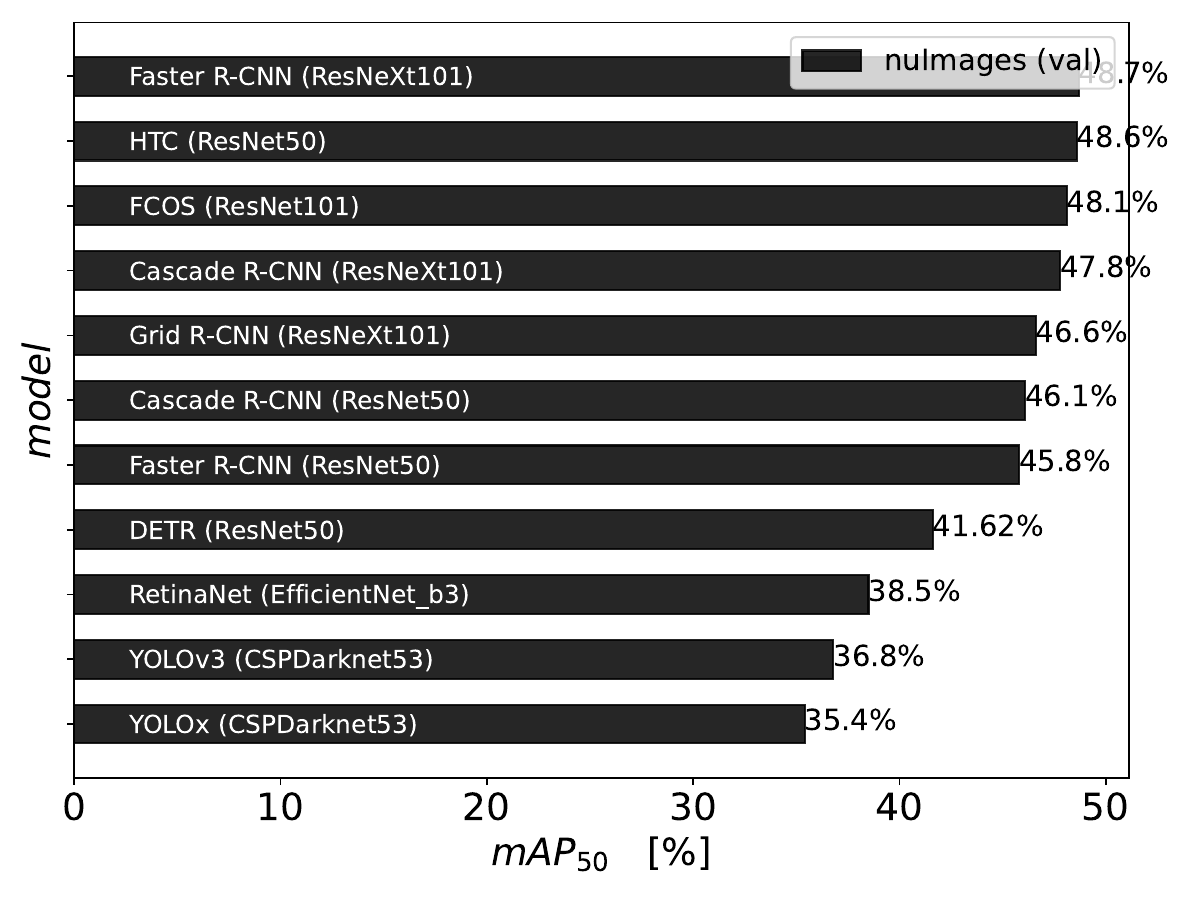}
    \caption{NuImages validation dataset mAP\_50 for selected models.}
    \label{fig:nuimages_val}
\end{figure}

Figs.~\ref{fig:app_rankings_nuimages_a} and~\ref{fig:app_rankings_nuimages_b} show the ranking with the baseline (defocus\_blur) for the selected object detection models on NuImages and all severities. 
Severities 1 and 2 are relatively similar to the baseline, but with a different offset. For severities 3-5 in Fig.~\ref{fig:app_rankings_nuimages_b}, the order of some models is reversed. Obviously, there are typical individual distribution shifts, which the models handle differently. 
For severity 4 in Fig.~\ref{fig:app_rankings_nuimages_b} Faster R-CNN has slightly higher mAP\_50 for trefoil compared to the neighbors YOLOv3 and YOLOx. The coma image corruption for severity 3 in Fig.~\ref{fig:app_rankings_nuimages_b} is handled worse for Cascade R-CNN than for Grid R-CNN, although they have comparable baseline mAP. 
\newcommand{\rankSz}{0.95\linewidth}

\begin{figure}
    \centering
    \begin{tabular}{@{}r@{}c@{}}
    \rotatebox{90}{\phantom{blabubbb}severity 1}&
        \includegraphics[width=\rankSz]{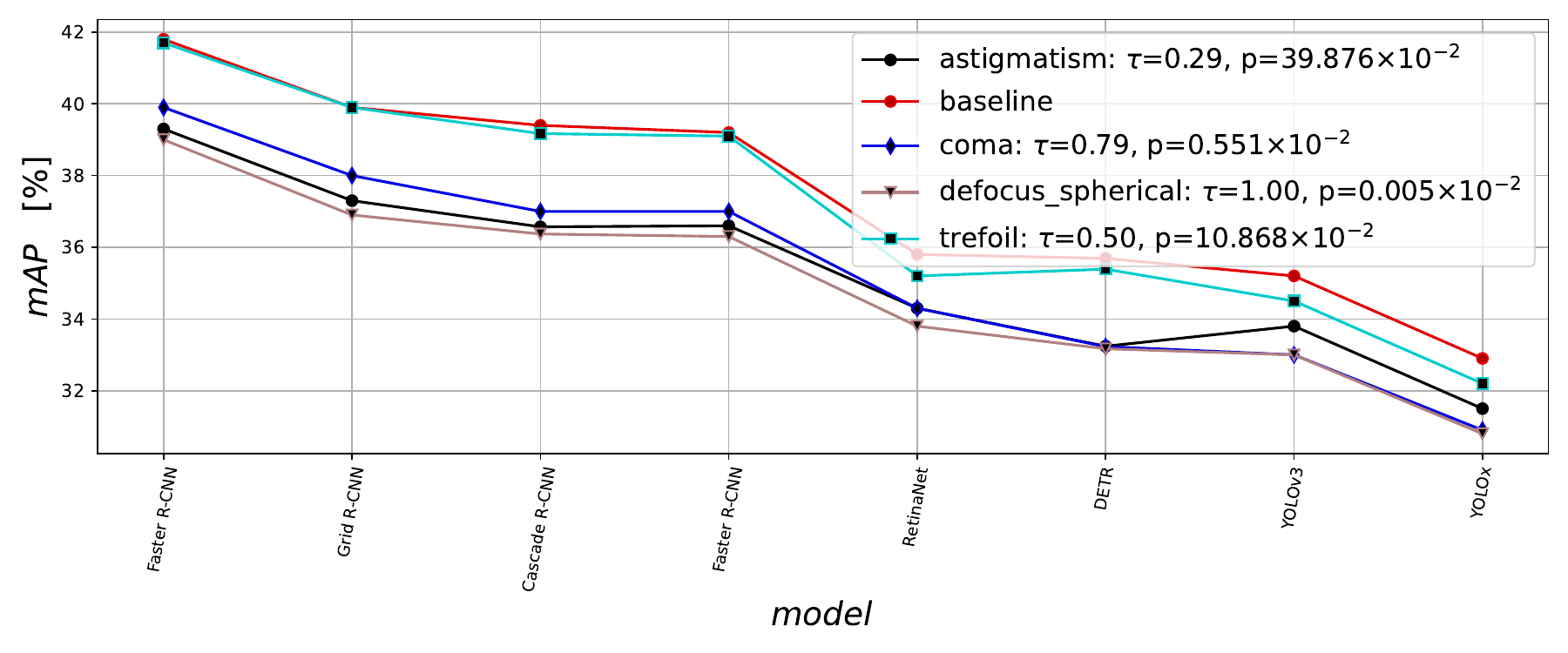}\\
         \rotatebox{90}{\phantom{blabubbb}severity 2}&
        \includegraphics[width=\rankSz]{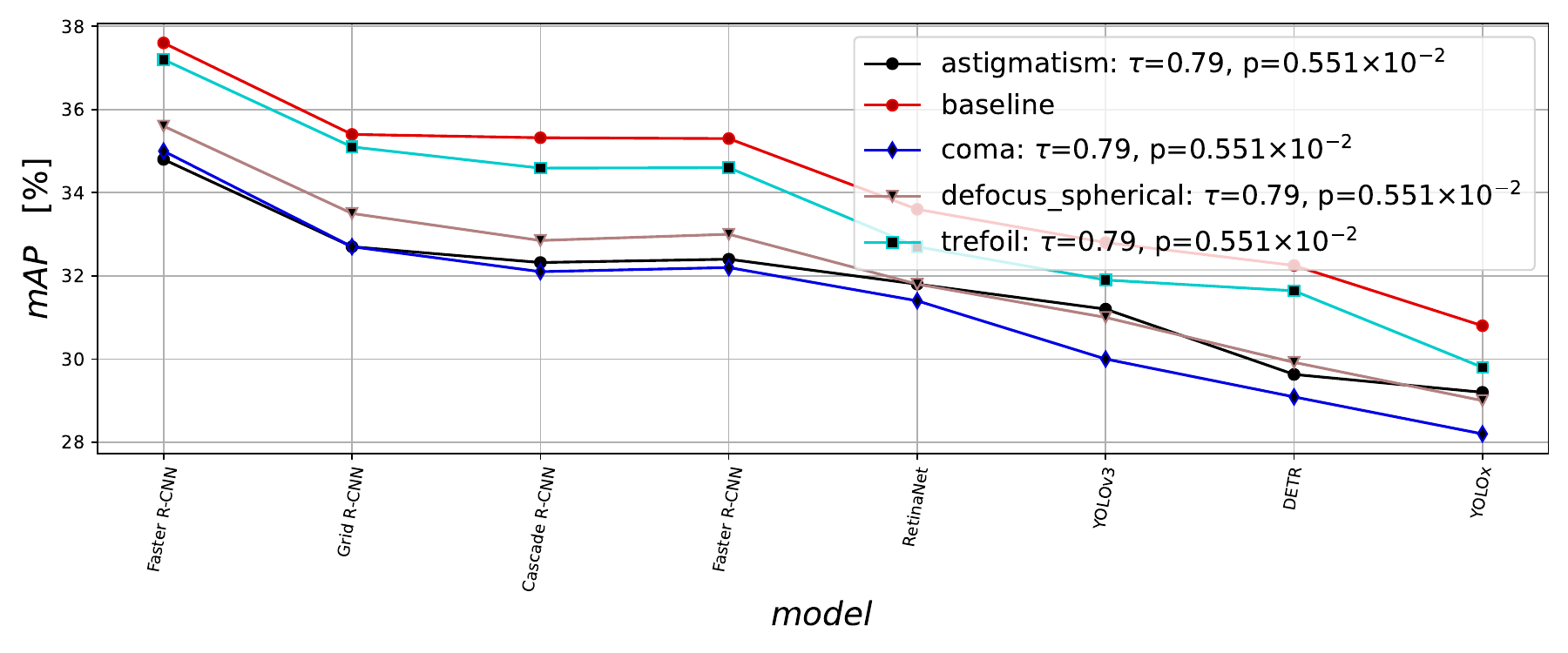}
    \end{tabular}
    \caption{Ranking comparisons on the NuImages validation dataset + OpticsBench. Each subfigure shows the mAP\_50 for all DNNs at a single severity.}\label{fig:app_rankings_nuimages_a}
\end{figure}
\begin{figure}
    \centering
    \begin{tabular}{@{}r@{}c@{}}
         \rotatebox{90}{\phantom{blabubbb}severity 3}&
        \includegraphics[width=\rankSz]{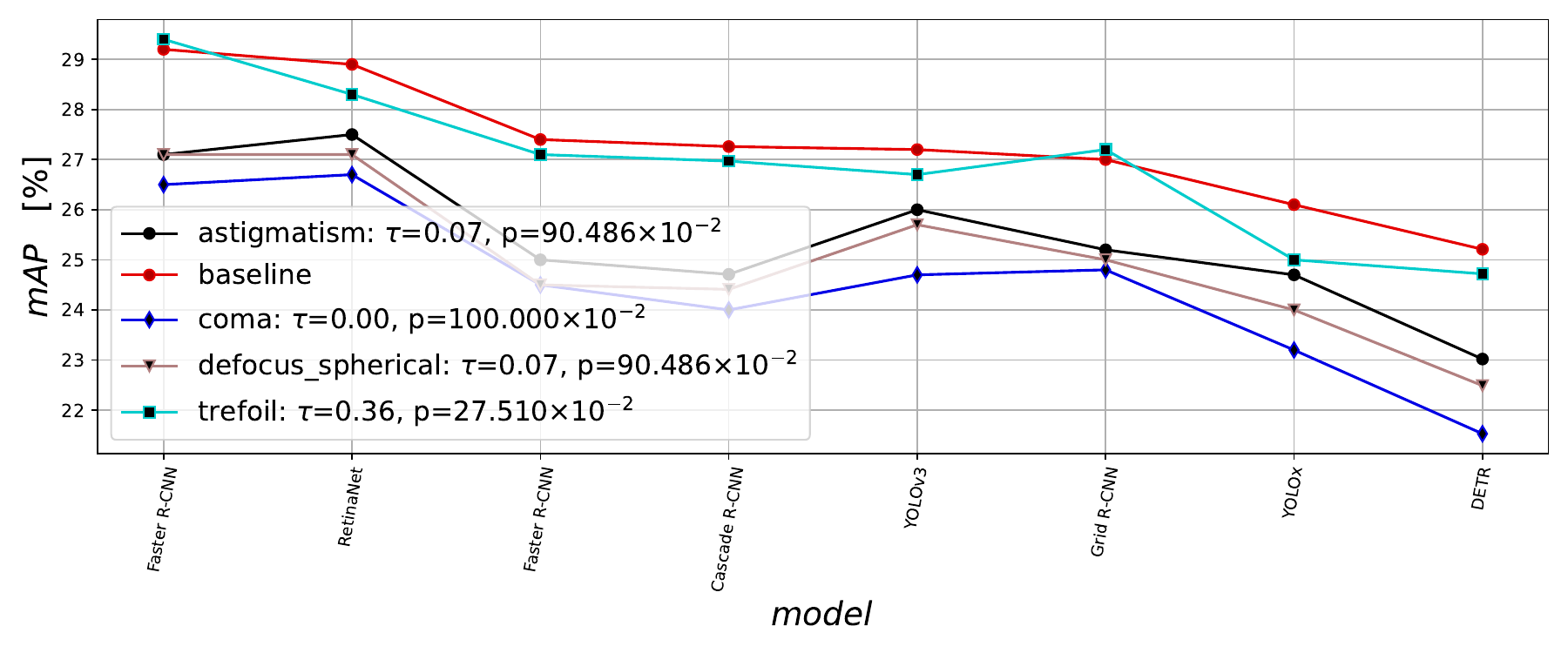}\\
         \rotatebox{90}{\phantom{blabubbb}severity 4}&
        \includegraphics[width=\rankSz]{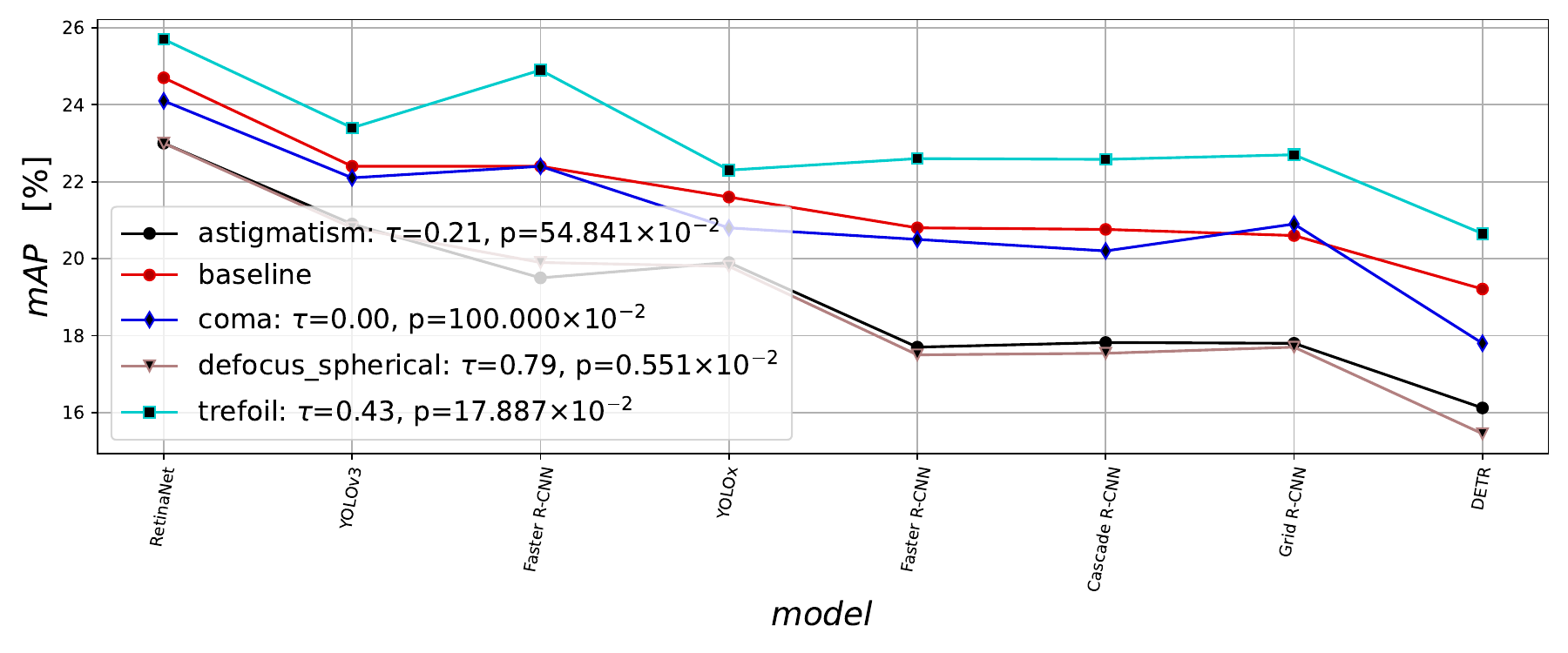}\\
         \rotatebox{90}{\phantom{blabubbb}severity 5}&
        \includegraphics[width=\rankSz]{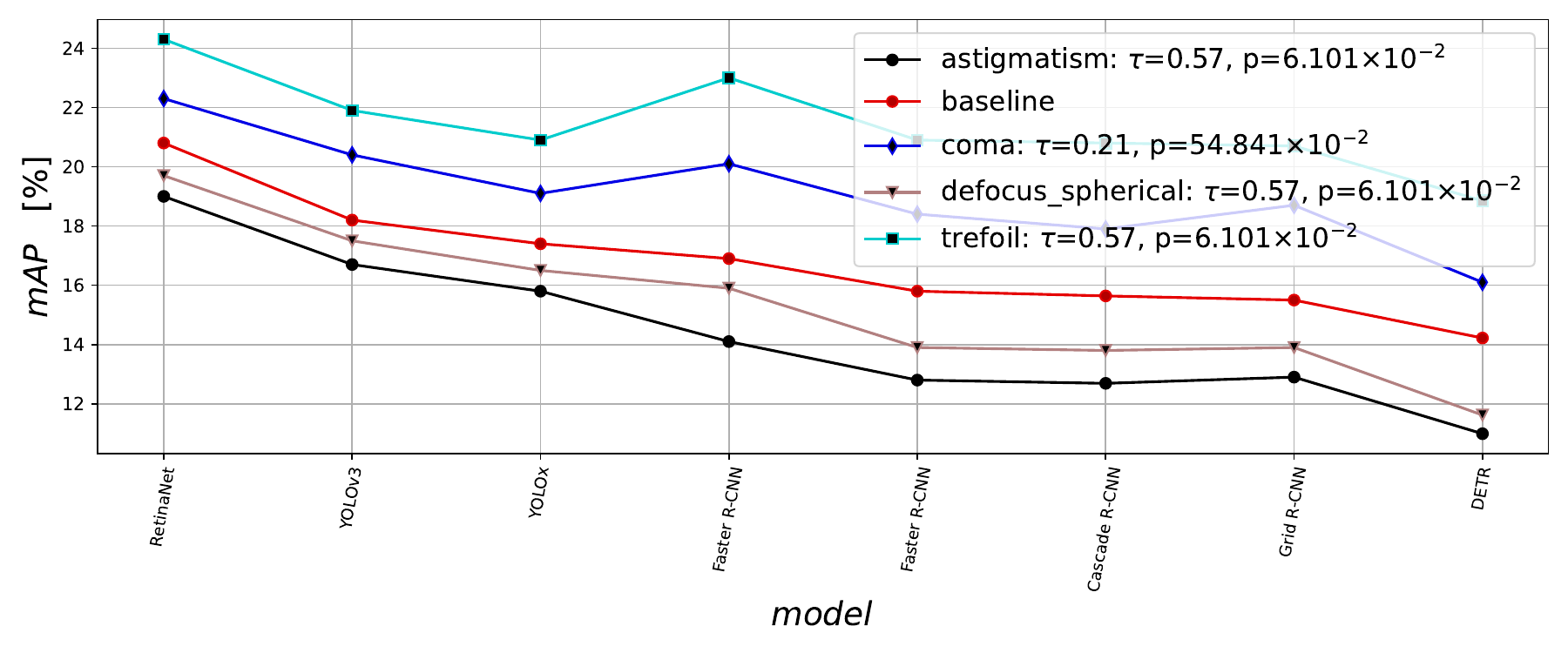}
    \end{tabular}
    \caption{Ranking comparisons on the NuImages validation dataset + OpticsBench. Each subfigure shows the mAP\_50 for all DNNs at a single severity.}\label{fig:app_rankings_nuimages_b}
\end{figure}

\section{Additional analysis on LensCorruptions}
\label{app:additional_lc}
We continue the analysis of the impact of LensCorruptions on individual models. 
The ConvNeXt model (dark blue) in Fig.~\ref{fig:distance_and_quality_classification_vs_accuracy} (bottom) has consistently the highest accuracy.
The base EfficientNet model (purple) has \SI{3.8}{\percent} less accuracy than ConvNeXt, which decreases to an average of \SI{11.6}{\percent} for the lens corruptions, which is especially seen at lower lens qualities. At high lens quality all three models have similar accuracies and approximate their validation accuracy. The ViT starts with a \SI{2.6}{\percent} higher validation accuracy than the EfficientNet and largely outperforms the smaller model by an average \SI{7.1}{\percent} on the lens corruptions. Compared to ConvNeXt, the ViT has \SI{1.2}{\percent} lower validation accuracy, which increases to \SI{4.5}{\percent} on the lens corruptions.

Table~\ref{tab:detection_robustness_lenses_overview} lists the evaluated detection models together with their mean Average Precision (mAP) on the MSCOCO subset without any corruption applied, the average LensCorruptions and OpticsBench mAPs. We continue here with a more detailed analysis. 
The large and slow DINO model has the absolute best average mAP on the lens corruptions with \SI{50.4}{\percent} mAP. While the DINO model has a reduction of \SI{8.5}{\percent} with respect to the validation mAP, YOLO-X is the most robust DNN with a reduction of \SI{6.9}{\percent} with respect to its initial validation mAP, followed by RetinaNet with a reduction of \SI{6.6}{\percent}. The latter is the most robust model on OpticsBench, while the large DINO model is the most robust in absolute terms. 

The combination of training script and model is crucial. YOLO-X is trained with HSV color augmentation and others, which may help to defend the added chromatic aberrations from OpticsBench. Using additional data augmentation such as OpticsAugment may be beneficial and improve the results.

\end{document}